\documentclass[final,12pt]{clear2025} 

\title[Shapley-PC]{Shapley-PC: Constraint-based Causal Structure Learning with a Shapley Inspired Framework}
\usepackage{times}

\clearauthor{\Name{Fabrizio Russo} \and
 \Name{Francesca Toni} \Email{$\{$fabrizio, ft$\}$@imperial.ac.uk}\\
 \addr Imperial College London, UK}

\makeatletter
\newcommand{\dontusepackage}[2][]{%
  \@namedef{ver@#2.sty}{9999/12/31}%
  \@namedef{opt@#2.sty}{#1}}
\makeatother

\dontusepackage{algorithm2e}

\usepackage[utf8]{inputenc} 
\usepackage[T1]{fontenc}    
\usepackage{booktabs}       
\usepackage{amsfonts}       
\usepackage{nicefrac}       
\usepackage{microtype}      
\usepackage{xcolor}         
\usepackage{tikz}
\usetikzlibrary{decorations.pathreplacing,calligraphy}
\tikzstyle{cnode}=[draw,circle,inner sep=1pt,minimum size=.5cm]

\usepackage{latexsym}
\usepackage{wrapfig}
\usepackage{multicol}

\SetCommentSty{mycommfont}

\usepackage{MnSymbol}
\usepackage{multirow}
\usepackage{algorithm}

\usepackage{array}

\def\Adjmat{\ensuremath{\mathbf{W}}}
\def\DAG{\ensuremath{\mathcal{G}}}
\def\SKEL{\ensuremath{\mathcal{C}}}

\def\feats{\ensuremath{\mathbf{X}}}

\def\node{\ensuremath{v}}

\def\nodeSet{\ensuremath{\mathbf{V}}}
\def\edgSet{\ensuremath{E}}
\def\noise{\ensuremath{u}}

\def\condSet{\ensuremath{\mathbf{S}}}
\newcommand{\given}{ \mid }

\def\nullH{\ensuremath{\mathcal{H}_0}}
\def\altH{\ensuremath{\mathcal{H}_0}}

\newcommand{\independent}{\,\mbox{\Large{\raisebox{-0.02em}{\rotatebox[origin=c]{90}{$\models$}}}}\,}
\newcommand{\dependent}{\mbox{\LARGE{\raisebox{-0.05em}{\rotatebox[origin=c]{-80}{{$\setminus$}}}}\!\!\!\!\independent}}
\newtheorem*{excont}{Example \continuation}
\newcommand{\continuation}{??}
\newenvironment{continueexample}[1]
 {\renewcommand{\continuation}{\ref{#1}}\excont[continued]}
 {\endexcont}

\definecolor{cadmiumgreen}{rgb}{0.0, 0.75, 0.40}
\definecolor{richcarmine}{rgb}{0.85,0.00,0.23}
\definecolor{JungleGreen}{HTML}{379f9f}

\usepackage{todonotes}

\begin{document}

\maketitle

\begin{abstract}%
Causal Structure Learning (CSL), also referred to as causal discovery, amounts to extracting causal relations among variables in data. CSL enables the estimation of causal effects from observational data alone, avoiding the need to perform real life experiments. Constraint-based CSL leverages conditional independence tests to perform causal discovery. We propose \emph{Shapley-PC}, a novel method to improve constraint-based CSL algorithms by using Shapley values over the possible conditioning sets, to decide which variables are responsible for the observed conditional (in)dependences. We prove soundness, completeness and asymptotic consistency of Shapley-PC and run a simulation study showing that our proposed algorithm is superior to existing versions of PC.
\end{abstract}


\section{Introduction}
\label{sec:intro}

Causal Structure Learning (CSL), also referred to as causal discovery, is the process of extracting causal relationships among variables in data, and represent them as graphs. Learning structural relations is important because of their causal interpretation. It corresponds to collecting and validating, with data, the assumptions necessary to perform causal inference, e.g. using causal graphical models~\citep{peters2017elements} or Functional Causal Models (FCM)~\citep{spirtes2000causation,pearl2009causality}. These models allow the estimation of causal effects, such as the impact of an action or treatment on an outcome. Causal effects are ideally discovered through 
real life experiments in the form of randomised control trials, but these can be expensive, time consuming or unethical, e.g. in establishing if smoking causes cancer, one would need some of the experiment's subjects to take up smoking.
Thus, it is important to be able to use observational, as opposed to experimental, data to study causes and effects~\citep{peters2017elements, scholkopf2021toward}.

CSL has been studied extensively in various settings and several methods have been proposed to address it (see e.g.~\citep{glymour2019review,vowels2022d, ZANGA2022survey} for overviews). 
The literature includes three classes of methods: constraint-based, score-based and FCM-based methods. In this paper, we focus on constraint-based methods, and provide a novel CSL algorithm of this class.

Constraint-based methods use conditional independence tests and graphical rules based on d-separation~\citep{pearl2009causality} to recover as much of the causal structure as possible, under different assumptions~\citep{colombo2014order}. 
Under the assumption of causal sufficiency, i.e. that no latent common causes are present in the data, the PC\footnote{From its creators' names: \textbf{P}eter Spirtes and \textbf{C}lark Glymour.} algorithm~\citep{spirtes2000causation} recovers graphs encoding as much of the discoverable relations as possible (see \S\ref{sec:related}).
Depending on the assumptions, the output of constraint-based methods may be sound and complete~\citep{spirtes2000causation} and asymptotically consistent~\citep{kalisch2007estimating, harris2013consistent}. However, with a finite sample, errors can emerge from the several conditional independence tests performed. 

The novel constraint-based method proposed in this paper improves the performance of PC, as well as other methods built on PC to mitigate its limitations on finite samples~\citep{ramsey2006CPC,colombo2014order,ramsey2016improving}, by using a novel perspective on CSL. Specifically, our method analyses the results of conditional independence tests using the game-theoretical concept of Shapley values~\citep{shapley1953value}. Generally, Shapley values quantify the contribution of individual entities to an output created by a group of entities. They have been used in settings ranging from economics~\citep{ichiishi1983game} to machine learning~\citep{lundberg2017shap,Frye2020asymshap,heskes2020causalshap,teneggi2023shapxrt} and root cause analysis~\citep{budhathoki2022causal}, but, to the best of our knowledge, not for CSL.
Overall, our contributions are as follows:
\vspace{-0.2cm}
\begin{itemize}
    \item We propose a novel decision rule that can be applied to constraint-based CSL algorithms to improve their robustness to errors in the independence tests (\S\ref{sec:method}).
    \vspace{-0.2cm}
    \item We propose the \emph{Shapley-PC algorithm}, 
   integrating the novel decision rule  within the PC-Stable algorithm~\citep{colombo2014order}, proving that Shapley-PC preserves, in the sample limit, the soundness, completeness and consistency of the original PC-algorithm (\S\ref{sec:method}).
    \vspace{-0.2cm}
    \item We provide an extensive evaluation of Shapley-PC, giving empirical evidence about the value-added of our decision rule when data distributions are ``close-to-unfaithful'' \citep{ramsey2006CPC}, 
    and showing that it consistently outperforms PC-based predecessors while using the same information extracted from data (\S\ref{sec:experiments}).
    \vspace{-0.2cm}
\end{itemize}

\section{Preliminaries}
\label{sec:background}

\paragraph{Graph Notions}
A graph $\DAG = (\nodeSet, \edgSet)$, is made up of a set of nodes $\nodeSet = \{X_1, \ldots , X_d\}$ and a set of edges $\edgSet\subseteq \nodeSet \times \nodeSet$. The nodes correspond to random variables,
while the edges reflect the relationships between variables. 
A graph can be \emph{directed} if it contains only directed edges $(\leftrightarrows)$; \emph{undirected} if it only has undirected edges $(-)$ and \emph{partially directed} if it has both. The \emph{skeleton} \SKEL{} of a (partially) directed graph is the result of replacing all directed edges with undirected ones. 
A graph is \emph{acyclic} if there is no directed path (collection of directed edges) that begins and ends with the same variable, in which case it is called a Directed Acyclic Graph (DAG).
If an edge exists between two nodes, then these are adjacent. A graph is \emph{complete} if all nodes are adjacent.
The set of nodes adjacent to a node $X_i$, according to a graph $\DAG$, is denoted by $\text{adj}(\DAG, X_i)$. 
A node $X_j \in \text{adj}(\DAG, X_i)$ is called a parent of $X_i$ if $X_j\rightarrow X_i$ and
$\text{pa}(\DAG, X_i)$ is the set of parents of $X_i$. $X_i$ is a descendant of $X_j$ if there is a directed path from the latter to the former.
A triple $(X_i, X_j, X_k)$ is called an \emph{Unshielded Triple (UT)} if $X_i$ and $X_k$ are not adjacent, but each is adjacent to $X_j$, represented as $X_i-X_j-X_k$. 

Each variable takes values from its own domain.
Two variables $X_i,X_j$ are \emph{independent}, given a conditioning set $\condSet \subseteq \nodeSet \setminus \{X_i,X_j\}$, if fixing the values of the variables in \condSet{}, $X_i$  or $X_j$ does not provide any additional information about $X_j$ or  $X_i$ (resp.). In this case, we write $X_i \independent X_j \given \condSet$, call \condSet{} a \emph{separating set} for $X_i,X_j$ and say that $\condSet$ \emph{d-separates} $X_i,X_j$, 
by rendering them independent 
(see \citep[Def. 1.2.3]{pearl2009causality} for a formal definition).
A UT can be oriented as a \emph{v-structure} $X_i\rightarrow X_j \leftarrow X_k$, where $X_j$ is called a \emph{collider}, 
by virtue of d-separation, as a collider is a variable that makes dependent other two variables that are independent otherwise. Hence, if we observe $X_1 \independent X_2\given \emptyset$ (denoted $X_1 \independent X_2$ from now on)  and $X_1 \dependent X_2 \given \{X_3\}$ 
(denoted $X_1 \dependent X_2 \given X_3$ from now on) we can infer that $X_3$ is a collider for $X_1$ and $X_2$, making $X_1-X_3-X_2$ a v-structure, i.e. $X_1\rightarrow X_3 \leftarrow X_2$.

A DAG can be interpreted causally when nodes linked by directed edges are associated to causes and effects~\citep{spirtes2000causation, pearl2009causality}. This allows manipulations that represent interventions (experiments)
to estimate the causal effect of a variable upon another, without performing the actual experiments~\citep{pearl2009causality}. 
Causal \emph{sufficiency} is the assumption that no latent common causes (confounders) are present in the data. Probabilistic measures are needed in practice to relate graphs to observational data. 

\paragraph{Statistical Notions}
A joint probability distribution $P$ factorizes according to a DAG $\DAG$ if $P(\nodeSet{}) = \prod_{i=1}^d P(X_i \given \text{pa}(\DAG, X_i))$. $P$ is said Markovian w.r.t. \DAG{} if it respects the conditional independence relations entailed by \DAG{} via d-separation. In turn, $P$ is \emph{faithful} to \DAG{} if the opposite is true, i.e. DAG \DAG\ reflects all conditional independences in $P$. 
Different DAGs can imply the same set of conditional independences, in which case they form a Markov Equivalence Class (MEC,~\citep{Richardson1996CCD}). DAGs in a MEC present the same adjacencies and v-structures and are uniquely represented by a \emph{Completed Partially} DAG (CPDAG)~\citep{chickering2002learning}. A CPDAG is a partially directed graph that has a directed edge if every DAG in the MEC has it, and an undirected edge if both directions appear in different DAGs in the MEC.

A \emph{Conditional Independence Test (CIT)}, e.g. Fisher's Z~\citep{fisher1970statistical}, HSIC~\citep{Gretton2007hsic}, KCI~\citep{zhang2011kci}, SCIT~\citep{Zhang23SCIT} and ARECI~\citep{chen2024ARECI} is a procedure whereby a test statistic measuring independence is constructed with a known asymptotic distribution under the null hypothesis \nullH{} of independence. Calculating the test statistic on a given dataset allows to estimate the $p$-value (or observed significance level) of the test for that dataset, under \nullH{}. This is a measure of evidence against \nullH{}~\citep{casella2002statistical}. Under \altH{},  $p$  is uniformly distributed in the interval $[0,1]$, which allows to set a significance level $\alpha$ that represents the pre-experiment Type I error rate (rejecting \nullH{} when it is true), whose expected value is at most $\alpha$~\citep{hung97pvalue}. 
A CIT, denoted by $I(X_i, X_j \given \condSet)$, outputs an observed significance level $p$. If $I(X_i, X_j \given \condSet) = p \geq \alpha$ then $X_i \independent X_j \given \condSet$. 
Instead, if $I(X_i, X_j \given \condSet) = p < \alpha$ then we can reject \nullH{} and declare the variables dependent: $X_i \dependent X_j \given \condSet$. 
Under the alternative hypothesis of dependence, the distribution of $p$ depends on the sample size and the true value of the test statistic. However, under any assumption, the distribution of $p$ monotonically decreases and is markedly skewed towards 0~\citep{hung97pvalue}. This allows to compare $p$-values, with the highest $p$ bearing the lowest likelihood of dependence~\citep{hung97pvalue, ramsey2016improving, raghu_comparison_2018}.

\paragraph{Shapley Values}
Consider a team $\mathbf{N} = \{1, \ldots , n\}$ of players collaborating to achieve a collective value $v(\mathbf{N})$, where $v$ is a value function that assigns a real number $v(\condSet)$ to any coalition $\condSet \subseteq \mathbf{N}$. The Shapley value $\phi_v(i)$~\citep{shapley1953value} quantifies the marginal contribution of a player $i \in \mathbf{N}$ when joining any possible coalition $\condSet$, averaged over all possible configurations of $\condSet$. This contribution is weighted according to the likelihood of each coalition's occurrence. Formally:
\begin{equation}
    \label{eq:shapley}
    \phi_v(i) = \sum_{\condSet \subseteq \mathbf{N} \backslash \{i\}} \frac{\lvert \condSet\lvert!(n-\lvert \condSet\lvert-1)!}{n!} [v(\condSet \cup \{i\}) - v(\condSet)]
\end{equation}
In our method (see \S\ref{sec:method}), the "players" are the variables in our model, and $v(\condSet)$ corresponds to a $p$-value (see Eq.~\ref{eq:SIV}).
The Shapley value is widely recognized as a fair solution to credit attribution, as it satisfies four axioms that underlie its fairness definition: symmetry, efficiency, law of aggregation and the null player corollary.\footnote{See \citep{shapley1953value} for details on the axioms and e.g., \citep{young1985monotonic} in regards to the fairness definition.} 

We note that the weighting factor in Eq.~\ref{eq:shapley} ensures equal treatment of coalitions of the same size. This guarantees that the value assigned to each player (or variable, in our case) depends solely on their marginal contribution, regardless of the coalition they join.

\section{PC-based Methods: State-of-the-art}
\label{sec:related}

The literature on CSL is broadly divided into three main approaches: constraint-based, score-based and Functional Causal Model (FCM)-based methods. In this work, we propose a novel method to improve constraint-based CSL, hence we focus on this category here. For overviews of all approaches we refer the reader to~\citep{glymour2019review, vowels2022d, ZANGA2022survey}.

Constraint-based CSL algorithms are based on CITs and graphical rules based on the d-separation criterion~\citep{pearl2009causality}. 
The PC-algorithm~\citep{spirtes2000causation} operates under the assumptions of acyclicity, sufficiency, and faithfulness. It consists of three steps: 
1) building a skeleton of the graph via adjacency search;
2) analysing UTs in the skeleton and orienting them as v-structures, and 
3) orienting as many of the remaining undirected edges without creating new v-structures or cycles, using the propagation rules from~\citep{Meek1995Orient}. 
The algorithm is computationally efficient, especially for sparse graphs, and has been shown to be sound, complete~\citep{spirtes2000causation} and consistent in the sample limit~\citep{kalisch2007estimating,harris2013consistent}. However, with finite samples, its results can vary depending on the variables' ordering.

To address this important limitation, PC-Stable~\citep{colombo2014order}, renders the first step of PC order-independent by removing edges only after all tests with a given conditioning set size are performed. \cite{Tsagris2018variation} instead uses one of the speed-up heuristics from~\citep[5.4.2.4, Heuristic 3]{spirtes2000causation} which prioritises the strongest adjacencies when choosing the next test to perform, according to some probabilistic measure.~\cite{DBLP:conf/pgm/AbellanGM06} also propose to choose edges using a measure of strength, Bayesian in this case, (i) between groups of three adjacent variables with some inconsistent test or (ii) to study the removal of an edge by determining a minimum size cut sets between two nodes. For the skeleton step, we adopt the strategy from PC-Stable.

For the second step,~\cite{ramsey2006CPC} break up the faithfulness assumption into adjacency-faithfulness and orientation-faithfulness. Assuming the former (i.e. that the edges are correctly identified) Conservative-PC (CPC) orients v-structures by checking that the latter assumption is satisfied in the data: for a UT $X_1-X_3-X_2$, $X_3$ is deemed a collider only if it is found in none of the separating sets for $X_1,X_2$. Majority-PC (MPC)~\citep{colombo2014order} relaxes the orientation-faithfulness check and orients v-structures if the potential collider appears in less than half of the separating sets of the other two nodes.
PC-Max~\citep{ramsey2016improving} selects the CIT with the maximum $p$-value and only orients the v-structure if the conditioning set for this test does not contain the collider under consideration.
\cite{Tsagris2018variation} propose some extra rules: checking for acyclicity (which we adopt in our algorithm too), checking for double colliders that violate orientation-faithfulness and checking for extra colliders created by Rule 1 of~\citep{Meek1995Orient}.
Finally, ML4C~\citep{dai2023ml4c} treats the v-structure orientation as a supervised learning problem: it trains a machine learning model on synthetic examples of v-structures and then predicts a binary label to decide upon UTs at test time. 

Overall, CPC is conservative, MPC proposes a middle-ground rule, PC-Max is informed by the observed significance level of the tests, and ML4C introduces a black-box model for the estimation. Our proposed decision rule uses the same information as MPC, CPC and PC-Max, hence we select these as our baselines for comparison
. Through our proposed rule, we analyse test results with Shapley values, lowering the dependence on single wrong tests from sample data, thus improving the discovery of v-structures, and the overall accuracy of the estimated causal graph. 

Lifting the sufficiency assumption, Fast Causal Inference (FCI)~\citep{spirtes2000causation, Colombo2012RFCI} outputs partial ancestral graphs to account for latent confounders. Variants like FCI-Max~\citep{raghu_comparison_2018}, analogous to PC-Max~\citep{ramsey2016improving}, adapt collider identification for this generalised setting. When the acyclicity assumption is relaxed, the CCD algorithm~\citep{Richardson1996CCD} recovers partially directed, cyclic graphs, and FCI has been shown to extend to cyclic settings as well~\citep{mooij2020fcicycle}. While we mention these algorithms for completeness, they fall outside the scope of this paper since we focus on acyclic and sufficient systems.

\section{Shapley-PC}
\label{sec:method}
\paragraph{Shapley Decision Rule} We propose to orient v-structures based on the Shapley value of the variable under consideration to be a collider in a UT. For this, we define a principled decision rule based on game theory, that analyses the behaviour of the $p$-value of the independence tests between two variables, when adding a candidate collider to the conditioning set. Shapley values are very well suited for the task, in that they calculate the contribution of a player (a variable) upon joining a team (a conditioning set). Note that $p$-values here are treated as a measure of association between variables, akin to their interpretation and usage in~\citep{tsamardinos2006MMHC}.

Let 
$\SKEL$ be a given skeleton, $X_i-X_j-X_k$  be a UT in $\SKEL$ and 
\begin{equation}
    \label{eq:adjsets}
\mathbf{N\!\!}=\!\!\{\condSet | \condSet \subseteq \text{adj}(\SKEL, X_i)\!\setminus\! \{ X_j\} \vee \condSet \subseteq \text{adj}(\SKEL, X_k)\!\setminus\! \{ X_j\} \}
\end{equation}
be the adjacency sets of the $X_i, X_k$.
Let $n$ be the number of variables in $\text{adj}(\SKEL, X_i) \cup \text{adj}(\SKEL, X_k)$.
Then,
we define
the \emph{Shapley Independence Value (SIV)} of $X_j$ in the 
given UT as follows:
\begin{equation}
    \label{eq:SIV}
\begin{split}
   \phi_{I}(X_j, \{X_i,X_k\}) =
   \sum_{\condSet \in \mathbf{N}} w_\condSet^n [I(X_i, X_k \given \condSet \cup \{X_j\}) - I(X_i, X_k \given \condSet)]
\end{split}
\end{equation}

where $w_\condSet^n=\frac{\lvert \condSet\lvert!(n-\lvert \condSet\lvert-1)!}{n!}$ is the weighting factor from Eq.~\ref{eq:shapley}. Note that this formulation is not guaranteed to respect 
some properties satisfied by Shapley values in general ~\citep{shapley1953value}, in particular
the efficiency and symmetry axiom,
but is guaranteed to satisfy other such properties, specifically the null player corollary, since it does not depend on the weighting. Although desirable in general, these properties are not fundamental in the context of this work
as they are not conducive to identifying colliders (see Lemma~\ref{th:shap_sign}).

Applying Eq.~\ref{eq:SIV}, we recover the marginal contribution $\phi_I(X_j, \{X_i,X_k\})$ of a candidate collider $X_j$  to the $p$-value of the independence test between the other two variables $X_i, X_k$ in the UT, when it enters the conditioning set \condSet{}, regardless of the order in which it enters. Following~\citep{hung97pvalue,ramsey2016improving,raghu_comparison_2018}, the higher the $p$-value, the higher the likelihood of independence. 
Thus, the lower $\phi_I(X_j, \{X_i,X_k\})$, the lower is the contribution of variable $X_j$ to the independence of the common parents $X_i, X_k$, hence the maximum likelihood of it being a collider. 
This leads to our \emph{decision rule}: 
\begin{quote} 
for any UT $X_i-X_j-X_k$, we declare $X_j$ a collider if it has negative SIV $\phi_I$.
\end{quote}

\begin{example}
\label{ex:ex1}
 For illustration, consider the DAG in the figure below (left) and the decision to orient the UT $X_1-X_3-X_2$ from the skeleton \SKEL{} on the right. Here, $\text{adj}(\SKEL,X_1)=\{X_3,X_4\}$ and $\text{adj}(\SKEL,X_2)=\{X_3\}$ so the following (correct) test results would be considered (e.g. for $\alpha=0.05$):
 \vspace{0.3cm}
 
 \hspace{-0.5cm}
 \begin{minipage}{0.3\textwidth}
     \begin{center}
            \begin{tikzpicture}[xscale=0.8]
                            \node[cnode,minimum size=0.65cm] (1) at (0,0.5) {$X_1$};
                            \node[cnode,minimum size=0.65cm] (0) at (0,-0.5) {$X_2$};
                            \node[cnode,minimum size=0.65cm] (2) at (1.1,0) {$X_3$};
                            \node[cnode,minimum size=0.65cm] (3) at (2.5,0) {$X_4$};
                            \path[->,thick,>=stealth]
                            (1) edge (2)
                            (0) edge (2)
                            (2) edge (3)
                            (1) edge[out=0,in=155] (3)
                            ;
            \begin{scope}
                [xshift=3.8cm]
                            \node[cnode,minimum size=0.65cm] (1) at (0,0.5) {$X_1$};
                            \node[cnode,minimum size=0.65cm] (0) at (0,-0.5) {$X_2$};
                            \node[cnode,minimum size=0.65cm] (2) at (1.1,0) {$X_3$};
                            \node[cnode,minimum size=0.65cm] (3) at (2.5,0) {$X_4$};
                             \path[->,thick,>=stealth]
                             (1) edge[-] (2)
                             (0) edge[-] (2)
                             (1) edge[-,out=0,in=155] (3)
                             (3) edge[-] (2)
                             ;
            \end{scope}
            \end{tikzpicture}
\end{center}
 \end{minipage}
 \hspace{0.8cm}
 \begin{minipage}{0.65\textwidth}
     \begin{itemize}
    \item $I(X_1,X_2)=1
    \geq \alpha$ 
    (thus $X_1 \independent X_2$)
    \vspace{-0.3cm}
    \item $I(X_1,X_2\given X_3)=0 < \alpha$ 
    (thus $X_1 \dependent X_2 \given X_3$)
    \vspace{-0.3cm}
    \item $I(X_1,X_2\given X_4)=0< \alpha$ 
(thus $X_1 \dependent X_2 \given X_4$)
    \vspace{-0.3cm}
    \item $I(X_1,X_2~\!\!\given~\!\!\{X_3,X_4\})~\!\!=0<\alpha$
 (so $X_1~\!\!\dependent~\!\!X_2~\!\!\given~\!\!\{X_3,X_4\}$) 
\end{itemize}
\end{minipage}
\vspace{0.3cm}

Then our decision rule would quantify the contribution of $X_3$ to the independence of $X_1,X_2$: $\phi_I(X_3,\{X_1,X_2\})=-0.5$. Here $n\!=\!2$, so $w_\condSet^n=0.5$. 

Thus, $\phi_I(X_3,\{X_1,X_2\})= w_\condSet^n[I(X_1,X_2~\!\!\given~\!\!X_3)-I(X_1,X_2)]+w_\condSet^n[I(X_1,X_2~\!\!\given~\!\!\{X_3,X_4\})-I(X_1,X_2~\!\!\given X_4)]=0.5(0-1)+0.5(0-0)=-0.5$.
We would therefore find that $X_3$ has negative contribution to the independence of $X_1,X_2$, and correctly identify it as a collider. 
\end{example}

The correspondence between the value function in Eq.~\ref{eq:shapley} and the one we employ in Eq.~\ref{eq:SIV}, comes from fixing $X_i, X_k$ and only changing $\condSet$ to calculate SIVs for each potential collider $X_j$. This makes $I(\cdot)$ a function of $\condSet$ alone, like $v(\cdot)$ in the original formulation of Eq.~\ref{eq:shapley}. 

With correct tests as in Example~\ref{ex:ex1},  the decision rules of all CPC, MPC, PC-Max and Shapley-PC correctly infer the v-structure from the marginal independence $X_1 \independent X_2$ and the conditional dependencies between $X_1$ and $X_2$ given all subsets of other variables. 
However, our decision rule can also deal with more realistic settings, as illustrated next.

\begin{continueexample}{ex:ex1}
    Consider the scenario with the following test results 
    obtained from the data: 
    \begin{itemize}
        \vspace{-0.2cm}
        \item $I(X_1,X_2)=0.7\geq \alpha$ (thus $X_1 \independent X_2$),
        \vspace{-0.2cm}
        \item $I(X_1,X_2\given X_3)=0.01< \alpha$ (thus $X_1 \dependent X_2 \given X_3$),
        \vspace{-0.2cm}
        \item $I(X_1,X_2\given X_4)=0.1\geq \alpha$ (thus $X_1 \independent X_2 \given X_4$),
        \vspace{-0.2cm}
        \item $I(X_1,X_2\given \{X_3,X_4\})=0.75\geq \alpha$ (thus $X_1 \independent X_2 \given \{X_3, X_4\}$).
    \end{itemize}
            \vspace{-0.2cm}
    The last two tests (wrongly) render an independence.\footnote{Note that this scenario is not unlikely from data. $I(X_1,X_2\given X_4)=0.1$ is just above $\alpha$ while $I(X_1,X_2\given \{X_3,X_4\})=0.75$ is entirely wrong: for increasing sizes of the conditioning sets the data sliced accordingly becomes thinner.} 
    Here, the SIV for $X_3$ for the UT $X_1-X_3-X_2$ is
    $\phi_I(X_3,\{X_1,X_2\})=-0.03$, and our decision rule is still able to correctly identify it as a collider and orient the v-structure. Instead, the decision rules employed by MPC and CPC do not orient it because of the inconsistency between $X_1 \dependent X_2 \given \{X_3\}$ and $X_1 \independent X_2 \given \{X_3, X_4\}$, while PC-Max' decision rule does not orient the v-structure because the maximum $p$-value test contains~$X_3$.  

    If, instead, $I(X_1, X_2 \given \{X_3, X_4\}) < 0.7$, PC-Max would also have identified the marginal independence $I(X_1, X_2)=0.7$ as the maximum $p$-value, and correctly oriented the v-structure.
    
    Suppose now that the same triple evaluated above was not a v-structure but instead a chain $X_1 \rightarrow X_3 \rightarrow X_2$. With perfect CITs, we would observe $X_1 \dependent X_2$, $X_1 \independent X_2 \given X_3$, $X_1 \dependent X_2 \given X_4$, and $X_1 \independent X_2 \given \{X_3, X_4\}$. If we had tests $I(X_1, X_2)=0.8$, $I(X_1, X_2 \given X_3)=0$, $I(X_1, X_2 \given X_4)=0$, $I(X_1, X_2 \given \{X_3, X_4\})=0.7$, with the first two incorrect, then Shapley-PC would calculate a negative contribution for $X_3$ $(\phi_I=-0.05)$, wrongly deeming it a collider. Similarly, PC-Max would pick up the wrong signal, as the highest $p$-value is $p=0.8$ (a wrong test). In contrast, CPC and MPC would not orient the structure due to the inconsistency between the tests involving $X_3$ and $\{X_3, X_4\}$.
    
    This highlights that the conservativeness of CPC and MPC can be a desirable trait, preventing false positives when orienting v-structures. Their stricter decision rules, which require consistency across multiple conditioning sets, act as safeguards in cases where tests are unreliable. However, their reliance on the consistency across tests can hinder the discovery of plausible v-structures, while Shapley-PC allows a more nuanced assessment based on the tests' strength.
\end{continueexample}

Having showcased how each decision rule has got merits and disadvantages, we next integrate our Shapley-based rule into the PC-Stable algorithm, analyse the theoretical guarantees of the resulting Shapley-PC algorithm, and test its empirical performance.

\paragraph{The Shapley-PC algorithm}
We now give our end-to-end CSL algorithm, integrating our novel Shapley-based orientation rule.
Our proposed  \emph{Shapley-PC algorithm} employs our novel decision rule as sketched 
in Alg.~\ref{alg:v_structures_shap}. 
The first step is the adjacency search that outputs a skeleton $\SKEL$, input of our decision rule in Step 2. Here, we start from a complete graph (line 1) and remove edges until no more independencies are found (lines 2-7).\footnote{Note that, for lack of space, Step 1 is presented in a simplified version: in the full version~\citep{colombo2014order}, the size of the conditioning set progressively increases, for efficiency.}
\begin{algorithm}[t]
    \caption{Shapley-PC}
    \label{alg:v_structures_shap}
    \SetAlgoNoEnd
    \SetAlgoLined
    \SetAlgoNoLine
        \textbf{Input}: $I(X_i, X_j \given \condSet)\text{ }\forall\text{ }X_i, X_j \in \nodeSet, \condSet \subseteq \nodeSet \setminus (X_i, X_j)$; $\alpha$
    
            \textbf{Step 1: Adjacency Search}~\citep{colombo2014order}
            
                    \hspace{0.5cm}\nl $\SKEL \coloneq \langle \nodeSet, \edgSet \rangle, \edgSet = \nodeSet \times \nodeSet$ \tcp*{Complete Graph over \nodeSet}
                    
                    \hspace{0.5cm}\nl \For{$X_i\in \SKEL$}{
                    \hspace{0.5cm}\nl    \For{$X_j\in \text{adj}(\SKEL,X_i)$}{
                    \hspace{0.5cm}\nl        \For{$\condSet \in \mathbf{N} $}{
                    \hspace{0.5cm}\nl            \If{$I(X_i, X_j \given \condSet)\geq \alpha$}{\hspace{0.5cm}\nl $\SKEL \coloneq \langle \nodeSet, \edgSet \setminus{(X_i-X_j)} \rangle$}
                        }}}
                \hspace{0.5cm}\nl \Return{$\SKEL$\tcp*{Skeleton}}
      
            \textbf{Step 2: Orient v-structures} (Our Decision Rule)
            
              \hspace{0.5cm}\nl  \For{$X_i-X_j-X_k\in \SKEL$}{
              \hspace{0.5cm}\nl        \If{$\phi_I(X_j, \{X_i,X_k\})<0$}{
              \hspace{0.5cm}\nl            \If{$ X_i-X_j-X_k\text{ not fully directed}$}{
              \hspace{0.5cm}\nl                \If{$\text{do not add a cycle or bi-directed edge} $}{\hspace{0.5cm}\nl  
                                $\text{orient: } X_i \rightarrow X_j \leftarrow X_k$
                            }
                        }
                    }
                }  
                \hspace{0.5cm}\nl \Return{$\SKEL$\tcp*{Partially Oriented DAG}}
        
            \textbf{Step 3: Pattern Completion}~\citep{Meek1995Orient}
            
            \hspace{0.5cm}\nl  Apply Meek's rules to \SKEL{} until no more edges can be oriented
            
            \hspace{0.5cm}\nl \Return{CPDAG \tcp*{MEC of the True DAG}}
\end{algorithm} 

In Step 2, we calculate SIVs for all candidate colliders in UTs within the skeleton $X_i-X_j-X_k\in \SKEL$ (line 8). While the number of tests in this step is the same as in CPC, MPC, and PC-Max, we obtain more granular information and analyze it using SIVs. Our decision rule (lines 9-12) is to declare $X_j$ a collider if it has a negative contribution to the observed significance level for $X_i,X_k$. We apply two additional conditions: as in PC-Max, we avoid bi-directed edges by checking existing orientations; additionally, following~\citep{Tsagris2018variation}, we check for acyclicity before making the orientation. If a bi-directed edge or a cycle is introduced, the UT is not oriented.

Finally, in Step 3 (line 14), groups of three and four adjacent variables are analysed and as many undirected edges as possible are oriented, using the rules from \citep{Meek1995Orient}. 

Compared to our reference versions of the PC algorithm in the literature, our proposed Shapley-PC also focuses on Step 2. Differently from CPC~\citep{ramsey2006CPC} and MPC~\citep{colombo2014order}, we use a continuous characterisation of the degree of independence rather than a dichotomous (in)dependence relation. Additionally, we add checks for cycles and bi-directed edges that avoid creating invalid DAGs. PC-Max \citep{ramsey2016improving} also uses $p$-values to decide about colliders and checks for bi-directed edges. However, PC-Max rule is over-reliant on the test with maximum $p$-value which makes it more prone to mistakes than our proposed method.

\paragraph{Theoretical Guarantees}
Having incorporated our SIVs into the PC algorithm, we now prove that Shapley-PC retains the theoretical guarantees of the original PC: soundness, completeness~\citep{spirtes2000causation} and high-dimensional consistency~\citep{kalisch2007estimating}. 
In order to prove soundness and completeness of Shapley-PC, we need a quantitative representation of perfect independence information. We define the concept of perfect conditional independence test, or \emph{perfect CIT}: a test that is able to extract perfect conditional independence information from data.
\begin{definition}
    \label{def:perfectCIT}
     For any $X_i, X_j \in \nodeSet$ and $\condSet \subseteq \nodeSet \setminus \{X_i, X_j\}$, a \emph{perfect} CIT is defined as:
        \begin{equation*}
            \begin{split}
                &I_{\infty}(X_i,X_j\given \condSet) \!=\! 
                \begin{cases}  
                    1
                    &\hspace*{-0.2cm}\text{ if } X_i \independent X_j\given \condSet\\
                    0
                    &\hspace*{-0.2cm}\text{ otherwise } 
                \end{cases}
            \end{split}
        \end{equation*}
\end{definition}
We can then show the consistent behaviour of SIVs for evaluating if UTs should be oriented as v-structures (all proofs are in Appendix \ref{sec:proofs}).

\begin{lemma}
    \label{th:shap_sign}
    Given a skeleton \(\SKEL\), a UT \( X_i-X_j-X_k \in \SKEL \), \( X_i, X_j, X_k \in \nodeSet \), and a perfect CIT \( I_{\infty} \), the SIV of variable \( X_j \) \(\phi_{I_{\infty}}(X_j, \{X_i, X_k\}) < 0\) if and only if \( X_j \) is a collider for \( X_i \) and \( X_k \).
\end{lemma}

Lemma~\ref{th:shap_sign} states that, given correct conditional independence information (i.e. a perfect CIT), our decision rule to identify colliders based on SIVs is correct. This allows us to prove that Shapley-PC algorithm is sound and complete when assuming faithfulness or infinite data~\citep{ramsey2006CPC}.

\begin{theorem}
    \label{th:correctness}
    Let $\mathbf{P}(\nodeSet)$ be a joint distribution faithful to a DAG $\DAG = (\nodeSet, \edgSet)$, and assume access to perfect conditional independence information for all pairs $(X_i, X_j) \in \nodeSet$ given subsets $\condSet \subseteq \nodeSet \setminus \{X_i, X_j\}$. Then the output of Shapley-PC is the CPDAG representing the MEC of $\DAG$.
\end{theorem}

Beyond asymptotic correctness, the original PC algorithm has also been shown to exhibit high-dimensional consistency, in the sample limit with the number of variables growing at a slower rate than the sample, for sparse graphs and multivariate Gaussian distributions~\citep{kalisch2007estimating} or Gaussian copulas~\citep{harris2013consistent}. These results are contingent on PC only performing CITs between pairs of variables, with the size of the conditioning sets less or equal to the maximal graph degree, i.e. the maximum of the number of edges linking any node to the others. Our Shapley-PC does not alter these features, hence the consistency results are equally applicable.

By systematically evaluating the marginal effect of adding a variable to different conditioning sets, Shapley values provide a structured framework for assessing the role of a candidate collider in an UT. While our analysis of SIVs focuses on the asymptotic setting, where the weighting scheme of Shapley values does not influence results, the Shapley value framework defines the terms to aggregate and remains key to their theoretical foundation. This highlights the value of SIVs over simpler heuristics, such as empirical averages, and sets the stage for extending our results to finite-sample analyses where the weighting and axioms may become more significant.

\paragraph{Additional Properties}
Shapley-PC improves the robustness of v-structure orientation in finite-sample settings by aggregating evidence across multiple CI tests with intersecting conditioning sets. Unlike CPC and PC-Max, which rely on a single test and risk false dependencies, Shapley Independence Values (SIVs) reduce the impact of individual errors, improving reliability. Our Shapley-based decision rule is also order-independent, akin to CPC and MPC~\citep{ramsey2006CPC, colombo2014order}.

Shapley-PC mitigates the reliance on a fixed significance threshold $\alpha$ in Step 2 of the algorithm by aggregating $p$-values across tests, removing the need for manual tuning~\citep{colombo2014order}. However, as discussed in \S\ref{sec:background}, uniformly distributed $p$-values under the null hypothesis can produce false signals, which could be mitigated by transforming them to probabilities~\citep{Claassen2012bayesian}, though this is out of scope here. 

In terms of computational overhead, Shapley-PC computes SIVs exactly, leveraging only the unshielded triples (UTs) identified in Step 1 of the PC algorithm. While Shapley values are generally expensive to compute~\citep{lundberg2017shap}, this remains feasible as the number of UTs depends on graph density: sparse graphs reduce computational costs. As in CPC, MPC, and PC-Max, additional tests in Step 2 depend on the graph degree, but the majority of computation still lies in Step 1~\citep{ramsey2006CPC}. Table~\ref{tab:runtime} empirically validates this claim. 

Finally, Shapley-PC classifies nodes as colliders if their Shapley value is negative, a decision rule that is theoretically sound with infinite data. Alternative SIV-based rules could account for context-driven desiderata, such as favouring minimal SIVs, or below a threshold, in line with heuristics like majority voting~\citep{colombo2014order} or $p$-value maximisation~\citep{ramsey2016improving}.

\vspace{-0.3cm}
\section{Empirical Evaluation}
\label{sec:experiments}
\vspace{-0.2cm}
We conduct a simulation study to compare Shapley-PC against existing versions of PC in the literature (see \S\ref{sec:background}). For all methods, we use Fisher's Z~\citep{fisher1970statistical} as CIT and, in line with \citep{ramsey2016improving}, we decrease the  significance threshold for the independence tests for increasing number of nodes ($\alpha = 0.1, 0.05, 0.01$ for $|\nodeSet| = 10, 20, 50$, respectively).Details on baselines and implementation, including a comparison with KCI~\citep{zhang2011kci} and $\chi^2$\citep{pearson1900x2} tests are in Appendix~\ref{sec:exp_det}. Code can be found at \url{https://github.com/briziorusso/ShapleyPC}.

\vspace{-0.1cm}
\paragraph{Data Generating Process (DGP)} Given our theoretical guarantees for faithful and infinite data, in this section, we aim at probing our proposed Shapley-PC in 
scenarios where the distributions in the data are ``close-to-unfaithful to the true graph'' \citep{ramsey2006CPC}, which poses a considerable challenge to reliable causal discovery~\citep{Robins2003consistency,ZhangSpirtes03strong}. To this end, we adapt the strategy proposed in \citep{ramsey2006CPC}, and generate data with a proportion of weak links, likely to lead to violations of orientation-faithfulness as defined in \citep{ramsey2006CPC}, whereby inconsistent separating sets are retrieved from the independence tests. 

The procedure is as follows.
In each experiment, we first generate 10 random graphs for each combination of three parameters: graph type Erd\"{o}s-Rényi (ER~\citep{erdds1959random}) and Scale Free (SF~\citep{barabasi1999emergence}), number of nodes $|\nodeSet| \in \{10, 20, 50\}$ and density $d=\{1, 2, 4\}$, with $|\edgSet|=|\nodeSet|\times d$. Graphs have a maximum degree of 10.
Given the ground truth DAG, we simulate 4 different additive noise Structural Equation Models (SEMs) of the type $X_j = f_j(\text{pa}(\DAG,X_j)) + \noise_j$ for all $j \in [1,\ldots,|\nodeSet|]$ in topological order.
In the SEMs, $f_j$ is linear, with coefficients $\Adjmat$ initialised from a uniform distribution with coefficients $[-1.5,-0.5]\cup[0.5,1.5]$ for $95\%$ of the effects, and $[-0.001,0.001]$ for the remaining $5\%$. Sampling from the range $[-0.001,0.001]$ simulates the presence of weak edges.
We then derive variables' values through the equation $\feats = \Adjmat^T \feats + \noise$ where the noise \noise{} is generated from Gaussian, Exponential, Gumbel and Uniform distributions. 
Finally, we vary the number of drawn samples ($N$) in function of the number of nodes: $N=s\times|\nodeSet|$, $s \in \{100,500,1000\}$ to check how data-hungry are the different algorithms.\footnote{Compared to~\citep{ramsey2006CPC}, we decreased the number of variations in nodes and densities to give space to the analysis of the effect of different types of graphs, noise distributions and sample sizes.}
More details on the DGPs are provided in Appendix~\ref{sec:dgp_sim_det}.

\paragraph{Evaluation Metrics} 
In line with~\citep{ramsey2006CPC, ramsey2016improving}, we analyse the ability to identify v-structures (colliders), which is the focus of our proposed algorithm, alongside the overall performance in recovering the causal arrows of the true graph. For the former, we summarise precision and recall in classifying correct UTs as v-structures, using F1 score (V-F1).\footnote{We isolate the errors in orienting correctly identified UTs, in line with adjacency-faithfulness \citep{ramsey2006CPC}.} Also for the arrows we use F1, but calculated on the number of (in)correct arrowheads (AH-F1). As a reminder, precision is the number of correct classifications out of the estimated ones, while recall is out of the true ones. F1 is the harmonic mean of precision and recall. All the metrics are calculated on the output CPDAGs.
Details about the metrics are in Appendix \ref{sec:met_det}, alongside breakdowns of F1 into precision and recall
in \S\ref{sec:synth_det}. 

\begin{table*}[t]
    \caption[Shapley-PC - Synthetic Experiments]{ArrowHead (AH) and V-structure (V) F1 Scores $\pm$ std for ER$d$ and SF$d$ graphs of nodes $\lvert\nodeSet\lvert \in \{10,50\}$. $d$ is the number of edges per node in the true DAG. Bold if significantly different from the runner-up according to a t-test (see Appendix \S\ref{sec:stats} for details). Observed significance levels: 0 `***' 0.001 `**' 0.01 `*' 0.05 `.' 0.1 ` ' 1. }
    \centering
    \begin{tabular}{cc|r|clclclcl}
    \toprule
    & & \textbf{Method} & \textbf{ER2} & & \textbf{ER4} & & \textbf{SF2} & & \textbf{SF4} & \\ 
    \midrule
    \multirow{10}{*}{\textbf{\rotatebox{90}{\textbf{$\lvert$V$\lvert$=10}}}} & \multirow{5}{*}{\textbf{\rotatebox{90}{\textbf{AH-F1}}}} 
    & PC-Stable & 0.36$\pm$0.28 & & 0.15$\pm$0.16 & & 0.67$\pm$0.23 & & 0.32$\pm$0.27 & \\
    & & CPC & 0.42$\pm$0.26 & & 0.15$\pm$0.17 & & 0.74$\pm$0.13 & & 0.34$\pm$0.23 & \\
    & & MPC & 0.42$\pm$0.26 & & 0.15$\pm$0.17 & & 0.74$\pm$0.13 & & 0.36$\pm$0.23 & \\
    & & PC-Max & 0.5$\pm$0.22 & & 0.07$\pm$0.12 & & 0.73$\pm$0.2 & & 0.39$\pm$0.26 & \\
    & & Shapley-PC & \textbf{0.63$\pm$0.16} & \!\!\!\!\!\!\textbf{**} & \textbf{0.25$\pm$0.16} & \!\!\!\!\!\!\textbf{***} & \textbf{0.82$\pm$0.08} & \!\!\!\!\!\!\textbf{*} & \textbf{0.54$\pm$0.16} & \!\!\!\!\!\!\textbf{**} \\ 
    \cmidrule{2-11}
    & \multirow{5}{*}{\textbf{\rotatebox{90}{\textbf{V-F1}}}} 
    & PC-Stable & 0.46$\pm$0.36 & & 0.22$\pm$0.32 & & 0.81$\pm$0.29 & & 0.49$\pm$0.41 & \\
    & & CPC & 0.59$\pm$0.33 & & 0.23$\pm$0.33 & & 0.88$\pm$0.18 & & 0.54$\pm$0.36 & \\
    & & MPC & 0.58$\pm$0.34 & & 0.22$\pm$0.33 & & 0.88$\pm$0.18 & & 0.59$\pm$0.36 & \\
    & & PC-Max & 0.67$\pm$0.3 & & 0.09$\pm$0.24 & & 0.9$\pm$0.22 & & 0.59$\pm$0.4 & \\
    & & Shapley-PC & \textbf{0.86$\pm$0.16} & \!\!\!\!\!\!\textbf{***} & \textbf{0.36$\pm$0.42} & \!\!\!\!\!\!\textbf{***} & \textbf{0.99$\pm$0.02} & \!\!\!\!\!\!\textbf{*} & \textbf{0.84$\pm$0.23} & \!\!\!\!\!\!\textbf{**} \\ 
    \midrule
    \multirow{10}{*}{\textbf{\rotatebox{90}{\textbf{$\lvert$V$\lvert$=50}}}} & \multirow{5}{*}{\textbf{\rotatebox{90}{\textbf{AH-F1}}}} 
    & PC-Stable & 0.25$\pm$0.3 & & 0.04$\pm$0.09 & & 0.63$\pm$0.37 & & 0.25$\pm$0.35 & \\
    & & CPC & 0.4$\pm$0.36 & & 0.06$\pm$0.1 & & 0.53$\pm$0.44 & & 0.4$\pm$0.39 & \\
    & & MPC & 0.35$\pm$0.34 & & 0.04$\pm$0.09 & & 0.51$\pm$0.44 & & 0.36$\pm$0.39 & \\
    & & PC-Max & 0.63$\pm$0.27 & & 0.05$\pm$0.11 & & 0.56$\pm$0.44 & & 0.59$\pm$0.37 & \\
    & & Shapley-PC & \textbf{0.75$\pm$0.06} & \!\!\!\!\!\!\textbf{**} & \textbf{0.19$\pm$0.15} & \!\!\!\!\!\!\textbf{***} & \textbf{0.9$\pm$0.04} & \!\!\!\!\!\!\!\!\textbf{***} & \textbf{0.83$\pm$0.07} & \!\!\!\!\!\!\textbf{***} \\
    \cmidrule{2-11}
    & \multirow{5}{*}{\textbf{\rotatebox{90}{\textbf{V-F1}}}} 
    & PC-Stable & 0.31$\pm$0.37 & & 0.08$\pm$0.17 & & 0.67$\pm$0.4 & & 0.28$\pm$0.4 & \\
    & & CPC & 0.5$\pm$0.44 & & 0.14$\pm$0.24 & & 0.58$\pm$0.48 & & 0.46$\pm$0.45 & \\
    & & MPC & 0.42$\pm$0.41 & & 0.08$\pm$0.18 & & 0.57$\pm$0.49 & & 0.42$\pm$0.45 & \\
    & & PC-Max & 0.81$\pm$0.34 & & 0.12$\pm$0.27 & & 0.61$\pm$0.48 & & 0.69$\pm$0.43 & \\
    & & Shapley-PC & \textbf{0.98$\pm$0.03} & \!\!\!\!\!\!\textbf{**} & \textbf{0.48$\pm$0.36} & \!\!\!\!\!\!\textbf{***} & \textbf{1.0$\pm$0.0} & \!\!\!\!\!\!\!\!\!\textbf{***} & \textbf{0.99$\pm$0.03} & \!\!\!\!\!\!\textbf{***} \\
    \bottomrule
    \end{tabular}
    \label{tab:main_res}
\end{table*}

\paragraph{Results}
We report the results for 10 and 50 nodes graphs of different type and density for $s=1000$ samples per node in Table~\ref{tab:main_res}. Results for $|\nodeSet|=20$, $d=1$ and $s \in \{100,500\}$ are provided in Appendix \ref{sec:synth_det}, since $|\nodeSet|=20$ and $s \in \{100,500\}$ corroborate the results in Table~\ref{tab:main_res}, while for $d=1$, 
no significant variations across methods were observed.

From Table~\ref{tab:main_res}, we can see that Shapley-PC is significantly better than all other versions of PC for both ER and SF graphs of density $d=2$ and $d=4$. We conduct pairwise t-tests for difference in means and highlight the best results in bold if the best method is significantly different from the runner-up, with a significance threshold $\alpha = 0.01$ and show the interval for the observed significance level. Details on the tests are in Appendix \ref{sec:stats}. 

Interesting variations in performance can be observed across graphs' types, densities and sizes. Firstly, ER graphs are generally more challenging to retrieve than SF. Secondly, increasing density on ER graphs results to have higher impact on all algorithms than for SF graphs as evidenced by the bigger drop in performance from $d=2$ to $d=4$. 
Thirdly, for the same density $d$, a larger number of nodes improves the results. This is because a density of $d=4$ edges per node means 40 edges for a 10 nodes graph, which is very close to the maximum number of edges for the graph to remain acyclic $(|\nodeSet|(|\nodeSet|-1))/2=45$. For a graph of 50 nodes, instead, having 200 edges is only about 15\% of the way to the maximum number of edges. PC-based methods, generally, perform best on sparse graphs~\citep{kalisch2007estimating}, Shapley-PC improves performance on denser graphs. We conjecture that this is because of the increased number of tests necessary to analyse denser graphs and the ability of our method to prevent the judgment of orientation based on single wrong tests.
\setlength\intextsep{0.15cm}
\begin{wraptable}{r}{8.7cm}
    \caption{Runtime for the experiments in Table~\ref{tab:main_res}: median elapsed time in seconds for ER and SF graphs with nodes $|\nodeSet|\in\{10,20,50\}$.}
    \centering
    \begin{tabular}{r|lll|lll}
    \toprule
     & \multicolumn{3}{c|}{\textbf{ER}} & \multicolumn{3}{c}{\textbf{SF}} \\
    $|\nodeSet|$ & 10 & 20 & 50 & 10 & 20 & 50 \\
    \midrule
    PC-Stable & 0.1 & 0.6 & 4.7 & 0.1 & 0.6 & 2.4 \\
    CPC & 0.1 & 0.7 & 5.5 & 0.1 & 0.8 & 4.3 \\
    MPC & 0.1 & 0.7 & 5.4 & 0.1 & 0.8 & 4.4 \\
    PC-Max & 0.1 & 0.7 & 5.5 & 0.1 & 0.8 & 4.5 \\
    Shapley-PC & 0.1 & 0.7 & 6.3 & 0.1 & 0.8 & 5.2 \\
    \bottomrule
    \end{tabular}
    \label{tab:runtime}    
\end{wraptable}

Besides the performance metrics in Table~\ref{tab:main_res}, we compare run times in Table~\ref{tab:runtime}. We can see that PC-Stable is the method that scales best with increasing number of nodes, while adding the SIVs calculation on top of the extra tests performed by CPC, MPC and PC-Max does not add considerable time (less than 1s for $|\nodeSet|=50$, $\sim\!\!15\%$ higher than PC-Max). 
Interestingly, the extra testing is more expensive for SF graphs, as demonstrated by the bigger difference, compared to ER, between PC-Stable and all other methods. This is possibly due to the morphology of SF graphs, presenting hubs of highly connected nodes.

\begin{figure}[t]
    \centering
    \includegraphics[width=0.8\textwidth]{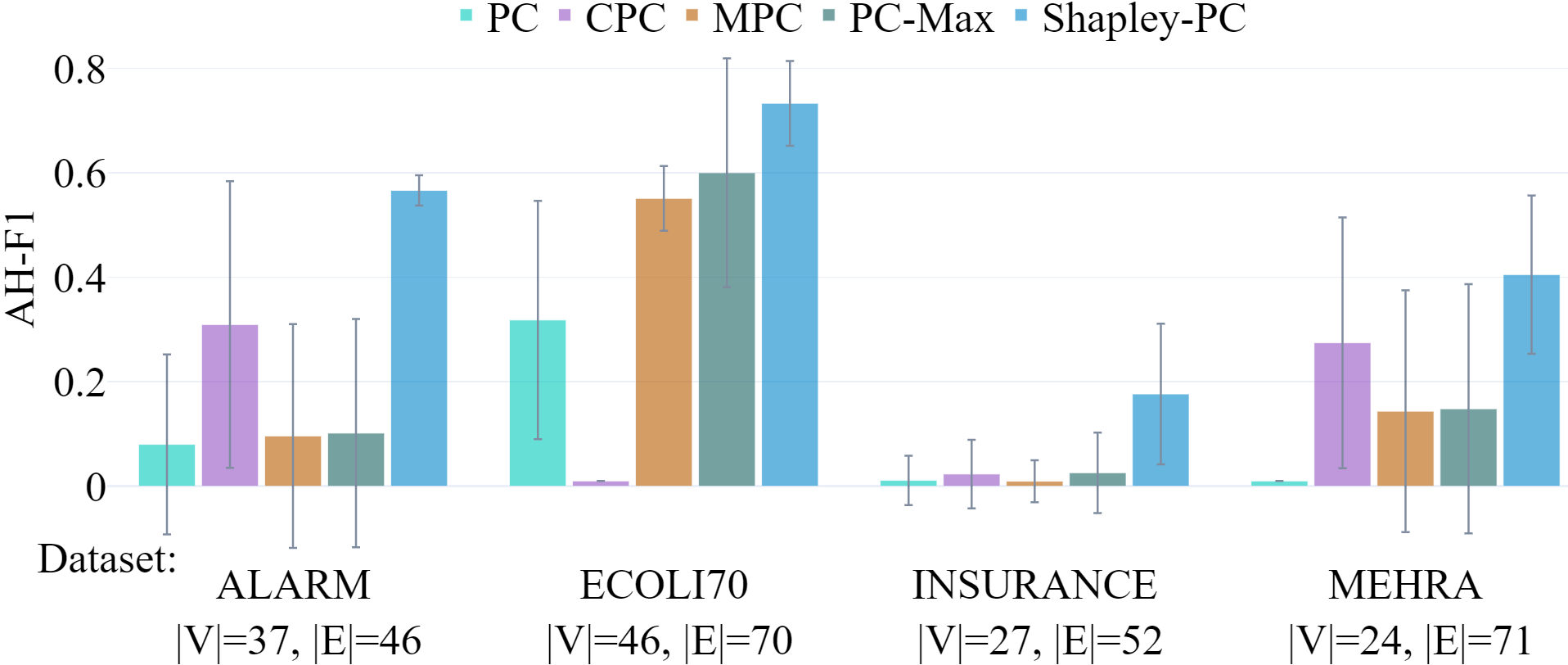}
    \caption{Mean and standard deviation ArrowHead F1 score for four datasets generated from pseudo-real Bayesian Networks from the \texttt{bnlearn} repository.}
    \label{fig:pseudoreal}
\end{figure}

\paragraph{Pseudo-Real Data}
In addition to the fully simulated data, we conduct experiments on datasets from the \href{https://www.bnlearn.com/bnrepository/}{\texttt{bnlearn}} repository.
The datasets are sampled from Bayesian Networks with fixed conditional probability tables, provided by previous studies and stored in the repository. We use datasets generated from all three categories available: discrete, Gaussian and Conditional Linear Gaussian Bayesian Networks. Alarm and Insurance are fully discrete, Ecoli70 fully continuous, while Mehra is mixed. We sample 50000 examples with 10 different seeds to measure performance and confidence intervals. 
Results for these data are shown in Figure~\ref{fig:pseudoreal} where we report the average ArrowHead F1 scores and their standard deviations. Shapley-PC ranks $1^{st}$ on all four datasets. However, it is significantly different from all other methods (according to a t-test, $\alpha=0.05$, see Appendix \S\ref{sec:real_det}) for Alarm and Insurance, while different from PC-Max on the Ecoli70 data only at the 0.1 significance level ($p=0.099$) and on par with CPC on the Mehra dataset. Details on the datasets, results for V-F1, as well as for additional datasets where significant differences were not observed, are in Appendix \ref{sec:real_det}.

\section{Conclusion and Future Work}
\label{sec:conclusion}
We proposed a decision rule for orienting v-structures in constraint-based CSL algorithms,  based on Shapley values, an established concept from game theory~\citep{shapley1953value}. We implemented our decision rule within the novel Shapley-PC algorithm and proved that it maintains the soundness, completeness and consistency guarantees of PC~\citep{spirtes2000causation}, that Shapley-PC is based on.
We carried out an extensive evaluation of Shapley-PC, showing that it outperforms its PC-based predecessors in orienting v-structures and more generally recovering causal directions when the data contains weak links driving orientation unfaithfulness \citep{ramsey2006CPC}.

Our proposed decision rule takes as input a skeleton $\SKEL{}$ to then analyse the strength of the associations between adjacent nodes and infer graph orientations. This procedure is directly transferable to constraint-based methods other than PC, possibly with less strict assumptions. One such algorithm is FCI~\citep{spirtes2000causation} which lifts both the sufficiency (no latent confounders) and the acyclicity assumptions~\citep{mooij2020fcicycle}. 

The applicability of our proposed method goes beyond constraint-based methods in that we can substitute $p$-values with any quantitative measure of association between variables. As shown in~\citep{ramsey2016improving},  in the context of the PC-Max algorithm, 
the scores underling score-based CSL methods such as GES~\citep{chickering2002learning} or FGS~\citep{ramsey2017million} can be used in the same guise. 
Additionally, hybrid methods combine independence tests and scores to estimate causal graphs. An example of such algorithms is MMHC~\citep{tsamardinos2006MMHC} that carries out a skeleton estimation before orienting edges using a score based on $p$-values. Our method could therefore also be easily extended to such methodologies.

Other directions for future work include the application of our decision rule to the skeleton estimation phase of constraint-based algorithms and to the Meek rules application. In fact, Meek rules can generate cyclic graphs~\citep{Tsagris2018variation} and, having to decide between arrows, one could use aggregated evidence in favour or against the orientation, from SIVs.
It would also be interesting to study more/less conservative versions
of our decision rule, to analyse the informativeness
thereof in interactive discovery processes involving humans, and, in line with~\citep{constantinou2021impact}, to compare it to the other categories of CSL methods in the literature.

\acks{The authors would like to thank Anna Rapberger, Ruben Menke, Antonio Rago, Avinash Kori, Torgunn Rings{\o} and all the anonymous referees for the helpful feedback in the build up to this work. Russo was supported by UK Research and Innovation (grant number EP/S023356/1), in the \href{www.safeandtrustedai.org}{UKRI Centre for Doctoral Training in Safe and Trusted Artificial Intelligence}. Toni was partially funded by the ERC under the EU’s Horizon 2020 research and innovation programme (grant agreement No. 101020934) and by J.P. Morgan and by the Royal Academy of Engineering under the Research Chairs and Senior Research Fellowships scheme. Any views or opinions expressed herein are solely those of the authors.}

\bibliography{main.bib}

\begin{thebibliography}{47}
\providecommand{\natexlab}[1]{#1}
\providecommand{\url}[1]{\texttt{#1}}
\expandafter\ifx\csname urlstyle\endcsname\relax
  \providecommand{\doi}[1]{doi: #1}\else
  \providecommand{\doi}{doi: \begingroup \urlstyle{rm}\Url}\fi

\bibitem[Abell{\'{a}}n et~al.(2006)Abell{\'{a}}n, G{\'{o}}mez{-}Olmedo, and Moral]{DBLP:conf/pgm/AbellanGM06}
Joaqu{\'{\i}}n Abell{\'{a}}n, Manuel G{\'{o}}mez{-}Olmedo, and Seraf{\'{\i}}n Moral.
\newblock Some variations on the {PC} algorithm.
\newblock In \emph{Third European Workshop on Probabilistic Graphical Models, 12-15 September 2006, Prague, Czech Republic. Electronic Proceedings}, pages 1--8, 2006.
\newblock URL \url{http://www.utia.cas.cz/files/mtr/pgm06/41\_paper.pdf}.

\bibitem[Barabási and Albert(1999)]{barabasi1999emergence}
Albert-László Barabási and Réka Albert.
\newblock Emergence of scaling in random networks.
\newblock \emph{Science}, 286\penalty0 (5439):\penalty0 509--512, 1999.
\newblock URL \url{http://www.doi.org/10.1126/science.286.5439.509}.

\bibitem[Budhathoki et~al.(2022)Budhathoki, Minorics, Bl{\"{o}}baum, and Janzing]{budhathoki2022causal}
Kailash Budhathoki, Lenon Minorics, Patrick Bl{\"{o}}baum, and Dominik Janzing.
\newblock Causal structure-based root cause analysis of outliers.
\newblock In \emph{International Conference on Machine Learning, {ICML} 2022, 17-23 July 2022, Baltimore, Maryland, {USA}}, volume 162 of \emph{Proceedings of Machine Learning Research}, pages 2357--2369. {PMLR}, 2022.
\newblock URL \url{https://proceedings.mlr.press/v162/budhathoki22a.html}.

\bibitem[Casella and Berger(2002)]{casella2002statistical}
G.~Casella and R.L. Berger.
\newblock \emph{Statistical Inference}.
\newblock Duxbury advanced series in statistics and decision sciences. Thomson Learning, 2002.
\newblock ISBN 9780534243128.
\newblock URL \url{https://books.google.de/books?id=0x_vAAAAMAAJ}.

\bibitem[Chen et~al.(2024)Chen, Qiao, Xie, Cai, Hao, and Zhang]{chen2024ARECI}
Zhengming Chen, Jie Qiao, Feng Xie, Ruichu Cai, Zhifeng Hao, and Keli Zhang.
\newblock Testing conditional independence between latent variables by independence residuals.
\newblock \emph{IEEE Transactions on Neural Networks and Learning Systems}, pages 1--13, 2024.
\newblock URL \url{https://doi.org/10.1109/TNNLS.2024.3368561}.

\bibitem[Chickering(2002)]{chickering2002learning}
David~Maxwell Chickering.
\newblock Learning equivalence classes of bayesian-network structures.
\newblock \emph{J. Mach. Learn. Res.}, 2:\penalty0 445--498, 2002.
\newblock URL \url{https://jmlr.org/papers/v2/chickering02a.html}.

\bibitem[Claassen and Heskes(2012)]{Claassen2012bayesian}
Tom Claassen and Tom Heskes.
\newblock {A Bayesian Approach to Constraint Based Causal Inference}.
\newblock In \emph{Proceedings of the Twenty-Eighth Conference on Uncertainty in Artificial Intelligence, Catalina Island, CA, USA, August 14-18, 2012}, pages 207--216. {AUAI} Press, 2012.
\newblock URL \url{https://doi.org/10.48550/arXiv.1210.4866}.

\bibitem[Colombo and Maathuis(2014)]{colombo2014order}
Diego Colombo and Marloes~H. Maathuis.
\newblock Order-independent constraint-based causal structure learning.
\newblock \emph{J. Mach. Learn. Res.}, 15\penalty0 (1):\penalty0 3741--3782, 2014.
\newblock URL \url{https://dl.acm.org/doi/10.5555/2627435.2750365}.

\bibitem[Colombo et~al.(2012)Colombo, Maathuis, Kalisch, and Richardson]{Colombo2012RFCI}
Diego Colombo, Marloes~H. Maathuis, Markus Kalisch, and Thomas~S. Richardson.
\newblock Learning high-dimensional directed acyclic graphs with latent and selection variables.
\newblock \emph{The Annals of Statistics}, 40\penalty0 (1):\penalty0 294--321, 2012.
\newblock ISSN 00905364, 21688966.
\newblock URL \url{http://www.jstor.org/stable/41713636}.

\bibitem[Constantinou et~al.(2023)Constantinou, Guo, and Kitson]{constantinou2021impact}
Anthony~C. Constantinou, Zhigao Guo, and Neville~Kenneth Kitson.
\newblock The impact of prior knowledge on causal structure learning.
\newblock \emph{Knowl. Inf. Syst.}, 65\penalty0 (8):\penalty0 3385--3434, 2023.
\newblock URL \url{https://doi.org/10.1007/s10115-023-01858-x}.

\bibitem[Dai et~al.(2023)Dai, Ding, Jiang, Han, and Zhang]{dai2023ml4c}
Haoyue Dai, Rui Ding, Yuanyuan Jiang, Shi Han, and Dongmei Zhang.
\newblock {ML4C:} seeing causality through latent vicinity.
\newblock In \emph{Proceedings of the 2023 {SIAM} International Conference on Data Mining, {SDM} 2023, Minneapolis-St. Paul Twin Cities, MN, USA, April 27-29, 2023}, pages 226--234. {SIAM}, 2023.
\newblock URL \url{https://doi.org/10.1137/1.9781611977653.ch26}.

\bibitem[Erdős and Rényi(1959)]{erdds1959random}
Paul Erdős and Alfréd Rényi.
\newblock On random graphs {I}.
\newblock \emph{Publ. math. debrecen}, 6\penalty0 (290-297):\penalty0 18, 1959.

\bibitem[Fisher(1970)]{fisher1970statistical}
Ronald~Aylmer Fisher.
\newblock Statistical methods for research workers.
\newblock In \emph{Breakthroughs in statistics: Methodology and distribution}, pages 66--70. Springer, 1970.

\bibitem[Frye et~al.(2020)Frye, Rowat, and Feige]{Frye2020asymshap}
Christopher Frye, Colin Rowat, and Ilya Feige.
\newblock Asymmetric shapley values: incorporating causal knowledge into model-agnostic explainability.
\newblock In \emph{Advances in Neural Information Processing Systems 33: Annual Conference on Neural Information Processing Systems 2020, NeurIPS 2020, December 6-12, 2020, virtual}, 2020.
\newblock URL \url{https://proceedings.neurips.cc/paper/2020/hash/0d770c496aa3da6d2c3f2bd19e7b9d6b-Abstract.html}.

\bibitem[Glymour et~al.(2019)Glymour, Zhang, and Spirtes]{glymour2019review}
Clark Glymour, Kun Zhang, and Peter Spirtes.
\newblock Review of causal discovery methods based on graphical models.
\newblock \emph{Front. Genet., 04 June 2019 Sec. Statistical Genetics and Methodology}, 10:\penalty0 524, 2019.
\newblock URL \url{https://doi.org/10.3389/fgene.2019.00524}.

\bibitem[Gretton et~al.(2007)Gretton, Fukumizu, Teo, Song, Sch{\"{o}}lkopf, and Smola]{Gretton2007hsic}
Arthur Gretton, Kenji Fukumizu, Choon~Hui Teo, Le~Song, Bernhard Sch{\"{o}}lkopf, and Alexander~J. Smola.
\newblock A kernel statistical test of independence.
\newblock In \emph{Advances in Neural Information Processing Systems 20, Proceedings of the Twenty-First Annual Conference on Neural Information Processing Systems, Vancouver, British Columbia, Canada, December 3-6, 2007}, pages 585--592. Curran Associates, Inc., 2007.
\newblock URL \url{https://proceedings.neurips.cc/paper/2007/hash/d5cfead94f5350c12c322b5b664544c1-Abstract.html}.

\bibitem[Harris and Drton(2013)]{harris2013consistent}
Naftali Harris and Mathias Drton.
\newblock {PC} algorithm for nonparanormal graphical models.
\newblock \emph{J. Mach. Learn. Res.}, 14\penalty0 (1):\penalty0 3365--3383, 2013.
\newblock URL \url{https://dl.acm.org/doi/10.5555/2567709.2567770}.

\bibitem[Heskes et~al.(2020)Heskes, Sijben, Bucur, and Claassen]{heskes2020causalshap}
Tom Heskes, Evi Sijben, Ioan~Gabriel Bucur, and Tom Claassen.
\newblock Causal shapley values: Exploiting causal knowledge to explain individual predictions of complex models.
\newblock In \emph{Advances in Neural Information Processing Systems 33: Annual Conference on Neural Information Processing Systems 2020, NeurIPS 2020, December 6-12, 2020, virtual}, 2020.
\newblock URL \url{https://proceedings.neurips.cc/paper/2020/hash/32e54441e6382a7fbacbbbaf3c450059-Abstract.html}.

\bibitem[Hung et~al.(1997)Hung, O'Neill, Bauer, and Kohne]{hung97pvalue}
H.~M.~James Hung, Robert~T. O'Neill, Peter Bauer, and Karl Kohne.
\newblock The behavior of the p-value when the alternative hypothesis is true.
\newblock \emph{Biometrics}, 53\penalty0 (1):\penalty0 11--22, 1997.
\newblock ISSN 0006341X, 15410420.
\newblock URL \url{http://www.jstor.org/stable/2533093}.

\bibitem[Ichiishi(1983)]{ichiishi1983game}
T.~Ichiishi.
\newblock \emph{Game Theory for Economic Analysis}.
\newblock Economic Theory, Econometrics, and Mathematical Economics. Elsevier Science, 1983.
\newblock ISBN 9780123701800.
\newblock URL \url{https://books.google.co.uk/books?id=zFm7AAAAIAAJ}.

\bibitem[Kalisch and B{\"{u}}hlmann(2007)]{kalisch2007estimating}
Markus Kalisch and Peter B{\"{u}}hlmann.
\newblock Estimating high-dimensional directed acyclic graphs with the pc-algorithm.
\newblock \emph{J. Mach. Learn. Res.}, 8:\penalty0 613--636, 2007.
\newblock URL \url{https://dl.acm.org/doi/10.5555/1314498.1314520}.

\bibitem[Lundberg and Lee(2017)]{lundberg2017shap}
Scott~M. Lundberg and Su{-}In Lee.
\newblock A unified approach to interpreting model predictions.
\newblock In \emph{Advances in Neural Information Processing Systems 30: Annual Conference on Neural Information Processing Systems 2017, December 4-9, 2017, Long Beach, CA, {USA}}, pages 4765--4774, 2017.
\newblock URL \url{https://proceedings.neurips.cc/paper/2017/hash/8a20a8621978632d76c43dfd28b67767-Abstract.html}.

\bibitem[Meek(1995)]{Meek1995Orient}
Christopher Meek.
\newblock Causal inference and causal explanation with background knowledge.
\newblock In \emph{{UAI} '95: Proceedings of the Eleventh Annual Conference on Uncertainty in Artificial Intelligence, Montreal, Quebec, Canada, August 18-20, 1995}, pages 403--410. Morgan Kaufmann, 1995.
\newblock URL \url{https://doi.org/10.48550/arXiv.1302.4972}.

\bibitem[Mooij and Claassen(2020)]{mooij2020fcicycle}
Joris~M. Mooij and Tom Claassen.
\newblock Constraint-based causal discovery using partial ancestral graphs in the presence of cycles.
\newblock In \emph{Proceedings of the Thirty-Sixth Conference on Uncertainty in Artificial Intelligence, {UAI} 2020, virtual online, August 3-6, 2020}, volume 124 of \emph{Proceedings of Machine Learning Research}, pages 1159--1168. {AUAI} Press, 2020.
\newblock URL \url{http://proceedings.mlr.press/v124/m-mooij20a.html}.

\bibitem[Pearl(2009)]{pearl2009causality}
Judea Pearl.
\newblock \emph{Causality}.
\newblock Cambridge University Press, 2 edition, 2009.
\newblock URL \url{https://doi.org/10.1017/CBO9780511803161}.

\bibitem[Pearson(1900)]{pearson1900x2}
Karl Pearson.
\newblock X. on the criterion that a given system of deviations from the probable in the case of a correlated system of variables is such that it can be reasonably supposed to have arisen from random sampling.
\newblock \emph{Philosophical Magazine Series 5}, 50\penalty0 (302):\penalty0 157--175, 1900.

\bibitem[Peters et~al.(2017)Peters, Janzing, and Sch\"olkopf]{peters2017elements}
J.~Peters, D.~Janzing, and B.~Sch\"olkopf.
\newblock \emph{Elements of Causal Inference: Foundations and Learning Algorithms}.
\newblock MIT Press, Cambridge, MA, USA, 2017.
\newblock URL \url{https://mitpress.mit.edu/9780262037310/elements-of-causal-inference/}.

\bibitem[Peters and B{\"{u}}hlmann(2015)]{peters2015structural}
Jonas Peters and Peter B{\"{u}}hlmann.
\newblock Structural intervention distance for evaluating causal graphs.
\newblock \emph{Neural Comput.}, 27\penalty0 (3):\penalty0 771--799, 2015.
\newblock URL \url{https://doi.org/10.1162/NECO\_a\_00708}.

\bibitem[Raghu et~al.(2018)Raghu, Ramsey, Morris, Manatakis, Spirtes, Chrysanthis, Glymour, and Benos]{raghu_comparison_2018}
Vineet~K. Raghu, Joseph~D. Ramsey, Alison Morris, Dimitrios~V. Manatakis, Peter Spirtes, Panos~K. Chrysanthis, Clark Glymour, and Panayiotis~V. Benos.
\newblock Comparison of strategies for scalable causal discovery of latent variable models from mixed data.
\newblock \emph{Int. J. Data Sci. Anal.}, 6\penalty0 (1):\penalty0 33--45, 2018.
\newblock URL \url{https://doi.org/10.1007/s41060-018-0104-3}.

\bibitem[Ramsey(2016)]{ramsey2016improving}
Joseph Ramsey.
\newblock Improving accuracy and scalability of the pc algorithm by maximizing p-value.
\newblock \emph{CoRR abs/1610.00378}, 2016.
\newblock URL \url{https://doi.org/10.48550/arXiv.1610.00378}.

\bibitem[Ramsey et~al.(2006)Ramsey, Zhang, and Spirtes]{ramsey2006CPC}
Joseph~D. Ramsey, Jiji Zhang, and Peter Spirtes.
\newblock Adjacency-faithfulness and conservative causal inference.
\newblock In \emph{{UAI} '06, Proceedings of the 22nd Conference in Uncertainty in Artificial Intelligence, Cambridge, MA, USA, July 13-16, 2006}. {AUAI} Press, 2006.
\newblock URL \url{https://www.cmu.edu/dietrich/philosophy/docs/spirtes/uai06.pdf}.

\bibitem[Ramsey et~al.(2017)Ramsey, Glymour, Sanchez{-}Romero, and Glymour]{ramsey2017million}
Joseph~D. Ramsey, Madelyn Glymour, Ruben Sanchez{-}Romero, and Clark Glymour.
\newblock A million variables and more: the fast greedy equivalence search algorithm for learning high-dimensional graphical causal models, with an application to functional magnetic resonance images.
\newblock \emph{Int. J. Data Sci. Anal.}, 3\penalty0 (2):\penalty0 121--129, 2017.
\newblock URL \url{https://doi.org/10.1007/s41060-016-0032-z}.

\bibitem[Richardson and Spirtes(1999)]{Richardson1996CCD}
Thomas Richardson and Peter Spirtes.
\newblock Automated discovery of linear feedback models.
\newblock In \emph{Computation, Causation, and Discovery}. AAAI Press, 05 1999.
\newblock ISBN 9780262315821.
\newblock URL \url{https://doi.org/10.7551/mitpress/2006.003.0010}.

\bibitem[Robins et~al.(2003)Robins, Scheines, Spirtes, and Wasserman]{Robins2003consistency}
James Robins, Richard Scheines, Peter Spirtes, and Larry Wasserman.
\newblock Uniform consistency in causal inference.
\newblock \emph{Biometrika}, 90, 09 2003.
\newblock URL \url{https://doi.org/10.1093/biomet/90.3.491}.

\bibitem[Sch{\"{o}}lkopf et~al.(2021)Sch{\"{o}}lkopf, Locatello, Bauer, Ke, Kalchbrenner, Goyal, and Bengio]{scholkopf2021toward}
Bernhard Sch{\"{o}}lkopf, Francesco Locatello, Stefan Bauer, Nan~Rosemary Ke, Nal Kalchbrenner, Anirudh Goyal, and Yoshua Bengio.
\newblock Toward causal representation learning.
\newblock \emph{Proc. {IEEE}}, 109\penalty0 (5):\penalty0 612--634, 2021.
\newblock URL \url{https://doi.org/10.1109/JPROC.2021.3058954}.

\bibitem[Shapley(1953)]{shapley1953value}
Lloyd~S Shapley.
\newblock A value for n-person games (1953).
\newblock \emph{Contribution to the Theory of Games}, 1953.
\newblock URL \url{https://www.rand.org/content/dam/rand/pubs/papers/2021/P295.pdf}.

\bibitem[Spirtes et~al.(2000)Spirtes, Glymour, and Scheines]{spirtes2000causation}
Peter Spirtes, Clark Glymour, and Richard Scheines.
\newblock \emph{Causation, Prediction, and Search, Second Edition}.
\newblock Adaptive computation and machine learning. {MIT} Press, 2000.
\newblock ISBN 978-0-262-19440-2.
\newblock URL \url{https://www.cs.cmu.edu/afs/andrew/scs/cs/15-381/archive/OldFiles/lib/cvsub/.g/group/sdss/.g/group2/g/scottd/fullbook.pdf}.

\bibitem[Teneggi et~al.(2023)Teneggi, Bharti, Romano, and Sulam]{teneggi2023shapxrt}
Jacopo Teneggi, Beepul Bharti, Yaniv Romano, and Jeremias Sulam.
\newblock {SHAP-XRT:} the shapley value meets conditional independence testing.
\newblock \emph{Trans. Mach. Learn. Res.}, 2023, 2023.
\newblock URL \url{https://openreview.net/forum?id=WFtTpQ47A7}.

\bibitem[Tsagris(2019)]{Tsagris2018variation}
Michail Tsagris.
\newblock Bayesian network learning with the {PC} algorithm: An improved and correct variation.
\newblock \emph{Appl. Artif. Intell.}, 33\penalty0 (2):\penalty0 101--123, 2019.
\newblock URL \url{https://doi.org/10.1080/08839514.2018.1526760}.

\bibitem[Tsamardinos et~al.(2006)Tsamardinos, Brown, and Aliferis]{tsamardinos2006MMHC}
Ioannis Tsamardinos, Laura~E. Brown, and Constantin~F. Aliferis.
\newblock The max-min hill-climbing bayesian network structure learning algorithm.
\newblock \emph{Mach. Learn.}, 65\penalty0 (1):\penalty0 31--78, 2006.
\newblock URL \url{https://doi.org/10.1007/s10994-006-6889-7}.

\bibitem[Vowels et~al.(2022)Vowels, Camgoz, and Bowden]{vowels2022d}
Matthew~J. Vowels, Necati~Cihan Camgoz, and Richard Bowden.
\newblock D’ya {Like} {DAGs}? {A} {Survey} on {Structure} {Learning} and {Causal} {Discovery}.
\newblock \emph{ACM Comput. Surv.}, 55\penalty0 (4), November 2022.
\newblock ISSN 0360-0300.
\newblock URL \url{https://doi.org/10.1145/3527154}.

\bibitem[Young(1985)]{young1985monotonic}
H.~Peyton Young.
\newblock Monotonic solutions of cooperative games.
\newblock \emph{International Journal of Game Theory}, 14\penalty0 (2):\penalty0 65--72, 1985.
\newblock URL \url{https://doi.org/10.1007/BF01769885}.

\bibitem[Zanga et~al.(2022)Zanga, Ozkirimli, and Stella]{ZANGA2022survey}
Alessio Zanga, Elif Ozkirimli, and Fabio Stella.
\newblock A survey on causal discovery: Theory and practice.
\newblock \emph{Int. J. Approx. Reason.}, 151:\penalty0 101--129, 2022.
\newblock URL \url{https://doi.org/10.1016/j.ijar.2022.09.004}.

\bibitem[Zhang et~al.(2023)Zhang, Xia, Zhang, Zhou, and Guan]{Zhang23SCIT}
Hao Zhang, Yewei Xia, Kun Zhang, Shuigeng Zhou, and Jihong Guan.
\newblock Conditional independence test based on residual similarity.
\newblock \emph{{ACM} Trans. Knowl. Discov. Data}, 17\penalty0 (8):\penalty0 117:1--117:18, 2023.
\newblock URL \url{https://doi.org/10.1145/3593810}.

\bibitem[Zhang and Spirtes(2003)]{ZhangSpirtes03strong}
Jiji Zhang and Peter Spirtes.
\newblock Strong faithfulness and uniform consistency in causal inference.
\newblock In \emph{{UAI} '03, Proceedings of the 19th Conference in Uncertainty in Artificial Intelligence, Acapulco, Mexico, August 7-10 2003}, pages 632--639. Morgan Kaufmann, 2003.
\newblock URL \url{https://dl.acm.org/doi/pdf/10.5555/2100584.2100661}.

\bibitem[Zhang et~al.(2011)Zhang, Peters, Janzing, and Sch{\"{o}}lkopf]{zhang2011kci}
Kun Zhang, Jonas Peters, Dominik Janzing, and Bernhard Sch{\"{o}}lkopf.
\newblock Kernel-based conditional independence test and application in causal discovery.
\newblock In \emph{{UAI} 2011, Proceedings of the Twenty-Seventh Conference on Uncertainty in Artificial Intelligence, Barcelona, Spain, July 14-17, 2011}, pages 804--813. {AUAI} Press, 2011.
\newblock URL \url{https://doi.org/10.48550/arXiv.1202.3775}.

\bibitem[Zheng et~al.(2024)Zheng, Huang, Chen, Ramsey, Gong, Cai, Shimizu, Spirtes, and Zhang]{zheng2024causal}
Yujia Zheng, Biwei Huang, Wei Chen, Joseph Ramsey, Mingming Gong, Ruichu Cai, Shohei Shimizu, Peter Spirtes, and Kun Zhang.
\newblock Causal-learn: Causal discovery in python.
\newblock \emph{Journal of Machine Learning Research}, 25\penalty0 (60):\penalty0 1--8, 2024.
\newblock URL \url{https://www.jmlr.org/papers/volume25/23-0970/23-0970.pdf}.

\end{thebibliography}

\clearpage
\appendix


\setcounter{section}{0}
\renewcommand{\thesection}{\Alph{section}}
\setcounter{theorem}{1}    
\section{Proofs}
\label{sec:proofs}

\begin{lemma}
    Given a skeleton \(\SKEL\), a UT \( X_i-X_j-X_k \in \SKEL \), \( X_i, X_j, X_k \in \nodeSet \), and a perfect CIT \( I_{\infty} \), the SIV of variable \( X_j \) \(\phi_{I_{\infty}}(X_j, \{X_i, X_k\}) < 0\) if and only if \( X_j \) is a collider for \( X_i \) and \( X_k \).
\end{lemma}

\begin{proof}
    Assume a perfect CIT \( I_{\infty} \) (Def.~\ref{def:perfectCIT}). The marginal contribution of \( X_j \in \nodeSet \) for a conditioning set \( \condSet \in \mathbf{C} \) (Eq.~\ref{eq:adjsets}) is:
    \(
    \Delta_{X_j}(\condSet) = I_{\infty}(X_i, X_k \given \condSet \cup \{X_j\}) - I_{\infty}(X_i, X_k \given \condSet).
    \)
    \paragraph{($\Rightarrow$)} Assume $X_j$ is indeed a collider for $X_i, X_k$ and consider the cases based on whether \(\condSet\) blocks paths other than the one passing through \( X_j \):

    \noindent \underline{Case 1 (\(\condSet\) blocks all other paths)}
    
    Conditioning on \( X_j \) opens a previously blocked path, inducing dependence:
    \[
    I_{\infty}(X_i, X_k \given \condSet) = 1, \quad I_{\infty}(X_i, X_k \given \condSet \cup \{X_j\}) = 0, \quad \Delta_{X_j}(\condSet) = -1.
    \]
    \underline{Case 2 (\(\condSet\) does not block all other paths)}
    
    Conditioning on \( X_j \) opens an additional path, not altering dependence:
        \[
        I_{\infty}(X_i, X_k \given \condSet) = I_{\infty}(X_i, X_k \given \condSet \cup \{X_j\}) = 0, \quad \Delta_{X_j}(\condSet) = 0.
        \]
    Given the UT configuration, $X_i \notin \text{adj}(X_k)$, and adjacency faithfulness ensures that at least one \(\condSet\) blocks all paths. Hence:
    \[
    \phi_{I_{\infty}}(X_j, \{X_i, X_k\}) = \sum_{\condSet \in \mathbf{C}} w_{\condSet}^n \Delta_{X_j}(\condSet) < 0.
    \]

    \paragraph{($\Leftarrow$)} Assume, instead, that $X_j$ is not a collider for $X_i, X_k$, and again by case distinction:

    \noindent \underline{Case 1 (\(\condSet\) blocks all other paths)}
    
    Conditioning on \( X_j \) blocks the only open path, inducing independence:
    \[
    I_{\infty}(X_i, X_k \given \condSet) = 0, \quad I_{\infty}(X_i, X_k \given \condSet \cup \{X_j\}) = 1, \quad \Delta_{X_j}(\condSet) = 1.
    \]
    \underline{Case 2 (\(\condSet\) does not block all other paths)}
    
    Conditioning on \( X_j \) blocks one of the open paths, but does not alter dependence:
    \[
    I_{\infty}(X_i, X_k \given \condSet) = I_{\infty}(X_i, X_k \given \condSet \cup \{X_j\}) = 0, \quad \Delta_{X_j}(\condSet) = 0.
    \]
    By adjacency faithfulness:
    \[
    \phi_{I_{\infty}}(X_j, \{X_i, X_k\}) = \sum_{\condSet \in \mathbf{C}} w_{\condSet}^n \Delta_{X_j}(\condSet) > 0.
    \]
    Therefore, \(\phi_{I_{\infty}}(X_j, \{X_i, X_k\}) < 0\) if and only if \( X_j \) is a collider for \( X_i \) and \( X_k \).
\end{proof}

\begin{theorem}
    Let $\mathbf{P}(\nodeSet)$ be a joint distribution faithful to a DAG $\DAG = (\nodeSet, \edgSet)$, and assume access to perfect conditional independence information for all pairs $(X_i, X_j) \in \nodeSet$ given subsets $\condSet \subseteq \nodeSet \setminus \{X_i, X_j\}$. Then the output of Shapley-PC is the CPDAG representing the MEC of $\DAG$.
\end{theorem}

\begin{proof}
    The proof follows straightforwardly from Lemma~\ref{th:shap_sign} and two additional results in the literature. Assume faithfulness and perfect CITs, then:
    \begin{itemize}
        \item \textbf{Step 1}: The skeleton is guaranteed to be correct~\citep[Thm. 2]{colombo2014order}.
        \item \textbf{Step 2}: Given a correct skeleton, by Lemma~\ref{th:shap_sign}, $\phi_I(X_j, \{X_i,X_k\}) < 0$ correctly identifies colliders in v-structures.
        \item \textbf{Step 3}: Given a PDAG, Meek's rules, are sound and complete~\citep[Thm. 2 and 3]{Meek1995Orient}. 
    \end{itemize}
    Thus, Shapley-PC outputs the correct CPDAG.
\end{proof}

\section{Details on Experiments}

\label{sec:exp_det}
In this section we provide additional details for the experiments in \S\ref{sec:experiments} of the main text.
\subsection{Baselines}
\label{sec:baseline_det}
We used the following four baselines with respective implementations (see \S\ref{sec:background} and \S\ref{sec:related} for context):
\begin{itemize}
    \item PC-Stable\footnote{Implemented in \href{https://github.com/py-why/causal-learn}{causal-learn}}~\citep{colombo2014order} consists of three steps: 
    \begin{enumerate}
        \item building a skeleton of the graph via adjacency search: conditional independence tests are performed for each pair of variables in the data. For efficiency, the algorithm starts by performing marginal independence tests (empty conditioning set) and gradually increases the size of the conditioning set once all pairs of variables have been tested. If an independence is found for a pair of variables, the edge is removed after all variables have been tested for that conditioning set size. The separating set is stored for the pair of variables found independent.\footnote{This is the difference of PC-Stable with the original PC~\citep{spirtes2000causation}, that instead removes edges as soon as an independence is found, being then subject to the order in which the variables are tested.} This step outputs a skeleton \SKEL.
        \item for each unshielded triple (UT) in the skeleton $X_i-X_j-X_k\in \SKEL$ output of step 1, the UT is oriented as a v-structure $X_i\rightarrow X_j\leftarrow X_k$ if $X_j$ is in the separating set for variables $X_i, X_k$. 
        \item all triangles (groups of three adjacent variables) and kites (group of four adjacent variables) are analysed with the rules for patterns~\citep{Meek1995Orient}. If certain configurations are obtained in step 2, further orientations are performed on the remaining undirected edges. The application of these rules returns a sound and complete CPDAG that represent the true DAG~\citep{Meek1995Orient}.
    \end{enumerate}
    \item Conservative-PC (CPC)\footnote{Our \href{https://github.com/briziorusso/ShapleyPC}{implementation}
    , based on \href{https://github.com/py-why/causal-learn}{causal-learn}}~\citep{ramsey2006CPC} is a modification of the PC(-Stable) algorithm. Step 1 can be the original or the stable version and 3 is the same, while the v-structure orientation rule is the main proposal. \cite{ramsey2006CPC} break up the faithfulness assumption into adjacency-faithfulness and orientation-faithfulness. Assuming the former (i.e. that the edges are correctly identified) CPC orients v-structures by checking that the latter assumption is satisfied in the data: for a UT $X_i-X_j-X_k$, $X_j$ is deemed a collider only if it is found in none of the separating sets for $X_i,X_k$. 
    
    \item Majority-PC (MPC)\footnote{Our \href{https://github.com/briziorusso/ShapleyPC}{implementation}
    , based on \href{https://github.com/py-why/causal-learn}{causal-learn}}~\citep{colombo2014order} relaxes the orientation-faithfulness check of CPC and orients v-structures if the potential collider appears in less than half of the separating sets of the other two nodes.

    \item PC-Max\footnote{Implemented in \href{https://github.com/py-why/causal-learn}{causal-learn}}~\citep{ramsey2016improving} again only modifies Step 2 of PC(-Stable). It selects the CIT with the maximum $p$-value for a given UT and only orients it as a v-structure if the conditioning set for the selected CIT does not contain the variable under consideration. 
\end{itemize}

\subsection{Implementation}
\label{sec:implem_det}
We provide an implementation of Shapley-PC based on the \texttt{causal-learn} python package~\citep{zheng2024causal}. Within \texttt{causal-learn}, we define a new PC function that accommodates our decision rule. The code is available at the following repository: 
\url{https://github.com/briziorusso/ShapleyPC}
In the repository, we also made available the code to reproduce all experiments and we saved all the plots, presented herein and in the main text, in HTML format. Downloading and opening them in a browser allows the inspection of all the numbers behind the plots in an interactive way. 

\paragraph{Hyperparameters} For all the methods we used Fisher's Z test~\citep{fisher1970statistical}, as implemented in \texttt{causal-learn}, with significance threshold $\alpha=0.01$ for the \texttt{bnlearn} dataset and with decreasing $\alpha$ for increasing number of nodes in the fully synthetic simulations: we used $\alpha=0.1, 0.05, 0.01$ for number of nodes $|\nodeSet|=10, 20, 50$, respectively.

\paragraph{Computing infrastructure} 
All experiments were ran on Intel(R) Xeon(R) w5-2455X CPU with 4600 max MHz and 128GB of RAM. We used python 3.10.12 on Ubuntu 22.04.

\subsection{Evaluation Metrics}
\label{sec:met_det}
In line with \citep{ramsey2006CPC, ramsey2016improving}, we analyse the ability to identify v-structures (colliders), which is the focus of our proposed algorithm, alongside the overall performance in recovering the causal arrows of the true graph. All the metrics are calculated on the output CPDAGs hence the (binary) adjacency matrices can have entries for both $(X_i,X_j)$ and $(X_j,X_i)$, in which case the edge is undirected.

For the accuracy in classifying v-structures, we use precision and recall in classifying correctly identified UTs and summarise it with the F1 Score. Specifically:
\begin{itemize}
    \item V-Precision = V-TP/(V-TP + V-FP)
    \item V-Recall = V-TP/(V-TP + V-FN)    
    \item V-F1 Score = $2\times (\text{V-P} \times \text{V-R}) / (\text{V-P}+\text{V-R})$
\end{itemize}
where V-True Positive (V-TP) is the number of correctly estimated v-structures; V-False Positive (V-FP) is the number of UTs wrongly deemed as v-structures; V-False Negative (V-FN) is the number of v-structures not deemed as such.

Also to evaluate the accuracy in identifying causal directions in the true graph we use F1, but calculated on the number of (in)correct arrowheads (AH-F1) as follows:
\begin{itemize}
    \item AH-Precision = AH-TP/(AH-TP + AH-FP)
    \item AH-Recall = AH-TP/(AH-TP + AH-FN)    
    \item AH-F1 Score = $2\times (\text{AH-P} \times \text{AH-R}) / (\text{AH-P}+\text{AH-R})$
\end{itemize}
where AH-True Positive (AH-TP) is the number of estimated edges with correct direction; False Positive (AH-FP) is the number of extra arrowheads; False Negative (AH-FN) is the number of missing arrowheads.

In addition to the metrics focusing on orientations and v-structures, we report two other commonly used metrics in CSL (see e.g.~\cite{constantinou2021impact}): Structural Hamming Distance (SHD)~\citep{tsamardinos2006MMHC} and Structural Intervention Distance (SID)~\citep{peters2015structural}.

SHD  = E + M + R, where Extra (E) is the set of extra edges, Missing (M) are the ones missing from the skeleton of the estimated graph and Reversed (R) have incorrect direction.

SID quantifies the agreement to a causal graph in terms of interventional distributions. It aims at quantifying the incorrect causal inference estimations stemming out of a mistake in the causal graph estimation, akin to a downstream task error on a pre-processing step where the task is causal inference and the pre-processing step is finding the right graph to inform it. Both missing/extra edges and incorrect orientation will play a role in the incorrect causal inferences.

\subsection{Synthetic Data}
\label{sec:synth_det}

Here we provide detail for the simulation study presented in \S\ref{sec:experiments}, in particular Table~\ref{tab:main_res} and \ref{tab:runtime}.

\subsubsection{DGP Details}
\label{sec:dgp_sim_det}

In each experiment, we first generate 10 random graphs with maximum degree of 10 for each combination of three parameters: 
\begin{itemize}
    \item graph type: Erd\"{o}s-Rényi (ER \citep{erdds1959random}) and Scale Free (SF \citep{barabasi1999emergence});
    \item number of nodes: $|\nodeSet| \in \{10, 20, 50\}$;
    \item density: $d=\{1, 2, 4\}$, with $|\edgSet|=|\nodeSet|\times d$.    
\end{itemize}

Given the ground truth DAGs $\DAG$, we simulate Structural Equation Models (SEMs) belonging to the Additive Noise Model, formally:
\begin{equation}
    X_j = f_j(\text{pa}(\DAG,X_j)) + \noise_j \hspace{0.2cm} \, \forall \, j \in [1,\ldots,|\nodeSet|]
\end{equation}

where $f_j$ is linear function with coefficients $\Adjmat$ and $\noise_j$ are samples from a noise distribution.

The coefficients $\Adjmat$ are sampled from a uniform distribution with parameters $[-1.5,-0.5]\cup[0.5,1.5]$ for $95\%$ of the effects, and $[-0.001,0.001]$ for the remaining $5\%$. Sampling effect magnitudes from the range $[-0.001,0.001]$ simulates the presence of weak edges to induce violations of orientation-faithfulness (see \S\ref{sec:experiments} and \citep{ramsey2006CPC}).

Finally, we sample $\feats = \Adjmat^T \feats + \noise$ where the noise \noise{} is generated from the following four distributions:
\begin{multicols}{2}
        \begin{itemize}
            \item Gaussian: $\noise \sim \mathcal{N}(0,1)$
            \item Exponential: $\noise \sim E(1)$
        \end{itemize}
    \columnbreak
        \begin{itemize}
            \item Gumbel: $\noise \sim G(0,1)$
            \item Uniform: $\noise \sim U(-1,1)$
        \end{itemize}
\end{multicols}

We vary the number of drawn samples ($N$) in function of the number of nodes $N=s\times|\nodeSet|$, $s \in \{100,500,1000\}$ and refer to $s$ as the proportional sample size. After sampling from the described DGPs we standardise the data using the standard scaler from \texttt{sklearn}.\footnote{\url{https://scikit-learn.org/stable/modules/generated/sklearn.preprocessing.StandardScaler.html}} Code to reproduce the simulated data is provided in our repository.

\subsubsection{Statistical Tests}\label{sec:stats}
Here we provide details for the statistical tests used to measure the significance of the difference in the results presented in Table~\ref{tab:main_res} in the main text. In Tables~\ref{tab:tests_ER_10}, \ref{tab:tests_ER_50}, \ref{tab:tests_SF_10} and \ref{tab:tests_SF_50} we provide t-statistics and $p$-values for graphs ER and SF graphs of 10 and 50 nodes, respectively. 

In each table we present pairwise comparisons of means, for V-F1 and AH-F1 scores presented in Table~\ref{tab:main_res} of the main text. We use two-sample, unequal variance t-tests, with degrees of freedom of 39 (10 seeds and 4 noise distributions, minus 1).

\subsubsection{Additional Results}
\label{sec:synth_plt_det}

Here we provide additional results that were not presented in the main text for space constraints. The results corroborate the ones presented in the main text.

\begin{table*}[ht]
    \centering
    \caption[AH-F1 and V-F1 for graphs of 10 and 50 nodes, $d=1$]{AH-F1 and V-F1 Scores $\pm$ std for ER1 and SF1 graphs of nodes $\lvert\nodeSet\lvert \in \{10,50\}$. No significant differences according to a t-test, $\alpha=0.05$.}
    \begin{tabular}{c|r|cc|cc}
    \toprule
    &\textbf{Method} & \multicolumn{2}{c|}{\textbf{ER1}} & \multicolumn{2}{c}{\textbf{SF1}} \\
    & & \textbf{$|\nodeSet|=10$} & \textbf{$|\nodeSet|=50$} & \textbf{$|\nodeSet|=10$} & \textbf{$|\nodeSet|=50$} \\
    \midrule
    \multirow{5}{*}{\textbf{\rotatebox{90}{\textbf{AH-F1}}}}
    & PC-Stable   & 0.91$\pm$0.14 & 0.82$\pm$0.14 & 0.9$\pm$0.12  & 0.91$\pm$0.07 \\
    & CPC         & 0.97$\pm$0.08 & 0.87$\pm$0.11 & 0.99$\pm$0.03 & 1.0$\pm$0.01 \\
    & MPC         & 0.96$\pm$0.09 & 0.87$\pm$0.11 & 0.98$\pm$0.05 & 0.99$\pm$0.02 \\
    & PC-Max      & 0.96$\pm$0.09 & 0.87$\pm$0.11 & 0.95$\pm$0.08 & 0.96$\pm$0.04 \\
    & Shapley-PC  & 0.93$\pm$0.12 & 0.9$\pm$0.07  & 0.95$\pm$0.08 & 0.96$\pm$0.04 \\
    \midrule
    \multirow{5}{*}{\textbf{\rotatebox{90}{\textbf{V-F1}}}}
    & PC-Stable   & 0.97$\pm$0.09 & 0.92$\pm$0.16 & 0.99$\pm$0.03 & 0.98$\pm$0.04 \\
    & CPC         & 1.0$\pm$0.0   & 0.93$\pm$0.13 & 1.0$\pm$0.0   & 1.0$\pm$0.0 \\
    & MPC         & 1.0$\pm$0.0   & 0.93$\pm$0.13 & 1.0$\pm$0.01  & 1.0$\pm$0.0 \\
    & PC-Max      & 1.0$\pm$0.0   & 0.97$\pm$0.08 & 1.0$\pm$0.0   & 1.0$\pm$0.0 \\
    & Shapley-PC  & 0.99$\pm$0.03 & 1.0$\pm$0.02  & 1.0$\pm$0.0   & 1.0$\pm$0.0 \\
    \bottomrule
    \end{tabular}
    \label{tab:sparse_tab}
\end{table*}
\paragraph{Sparsest Graphs (d=1)}
The results presented in Table~\ref{tab:main_res} in the main text show AH-F1 and V-F1 for ER and SF graphs of density $d=\{2, 4\}$. Here we complete the picture and provide results for the sparsest graphs analysed: $d=1$. We can see from Table~\ref{tab:sparse_tab}, that all methods perform quite well of very sparse graphs. This result is in line with~\citep{kalisch2007estimating}. Given the limited opportunity for improvement, no significant differences between the various methods is observed.

\paragraph{Graphs of 20 Nodes}

The results presented in Table~\ref{tab:main_res} in the main text show AH-F1 and V-F1 for ER and SF graphs of 10 anf 50 nodes. Here we complete the picture and provide results for graphs of 20 nodes. The results corroborate the ones presented in the main paper. From Table~\ref{tab:20_nodes}, we can see that Shapley-PC outperforms all other methods on ER4, SF2 and SF4. On ER2 it is not significantly different from PC-Max (according to a t-test, $\alpha=0.05$), but better than all other methods.

\begin{table*}[ht]
    \centering
    \caption{AH-F1 and V-F1 Scores $\pm$ std for ER2, ER4, SF2, and SF4 graphs of 20 nodes. Bold if significantly different from the runner-up (according to a t-test, $\alpha=0.05$).}
    \begin{tabular}{c|r|cc|cc}
    \toprule
    & \textbf{Method} & \textbf{ER2} & \textbf{ER4} & \textbf{SF2} & \textbf{SF4} \\
    \midrule
    \multirow{5}{*}{\textbf{\rotatebox{90}{\textbf{AH-F1}}}}
    & PC-Stable   & 0.28$\pm$0.26         & 0.08$\pm$0.10         & 0.59$\pm$0.30         & 0.22$\pm$0.25 \\
    & CPC         & 0.37$\pm$0.25         & 0.11$\pm$0.11         & 0.52$\pm$0.35         & 0.25$\pm$0.27 \\
    & MPC         & 0.31$\pm$0.26         & 0.11$\pm$0.11         & 0.50$\pm$0.36         & 0.25$\pm$0.26 \\
    & PC-Max      & \textbf{0.51$\pm$0.24} & 0.08$\pm$0.11         & 0.63$\pm$0.32         & 0.37$\pm$0.28 \\
    & Shapley-PC  & \textbf{0.56$\pm$0.19} & \textbf{0.17$\pm$0.12} & \textbf{0.81$\pm$0.04} & \textbf{0.61$\pm$0.09} \\
    \midrule
    \multirow{5}{*}{\textbf{\rotatebox{90}{\textbf{V-F1}}}}
    & PC-Stable   & 0.35$\pm$0.35         & 0.14$\pm$0.21         & 0.71$\pm$0.37         & 0.31$\pm$0.38 \\
    & CPC         & 0.52$\pm$0.36         & 0.23$\pm$0.26         & 0.63$\pm$0.43         & 0.36$\pm$0.40 \\
    & MPC         & 0.42$\pm$0.37         & 0.19$\pm$0.24         & 0.60$\pm$0.43         & 0.36$\pm$0.38 \\
    & PC-Max      & \textbf{0.71$\pm$0.34} & 0.22$\pm$0.32         & 0.77$\pm$0.39         & 0.54$\pm$0.41 \\
    & Shapley-PC  & \textbf{0.82$\pm$0.27} & \textbf{0.42$\pm$0.33} & \textbf{0.99$\pm$0.02} & \textbf{0.97$\pm$0.05} \\
    \bottomrule
    \end{tabular}
    \label{tab:20_nodes}
\end{table*}

\begin{figure*}[ht]
    \centering
    \includegraphics[width=\linewidth]{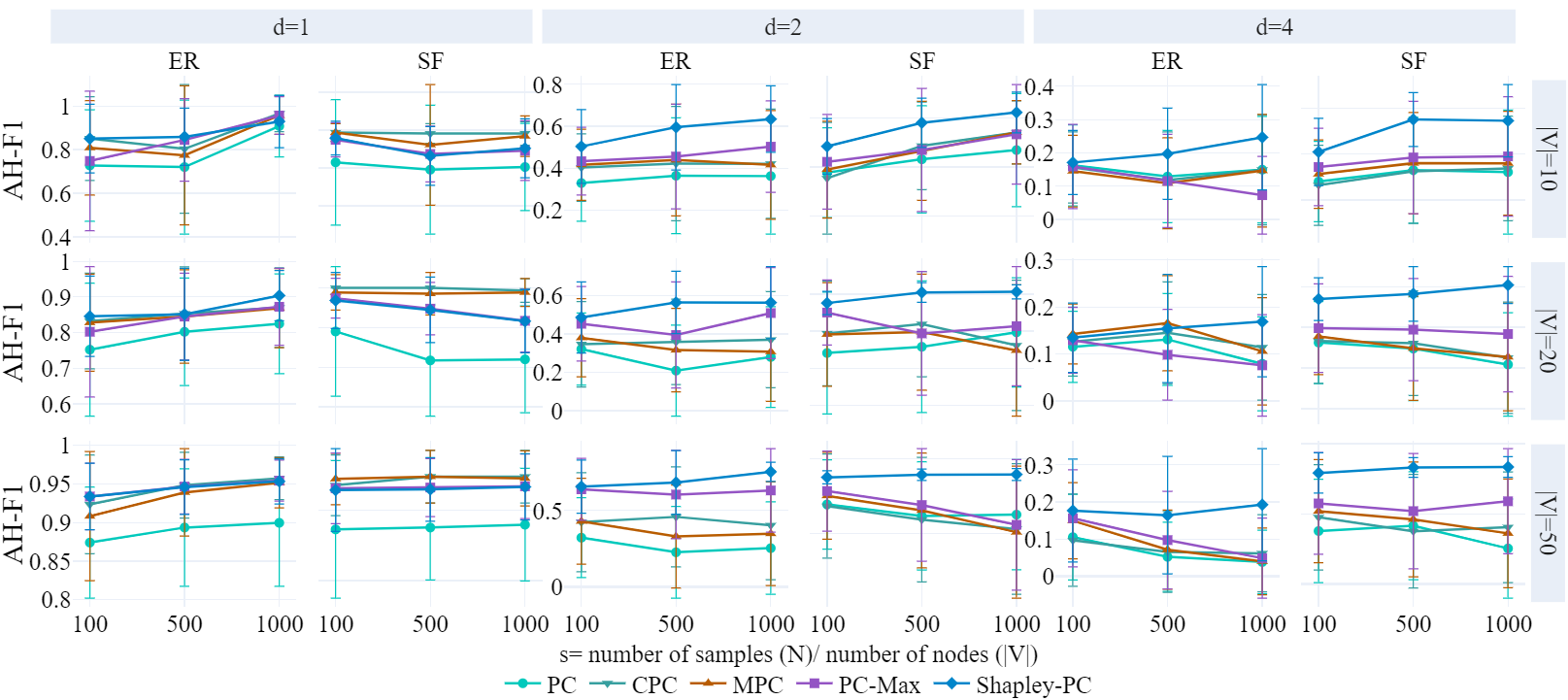}
    \caption{ArrowHead F1 scores by proportional sample size ($s \in \{100, 500, 1000\}$) for the fully synthetic data in \S\ref{sec:experiments}.}
    \label{fig:AHF1_sample}
\end{figure*}

\begin{figure*}[t]
    \centering
    \includegraphics[width=\linewidth]{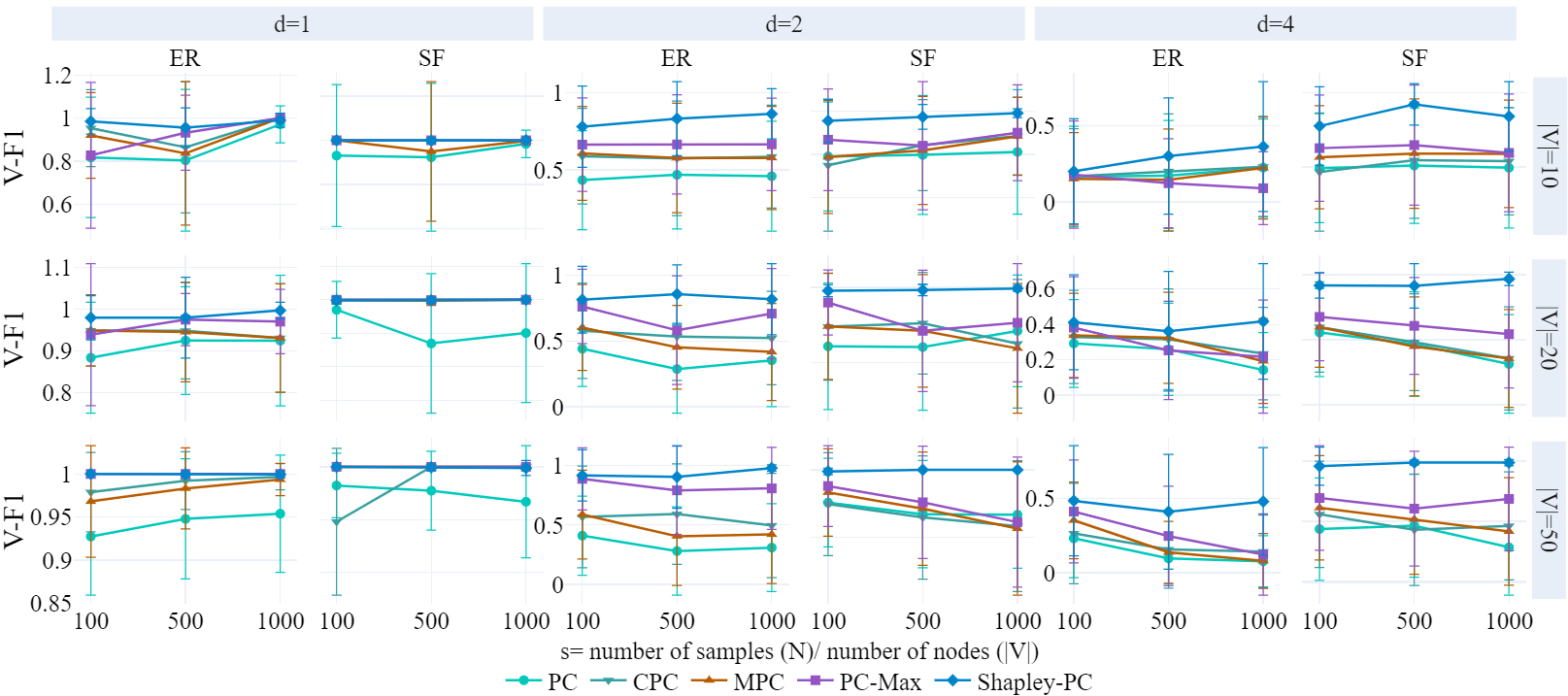}
    \caption{V-structure F1 scores by proportional sample size ($s \in \{100, 500, 1000\}$) for the fully synthetic data in \S\ref{sec:experiments}.}
    \label{fig:VF1_sample}
\end{figure*}

\paragraph{Proportional Sample Size} The results presented in Table~\ref{tab:main_res} in the main text show AH-F1 and V-F1 for proportional sample size $s=1000$. The proportional sample size is the number of samples per node in the dataset, with total number of samples $N=s*|V|$. Here we show the trends for $s \in \{100, 500, 1000\}$, in Fig.~\ref{fig:AHF1_sample} and Fig.~\ref{fig:VF1_sample} for AH-F1 and V-F1, respectively. From the plots, we notice that the trends are mostly flat, demonstrating that none of the methods compared is very ``data-hungry.''


\paragraph{Noise Distributions} Plots by noise distribution are provided as interactive plots in our \href{https://github.com/briziorusso/ShapleyPC/tree/mainpub/results/figs}{repository}, as they would not be easily displayed on A4 paper. No majour differences are observed across noise distributions.

\subsubsection{Additional Metrics}

\paragraph{Precision and Recall}
The results presented in Table~\ref{tab:main_res} in the main text show F1 scores for arrowheads and v-structures' classification. Here we provide a breakdown of the F1 scores into their components: Precision and Recall. In Table~\ref{tab:prec_rec_AH} we can see that Shapley-PC is significantly better (according to a t-test, $\alpha=0.05$) than all other methods, for both precision and recall wrt ArrowHead classifications. Results are analogous for v-structure classifications, shown in Table~\ref{tab:prec_rec_V}.

\paragraph{SHD and SID}
In addition to the results that focus on the v-structures and arrowhead orientations, as presented in main text and the additional results in this section, we also present results using SID~\citep{peters2015structural} in Table~\ref{tab:sid} and SHD~\citep{tsamardinos2006MMHC} in Table~\ref{tab:shd}. Both metrics measure error, hence the lower the better.
Since we calculate all metrics on CPDAGs, SID estimates a best and worst scenario (SID-Low and High, respectively) depending on the orientation of the undirected edges in the output CPDAG.

From Table~\ref{tab:sid}, we can see that Shapley-PC is significantly better than all other methods for the best case scenario (SID-Low) on 10 nodes graphs. For the worst case scenario (SID-High) all methods are on par for ER4, and Shapley-PC and PC-Max are on par, and better than all others for SF4. Shapley-PC is significantly better than all other methods for the remaining types of 10 nodes graphs. For 50 nodes graphs, which are sparser (see results discussion in the main text \S\ref{sec:experiments}), Shapley-PC is better than PC, CPC and MPC, but not significantly better than PC-Max. 

Comparison of Shapley-PC with our baselines based SHD are shown in Table~\ref{tab:shd}. We can see that PC-Stable is significantly worse than all other methods for ER1, ER2 and SF2, while no significant differences are observed for the remaining three graph types.

Overall, Shapley-PC is never worse than any other baseline, based on both SID and SHD.

We remark that SHD and SID are more general graphical metrics, that do not take into account that there can be errors in skeleton and orientations, and that these can be isolated one from the other. With ArrowHead F1, we measure the orientation capabilities of the different methods, that with these metrics are confounded by errors in the skeleton.

\begin{table*}[ht]
    \caption{Details of the Bayesian Network from \texttt{bnlearn} used to generate the pseudo-real datasets in \S\ref{sec:experiments}. "Cat" and "Cont" are counts of the categorical and continuous variables in the produced datasets, respectively. $|N|$ and $|E|$ are the number of nodes and edges in the graph and $d=|E|/|N|$ their proportion.}
    \label{tab:real_data_det}
    \centering
    \begin{tabular}{r|c|c|ccccc}
Dataset Name & Type & Cat & Cont & $|N|$ & $|E|$ & $d$ \\
\hline
\href{https://www.bnlearn.com/bnrepository/discrete-medium.html#alarm}{ALARM}              & Discrete    & 37 & 0 & 37  & 46   & 1.24\\
\href{https://www.bnlearn.com/bnrepository/discrete-medium.html#child}{CHILD}              & Discrete    & 20 & 0 & 20  & 25   & 1.25\\
\href{https://www.bnlearn.com/bnrepository/discrete-large.html#hepar2}{HEPAR2}             & Discrete    & 70 & 0 & 70  & 123  & 1.76\\
\href{https://www.bnlearn.com/bnrepository/discrete-medium.html#insurance}{INSURANCE}      & Discrete    & 27 & 0 & 27  & 52   & 1.93\\
\href{https://www.bnlearn.com/bnrepository/gaussian-verylarge.html#arth150}{ARTH150}       & Gaussian    & 0  & 107 & 107 & 150  & 1.40\\
\href{https://www.bnlearn.com/bnrepository/gaussian-medium.html#ecoli70}{ECOLI70}          & Gaussian    & 0  & 46  & 46  & 70   & 1.52\\
\href{https://www.bnlearn.com/bnrepository/gaussian-large.html#magic-irri}{MAGIC-IRRI}     & Gaussian    & 0  & 64  & 64  & 102  & 1.59\\
\href{https://www.bnlearn.com/bnrepository/gaussian-medium.html#magic-niab}{MAGIC-NIAB}    & Gaussian    & 0  & 44  & 44  & 66   & 1.50\\
\href{https://www.bnlearn.com/bnrepository/clgaussian-medium.html#mehra}{MEHRA}            & Linear Gaussian & 8  & 16  & 24  & 71   & 2.96\\
\href{https://www.bnlearn.com/bnrepository/clgaussian-small.html#sangiovese}{SANGIOVESE}   & Linear Gaussian & 1  & 14  & 15  & 55   & 3.67\\
    \end{tabular}
\end{table*}

\subsection{Pseudo-Real Data}
\label{sec:real_det}

Here we report details on the experiments with pseudo-real data from \S\ref{sec:experiments} of the main text.

\subsubsection{Datasets Details}
\label{sec:bnlearn}
\setlength\intextsep{0pt}

For the experiments on pseudo-real data, we used ten datasets from the \href{https://www.bnlearn.com/bnrepository/}{\texttt{bnlearn}} repository
which is widely used for research in CSL. The datasets are sampled from Bayesian Networks (BN) with fixed conditional probability tables stored in the repository. The BNs used in our experiments are from all three categories in the repository: Discrete, Gaussian and Conditional Linear Gaussian.
The number of nodes vary from 15 to 107 and the number of edges from 25 to 150. Details on the number of nodes, edges and density of the DAGs underlying these data are reported in Table~\ref{tab:real_data_det}, together with links to a more detailed description from the \texttt{bnlearn} repository.

Having downloaded all the .bif or .rda files from the repository, we load the Bayesian network and the associated conditional probability tables and sample 50000 observations, with 10 different seeds. We encoded the labels using the label encoder from \texttt{sklearn} for categorical variables and applied standard scaling from \texttt{sklearn} for the continuous ones. The BNs, together with the code to reproduce the dataset, is provided in our \href{https://github.com/briziorusso/ShapleyPC/}{repository}.

\subsubsection{Statistical Tests}\label{sec:stats_real}
Here we present details of the statistical tests used to measure the significance of the difference in the results presented in Fig.~\ref{fig:pseudoreal} in the main text. In Tables~\ref{tab:tests_alarm_insurance} and \ref{tab:tests_ecoli70_mehra} we provide t-statistics and $p$-values for the Alarm, Insurance, Ecoli70 and Mehra datasets. In each table we present pairwise comparisons of means (shown in brackets together with standard deviations), for the AH-F1 and V-F1 presented in Fig.~\ref{fig:pseudoreal} of the main text.

\paragraph{Additional Metrics}

\begin{figure*}
    \centering
    \includegraphics[width=0.7\linewidth]{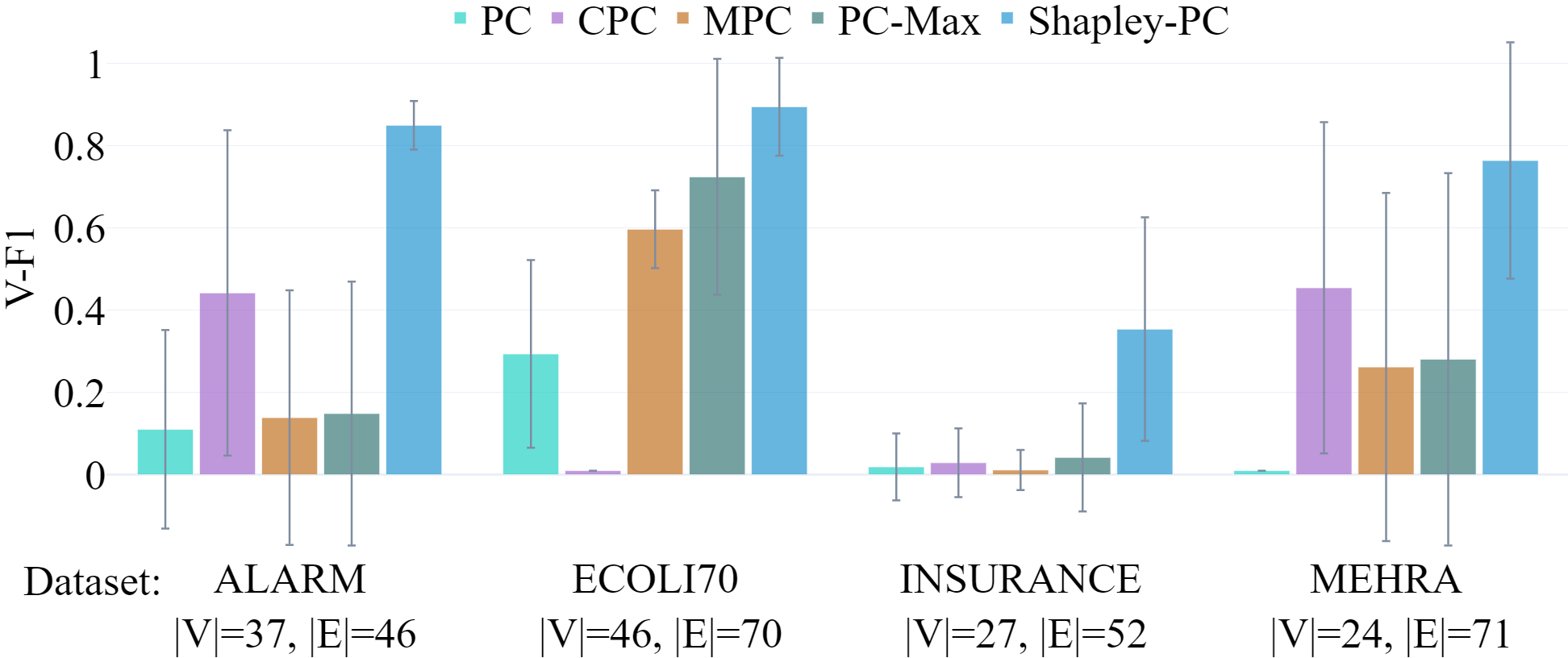}
    \caption{V-structure F1 for the datasets in Fig.~\ref{fig:pseudoreal} in the main text.}
    \label{fig:VF1_real}
\end{figure*}

\begin{figure*}
    \centering
    \includegraphics[width=0.7\linewidth]{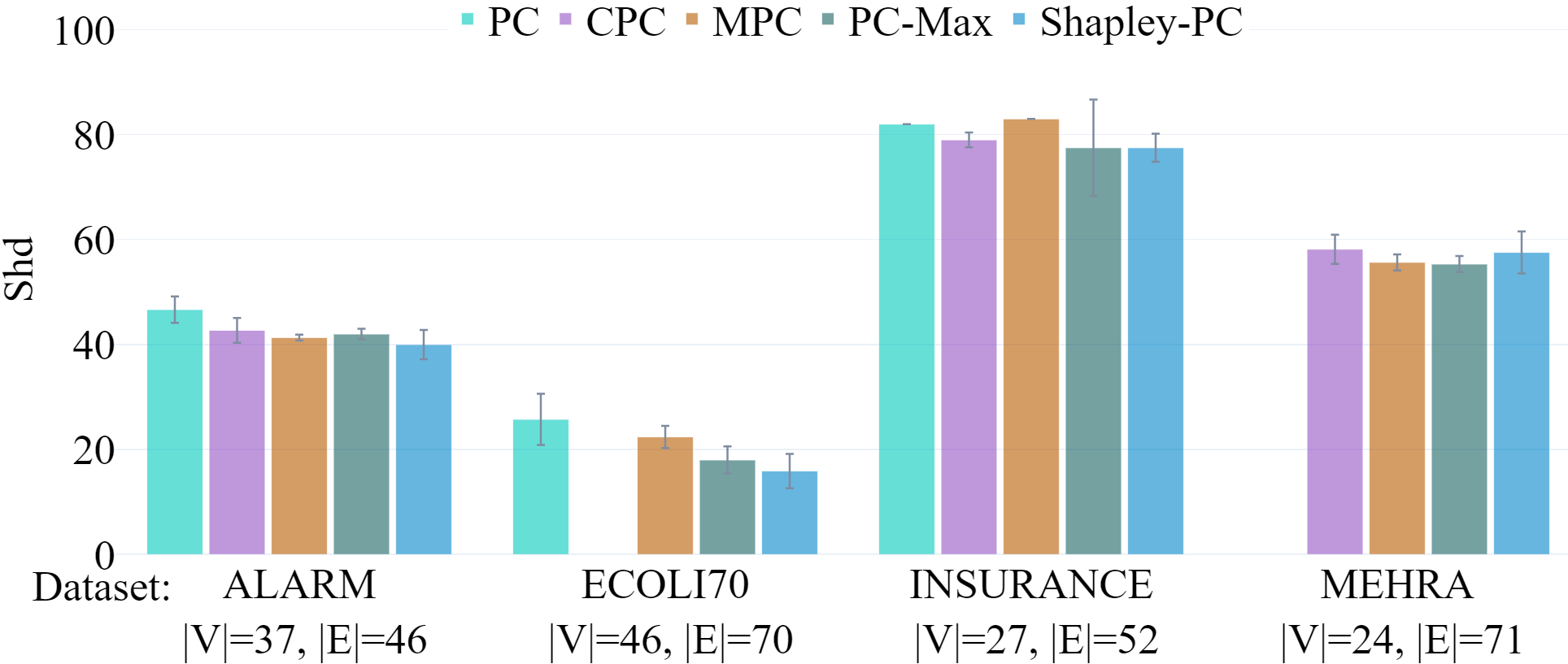}
    \caption{SHD, the lower the better, for the datasets in Fig.~\ref{fig:pseudoreal} in the main text.}
    \label{fig:SHD_real}
\end{figure*}

\begin{figure*}
    \centering
    \includegraphics[width=\linewidth]{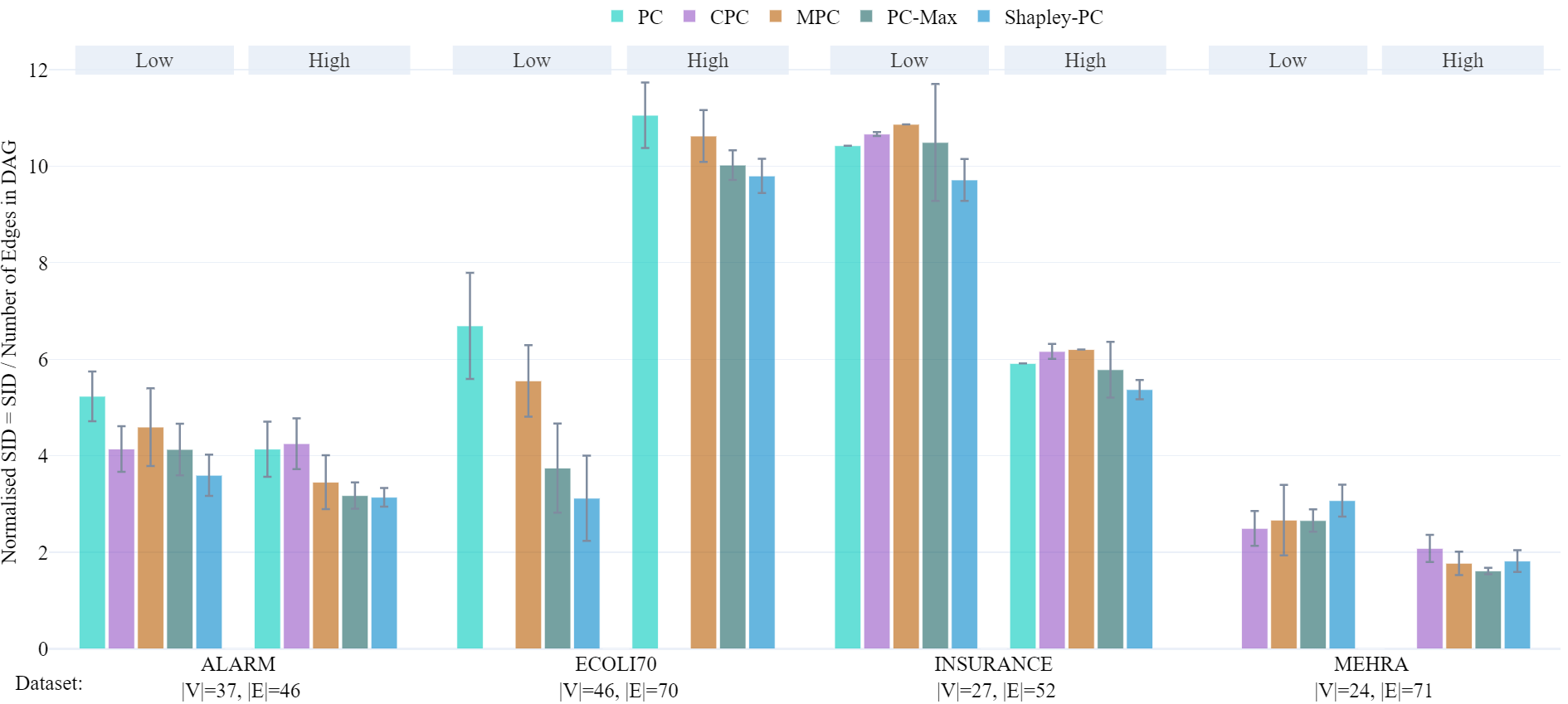}
    \caption{Normalised SID, the lower the better, for the datasets in Fig.~\ref{fig:pseudoreal} in the main text.}
    \label{fig:SID_real}
\end{figure*}

In Fig.~\ref{fig:pseudoreal} in the main text, we show the results on four of the ten dataset detailed in Table~\ref{tab:real_data_det}, according to ArrowHead F1 Score. In this section we report additional metrics, in line with the experiments on synthetic data. In particular, we visualise V-F1 (Fig.~\ref{fig:VF1_real}), SHD (Fig.~\ref{fig:SHD_real}) and SID (Fig.~\ref{fig:SID_real}). 
Precision and Recall are left out because they show very similar trends to AH-F1 and V-F1 presented herein, but are provided in our repository as interactive plots. 

In Fig.~\ref{fig:VF1_real}, we report V-F1 scores for the same set of datasets as in the main text. For AH-F1 (Fig.~\ref{fig:pseudoreal} in the main text), Shapley-PC is significantly better than all other methods on Alarm and Insurance. For V-F1, Shapley-PC is better than all other methods on Alarm, Insurance and Mehra. For Ecoli70, we are on par with PC-Max, and better than all others.
According to SHD (Fig.~\ref{fig:SHD_real}) and SID (Fig.~\ref{fig:SID_real}), no significant differences are observed across datasets and methods. 

\subsubsection{Additional Datasets}
\label{sec:real_add_det}
\begin{figure*}[t]
    \centering
    \includegraphics[width=\linewidth]{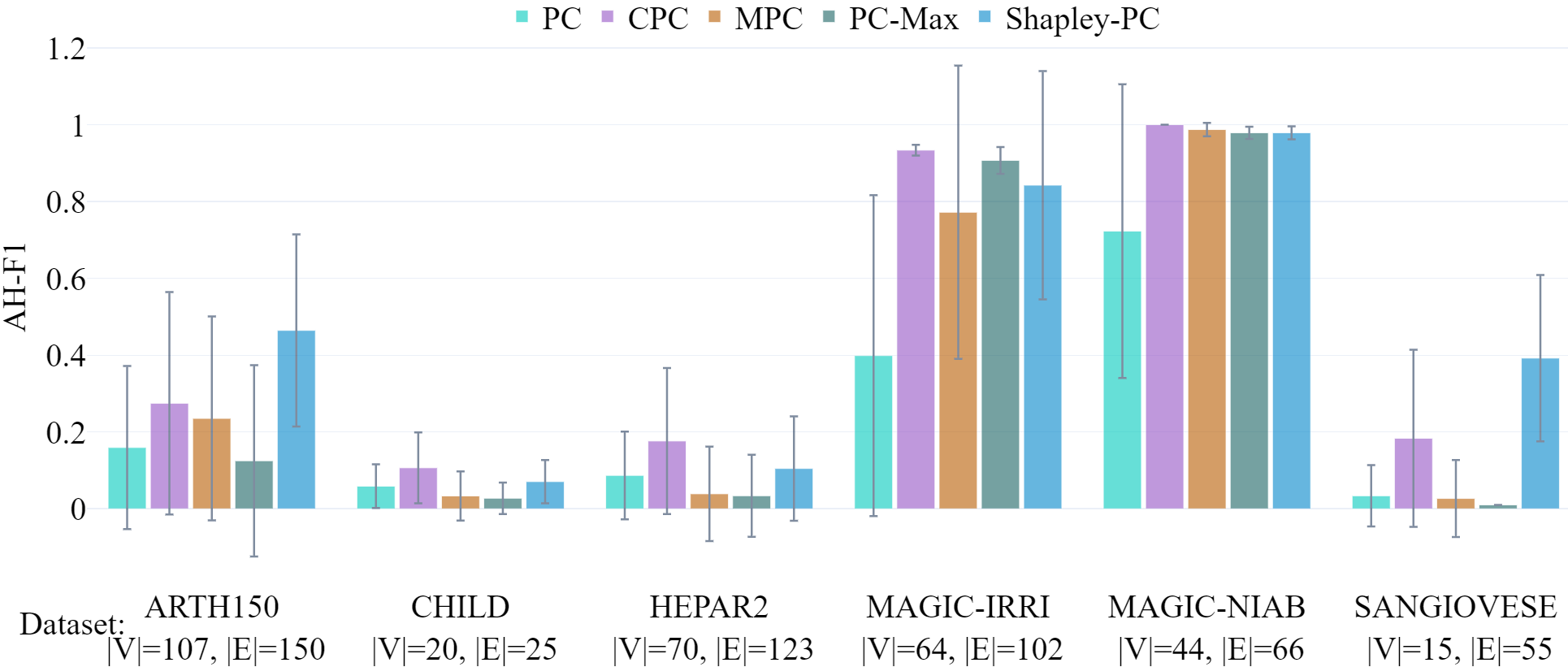}
    \caption{ArrowHead F1 for additional datasets in the \texttt{bnlearn} repository.}
    \label{fig:AHF1_real_extra}
\end{figure*}

\begin{figure*}[t]
    \centering
    \includegraphics[width=\linewidth]{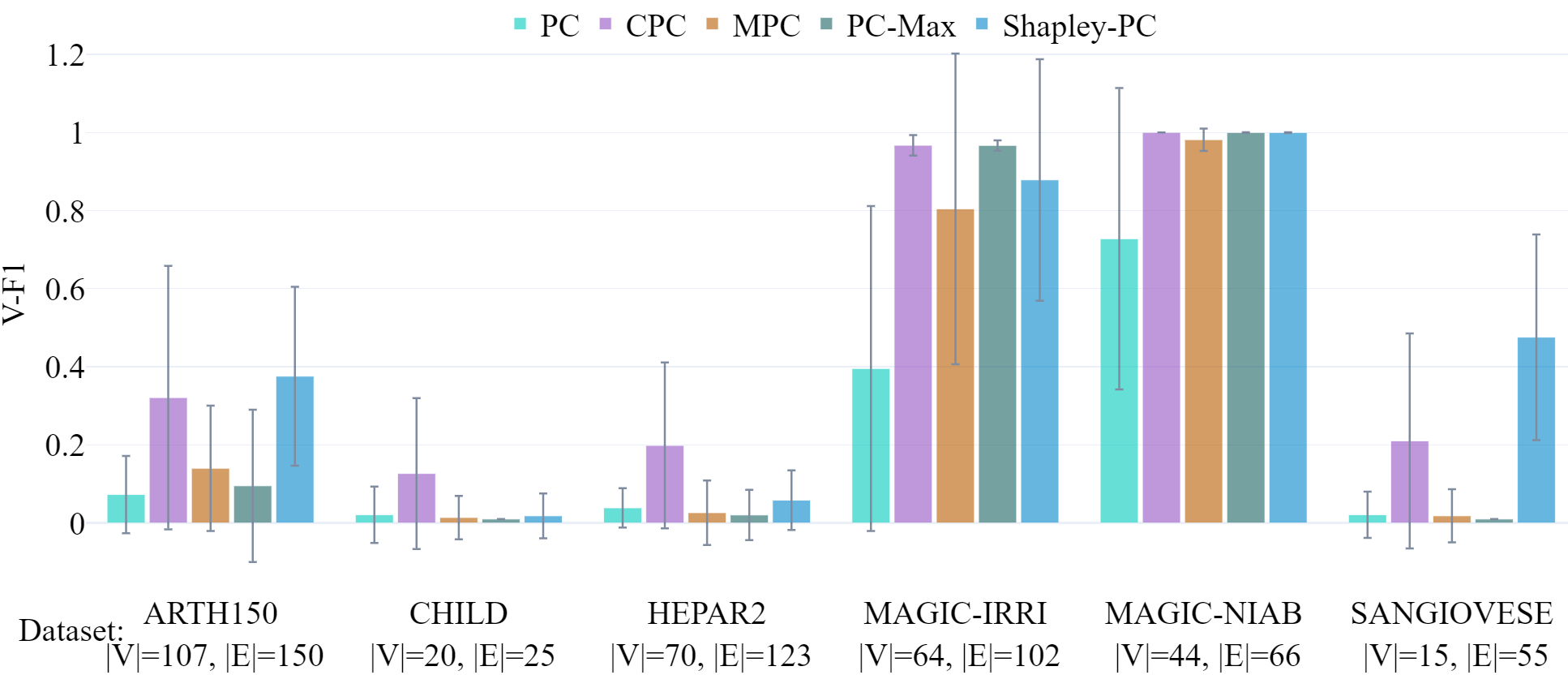}
    \caption{V-structure F1 for additional datasets in the \texttt{bnlearn} repository.}
    \label{fig:VF1_real_extra}
\end{figure*}

The results presented in Fig.~\ref{fig:pseudoreal} show AH-F1 for four datasets out of the ten analysed. We show results for the remaining six datasets (Arth150, Child, Hepar2, Magic-irri, Magic-niab and Sangiovese) in Fig.~\ref{fig:AHF1_real_extra} (AH-F1) and Fig.~\ref{fig:VF1_real_extra} (V-F1). Out of these six datasets, Shapley-PC results to be significantly better than all other methods according to AH-F1 on Arth150 and Sangiovese. According to V-F1, Shapley-PC is better than all others on Sangiovese, and on par with CPC, improving on all other methods, on Arth150. No significant differences are observed on the remaining four datasets, apart from PC-Stable being worse than all other methods on the Magic-irri and Magic-niab datasets.

\clearpage

\begin{table}[t]
    \caption{Two-sample, unequal variance t-tests for difference in means for ER graphs with $|\nodeSet|=10$. Significance levels: 0 '***', 0.001 '**', 0.01 '*', 0.05 '.', 0.1 ' ' 1. DoF: $n_a= n_b=39$.}
    \label{tab:tests_ER_10}
    \vspace{-0.2cm}
    \centering
    \setlength{\tabcolsep}{4pt} 
    \begin{tabular}{c|c|c|c|c|rl}
        \toprule
        \textbf{Type} & \textbf{Metric} & \textbf{Methods} & \textbf{Means$\pm$Std} & \textbf{t} & \multicolumn{2}{c}{\textbf{p-value}} \\
        \midrule
        \multirow{20}{*}{\textbf{ER2}} 
        & \multirow{10}{*}{\textbf{V-F1}}
            & PC-S vs CPC &$0.46\pm0.4$ vs $0.59\pm0.3$&$-1.64$&$0.104$&\\
        & & PC-S vs MPC &$0.46\pm0.4$ vs $0.58\pm0.3$&$-1.54$&$0.128$&\\
        & & PC-S vs PC-M &$0.46\pm0.4$ vs $0.67\pm0.3$&$-2.80$&$0.006$&\!\!\!\!**\\
        & & PC-S vs SPC &$0.46\pm0.4$ vs $0.86\pm0.2$&$-6.53$&$0.000$&\!\!\!\!***\\
        & & CPC vs MPC &$0.59\pm0.3$ vs $0.58\pm0.3$&$0.11$&$0.916$&\\
        & & CPC vs PC-M &$0.59\pm0.3$ vs $0.67\pm0.3$&$-1.12$&$0.265$&\\
        & & CPC vs SPC &$0.59\pm0.3$ vs $0.86\pm0.2$&$-4.73$&$0.000$&\!\!\!\!***\\
        & & MPC vs PC-M &$0.58\pm0.3$ vs $0.67\pm0.3$&$-1.23$&$0.222$&\\
        & & MPC vs SPC &$0.58\pm0.3$ vs $0.86\pm0.2$&$-4.85$&$0.000$&\!\!\!\!***\\
        & & PC-M vs SPC &$0.67\pm0.3$ vs $0.86\pm0.2$&$-3.66$&$0.001$&\!\!\!\!***\\
        \cmidrule{2-7}
        & \multirow{10}{*}{\textbf{AH-F1}} 
            & PC-S vs CPC &$0.36\pm0.3$ vs $0.42\pm0.3$&$-0.98$&$0.331$&\\
        & & PC-S vs MPC &$0.36\pm0.3$ vs $0.42\pm0.3$&$-0.89$&$0.377$&\\
        & & PC-S vs PC-M &$0.36\pm0.3$ vs $0.50\pm0.2$&$-2.52$&$0.014$&\!\!\!\!*\\
        & & PC-S vs SPC &$0.36\pm0.3$ vs $0.63\pm0.2$&$-5.39$&$0.000$&\!\!\!\!***\\
        & & CPC vs MPC &$0.42\pm0.3$ vs $0.42\pm0.3$&$0.09$&$0.926$&\\
        & & CPC vs PC-M &$0.42\pm0.3$ vs $0.50\pm0.2$&$-1.52$&$0.133$&\\
        & & CPC vs SPC &$0.42\pm0.3$ vs $0.63\pm0.2$&$-4.42$&$0.000$&\!\!\!\!***\\
        & & MPC vs PC-M &$0.42\pm0.3$ vs $0.50\pm0.2$&$-1.62$&$0.109$&\\
        & & MPC vs SPC &$0.42\pm0.3$ vs $0.63\pm0.2$&$-4.54$&$0.000$&\!\!\!\!***\\
        & & PC-M vs SPC &$0.50\pm0.2$ vs $0.63\pm0.2$&$-3.08$&$0.003$&\!\!\!\!**\\
        \midrule
        \multirow{20}{*}{\textbf{ER4}} 
        & \multirow{10}{*}{\textbf{V-F1}} 
            & PC-S vs CPC &$0.23\pm0.3$ vs $0.23\pm0.3$&$-0.08$&$0.940$&\\
        & & PC-S vs MPC &$0.23\pm0.3$ vs $0.22\pm0.3$&$0.03$&$0.973$&\\
        & & PC-S vs PC-M &$0.23\pm0.3$ vs $0.09\pm0.2$&$2.15$&$0.035$&\!\!\!\!*\\
        & & PC-S vs SPC &$0.23\pm0.3$ vs $0.36\pm0.4$&$-1.64$&$0.106$&\\
        & & CPC vs MPC &$0.23\pm0.3$ vs $0.22\pm0.3$&$0.11$&$0.915$&\\
        & & CPC vs PC-M &$0.23\pm0.3$ vs $0.09\pm0.2$&$2.19$&$0.032$&\!\!\!\!*\\
        & & CPC vs SPC &$0.23\pm0.3$ vs $0.36\pm0.4$&$-1.55$&$0.125$&\\
        & & MPC vs PC-M &$0.22\pm0.3$ vs $0.09\pm0.2$&$2.05$&$0.044$&\!\!\!\!*\\
        & & MPC vs SPC &$0.22\pm0.3$ vs $0.36\pm0.4$&$-1.64$&$0.106$&\\
        & & PC-M vs SPC &$0.09\pm0.2$ vs $0.36\pm0.4$&$-3.54$&$0.001$&\!\!\!\!***\\
        \cmidrule{2-7}
        & \multirow{10}{*}{\textbf{AH-F1}} 
            & PC-S vs CPC &$0.15\pm0.2$ vs $0.15\pm0.2$&$0.01$&$0.989$&\\
        & & PC-S vs MPC &$0.15\pm0.2$ vs $0.15\pm0.2$&$0.09$&$0.930$&\\
        & & PC-S vs PC-M &$0.15\pm0.2$ vs $0.07\pm0.1$&$2.45$&$0.017$&\!\!\!\!*\\
        & & PC-S vs SPC &$0.15\pm0.2$ vs $0.25\pm0.2$&$-2.73$&$0.008$&\!\!\!\!**\\
        & & CPC vs MPC &$0.15\pm0.2$ vs $0.15\pm0.2$&$0.07$&$0.943$&\\
        & & CPC vs PC-M &$0.15\pm0.2$ vs $0.07\pm0.1$&$2.38$&$0.020$&\!\!\!\!*\\
        & & CPC vs SPC &$0.15\pm0.2$ vs $0.25\pm0.2$&$-2.70$&$0.009$&\!\!\!\!**\\
        & & MPC vs PC-M &$0.15\pm0.2$ vs $0.07\pm0.1$&$2.26$&$0.027$&\!\!\!\!*\\
        & & MPC vs SPC &$0.15\pm0.2$ vs $0.25\pm0.2$&$-2.74$&$0.008$&\!\!\!\!**\\
        & & PC-M vs SPC &$0.07\pm0.1$ vs $0.25\pm0.2$&$-5.59$&$0.000$&\!\!\!\!***\\
        \bottomrule
    \end{tabular}
\end{table}

\begin{table}[t]
    \caption{Two-sample, unequal variance t-tests for difference in means for SF graphs with $|\nodeSet|=10$. Significance levels: 0 '***', 0.001 '**', 0.01 '*', 0.05 '.', 0.1 ' ' 1. DoF: $n_a=n_b=39$.}
    \label{tab:tests_SF_10}
    \centering
    \vspace{-0.2cm}
    \setlength{\tabcolsep}{4pt} 
    \begin{tabular}{c|c|c|c|c|rl}
        \toprule
        \textbf{Type} & \textbf{Metric} & \textbf{Methods} & \textbf{Means$\pm$Std} & \textbf{t} & \multicolumn{2}{c}{\textbf{p-value}} \\
        \midrule
        \multirow{20}{*}{\textbf{SF2}} 
        & \multirow{10}{*}{\textbf{V-F1}}
            & PC-S vs CPC &$0.81\pm0.3$ vs $0.88\pm0.2$&$-1.36$&$0.179$&\\
        & & PC-S vs MPC &$0.81\pm0.3$ vs $0.88\pm0.2$&$-1.35$&$0.183$&\\
        & & PC-S vs PC-M &$0.81\pm0.3$ vs $0.90\pm0.2$&$-1.55$&$0.125$&\\
        & & PC-S vs SPC &$0.81\pm0.3$ vs $0.99\pm0.0$&$-3.93$&$0.000$&\!\!\!\!***\\
        & & CPC vs MPC &$0.88\pm0.2$ vs $0.88\pm0.2$&$0.01$&$0.989$&\\
        & & CPC vs PC-M &$0.88\pm0.2$ vs $0.90\pm0.2$&$-0.36$&$0.721$&\\
        & & CPC vs SPC &$0.88\pm0.2$ vs $0.99\pm0.0$&$-3.73$&$0.001$&\!\!\!\!***\\
        & & MPC vs PC-M &$0.88\pm0.2$ vs $0.90\pm0.2$&$-0.37$&$0.712$&\\
        & & MPC vs SPC &$0.88\pm0.2$ vs $0.99\pm0.0$&$-3.72$&$0.001$&\!\!\!\!***\\
        & & PC-M vs SPC &$0.90\pm0.2$ vs $0.99\pm0.0$&$-2.56$&$0.014$&\!\!\!\!*\\
        \cmidrule{2-7}
        & \multirow{10}{*}{\textbf{AH-F1}} 
            & PC-S vs CPC &$0.67\pm0.2$ vs $0.74\pm0.1$&$-1.72$&$0.091$&\\
        & & PC-S vs MPC &$0.67\pm0.2$ vs $0.74\pm0.1$&$-1.73$&$0.090$&\\
        & & PC-S vs PC-M &$0.67\pm0.2$ vs $0.73\pm0.2$&$-1.32$&$0.191$&\\
        & & PC-S vs SPC &$0.67\pm0.2$ vs $0.82\pm0.1$&$-3.95$&$0.000$&\!\!\!\!***\\
        & & CPC vs MPC &$0.74\pm0.1$ vs $0.74\pm0.1$&$-0.01$&$0.992$&\\
        & & CPC vs PC-M &$0.74\pm0.1$ vs $0.73\pm0.2$&$0.20$&$0.840$&\\
        & & CPC vs SPC &$0.74\pm0.1$ vs $0.82\pm0.1$&$-3.36$&$0.001$&\!\!\!\!**\\
        & & MPC vs PC-M &$0.74\pm0.1$ vs $0.73\pm0.2$&$0.21$&$0.834$&\\
        & & MPC vs SPC &$0.74\pm0.1$ vs $0.82\pm0.1$&$-3.34$&$0.001$&\!\!\!\!**\\
        & & PC-M vs SPC &$0.73\pm0.2$ vs $0.82\pm0.1$&$-2.57$&$0.013$&\!\!\!\!*\\
        \midrule
        \multirow{20}{*}{\textbf{SF4}} 
        & \multirow{10}{*}{\textbf{V-F1}} 
            & PC-S vs CPC &$0.49\pm0.4$ vs $0.54\pm0.4$&$-0.52$&$0.606$&\\
        & & PC-S vs MPC &$0.49\pm0.4$ vs $0.59\pm0.4$&$-1.09$&$0.280$&\\
        & & PC-S vs PC-M &$0.49\pm0.4$ vs $0.59\pm0.4$&$-1.12$&$0.266$&\\
        & & PC-S vs SPC &$0.49\pm0.4$ vs $0.84\pm0.2$&$-4.67$&$0.000$&\!\!\!\!***\\
        & & CPC vs MPC &$0.54\pm0.4$ vs $0.59\pm0.4$&$-0.61$&$0.546$&\\
        & & CPC vs PC-M &$0.54\pm0.4$ vs $0.59\pm0.4$&$-0.66$&$0.511$&\\
        & & CPC vs SPC &$0.54\pm0.4$ vs $0.84\pm0.2$&$-4.43$&$0.000$&\!\!\!\!***\\
        & & MPC vs PC-M &$0.59\pm0.4$ vs $0.59\pm0.4$&$-0.08$&$0.936$&\\
        & & MPC vs SPC &$0.59\pm0.4$ vs $0.84\pm0.2$&$-3.70$&$0.000$&\!\!\!\!***\\
        & & PC-M vs SPC &$0.59\pm0.4$ vs $0.84\pm0.2$&$-3.38$&$0.001$&\!\!\!\!**\\
        \cmidrule{2-7}
        & \multirow{10}{*}{\textbf{AH-F1}} 
            & PC-S vs CPC &$0.32\pm0.3$ vs $0.34\pm0.2$&$-0.29$&$0.774$&\\
        & & PC-S vs MPC &$0.32\pm0.3$ vs $0.36\pm0.2$&$-0.69$&$0.493$&\\
        & & PC-S vs PC-M &$0.32\pm0.3$ vs $0.39\pm0.3$&$-1.16$&$0.252$&\\
        & & PC-S vs SPC &$0.32\pm0.3$ vs $0.54\pm0.2$&$-4.52$&$0.000$&\!\!\!\!***\\
        & & CPC vs MPC &$0.34\pm0.2$ vs $0.36\pm0.2$&$-0.44$&$0.662$&\\
        & & CPC vs PC-M &$0.34\pm0.2$ vs $0.39\pm0.3$&$-0.96$&$0.341$&\\
        & & CPC vs SPC &$0.34\pm0.2$ vs $0.54\pm0.2$&$-4.74$&$0.000$&\!\!\!\!***\\
        & & MPC vs PC-M &$0.36\pm0.2$ vs $0.39\pm0.3$&$-0.55$&$0.583$&\\
        & & MPC vs SPC &$0.36\pm0.2$ vs $0.54\pm0.2$&$-4.25$&$0.000$&\!\!\!\!***\\
        & & PC-M vs SPC &$0.39\pm0.3$ vs $0.54\pm0.2$&$-3.23$&$0.002$&\!\!\!\!**\\
        \bottomrule
    \end{tabular}
\end{table}

\clearpage

\begin{table}[t]
    \caption{Two-sample, unequal variance t-tests for difference in means for ER graphs with $|\nodeSet|=50$. Significance levels: 0 '***', 0.001 '**', 0.01 '*', 0.05 '.', 0.1 ' ' 1. DoF: $n_a=n_b=39$.}
    \label{tab:tests_ER_50}
    \centering
    \vspace{-0.2cm}
    \setlength{\tabcolsep}{4pt} 
    \begin{tabular}{c|c|c|c|c|rl}
        \toprule
        \textbf{Type} & \textbf{Metric} & \textbf{Methods} & \textbf{Means$\pm$Std} & \textbf{t} & \multicolumn{2}{c}{\textbf{p-value}} \\
        \midrule
        \multirow{20}{*}{\textbf{ER2}} 
        & \multirow{10}{*}{\textbf{V-F1}}
            & PC-S vs CPC &$0.31\pm0.4$ vs $0.50\pm0.4$&$-2.05$&$0.043$&\!\!\!\!*\\
        & & PC-S vs MPC &$0.31\pm0.4$ vs $0.42\pm0.4$&$-1.26$&$0.210$&\\
        & & PC-S vs PC-M &$0.31\pm0.4$ vs $0.81\pm0.3$&$-6.24$&$0.000$&\!\!\!\!***\\
        & & PC-S vs SPC &$0.31\pm0.4$ vs $0.98\pm0.0$&$-11.45$&$0.000$&\!\!\!\!***\\
        & & CPC vs MPC &$0.50\pm0.4$ vs $0.42\pm0.4$&$0.79$&$0.429$&\\
        & & CPC vs PC-M &$0.50\pm0.4$ vs $0.81\pm0.3$&$-3.53$&$0.001$&\!\!\!\!***\\
        & & CPC vs SPC &$0.50\pm0.4$ vs $0.98\pm0.0$&$-6.95$&$0.000$&\!\!\!\!***\\
        & & MPC vs PC-M &$0.42\pm0.4$ vs $0.81\pm0.3$&$-4.56$&$0.000$&\!\!\!\!***\\
        & & MPC vs SPC &$0.42\pm0.4$ vs $0.98\pm0.0$&$-8.56$&$0.000$&\!\!\!\!***\\
        & & PC-M vs SPC &$0.81\pm0.3$ vs $0.98\pm0.0$&$-3.13$&$0.003$&\!\!\!\!**\\
        \cmidrule{2-7}
        & \multirow{10}{*}{\textbf{AH-F1}} 
            & PC-S vs CPC &$0.25\pm0.3$ vs $0.40\pm0.4$&$-2.02$&$0.047$&\!\!\!\!*\\
        & & PC-S vs MPC &$0.25\pm0.3$ vs $0.35\pm0.3$&$-1.32$&$0.191$&\\
        & & PC-S vs PC-M &$0.25\pm0.3$ vs $0.63\pm0.3$&$-5.86$&$0.000$&\!\!\!\!***\\
        & & PC-S vs SPC &$0.25\pm0.3$ vs $0.75\pm0.1$&$-10.22$&$0.000$&\!\!\!\!***\\
        & & CPC vs MPC &$0.40\pm0.4$ vs $0.35\pm0.3$&$0.70$&$0.485$&\\
        & & CPC vs PC-M &$0.40\pm0.4$ vs $0.63\pm0.3$&$-3.21$&$0.002$&\!\!\!\!**\\
        & & CPC vs SPC &$0.40\pm0.4$ vs $0.75\pm0.1$&$-6.12$&$0.000$&\!\!\!\!***\\
        & & MPC vs PC-M &$0.35\pm0.3$ vs $0.63\pm0.3$&$-4.11$&$0.000$&\!\!\!\!***\\
        & & MPC vs SPC &$0.35\pm0.3$ vs $0.75\pm0.1$&$-7.42$&$0.000$&\!\!\!\!***\\
        & & PC-M vs SPC &$0.63\pm0.3$ vs $0.75\pm0.1$&$-2.75$&$0.009$&\!\!\!\!**\\
        \midrule
        \multirow{20}{*}{\textbf{ER4}} 
        & \multirow{10}{*}{\textbf{V-F1}} 
            & PC-S vs CPC &$0.08\pm0.2$ vs $0.14\pm0.2$&$-1.38$&$0.172$&\\
        & & PC-S vs MPC &$0.08\pm0.2$ vs $0.08\pm0.2$&$-0.06$&$0.954$&\\
        & & PC-S vs PC-M &$0.08\pm0.2$ vs $0.12\pm0.3$&$-0.88$&$0.380$&\\
        & & PC-S vs SPC &$0.08\pm0.2$ vs $0.48\pm0.4$&$-6.32$&$0.000$&\!\!\!\!***\\
        & & CPC vs MPC &$0.14\pm0.2$ vs $0.08\pm0.2$&$1.30$&$0.198$&\\
        & & CPC vs PC-M &$0.14\pm0.2$ vs $0.12\pm0.3$&$0.35$&$0.731$&\\
        & & CPC vs SPC &$0.14\pm0.2$ vs $0.48\pm0.4$&$-4.86$&$0.000$&\!\!\!\!***\\
        & & MPC vs PC-M &$0.08\pm0.2$ vs $0.12\pm0.3$&$-0.82$&$0.414$&\\
        & & MPC vs SPC &$0.08\pm0.2$ vs $0.48\pm0.4$&$-6.21$&$0.000$&\!\!\!\!***\\
        & & PC-M vs SPC &$0.12\pm0.3$ vs $0.48\pm0.4$&$-4.96$&$0.000$&\!\!\!\!***\\
        \cmidrule{2-7}
        & \multirow{10}{*}{\textbf{AH-F1}} 
            & PC-S vs CPC &$0.04\pm0.1$ vs $0.06\pm0.1$&$-1.10$&$0.275$&\\
        & & PC-S vs MPC &$0.04\pm0.1$ vs $0.04\pm0.1$&$-0.08$&$0.933$&\\
        & & PC-S vs PC-M &$0.04\pm0.1$ vs $0.05\pm0.1$&$-0.48$&$0.635$&\\
        & & PC-S vs SPC &$0.04\pm0.1$ vs $0.19\pm0.2$&$-5.63$&$0.000$&\!\!\!\!***\\
        & & CPC vs MPC &$0.06\pm0.1$ vs $0.04\pm0.1$&$1.00$&$0.320$&\\
        & & CPC vs PC-M &$0.06\pm0.1$ vs $0.05\pm0.1$&$0.55$&$0.584$&\\
        & & CPC vs SPC &$0.06\pm0.1$ vs $0.19\pm0.2$&$-4.52$&$0.000$&\!\!\!\!***\\
        & & MPC vs PC-M &$0.04\pm0.1$ vs $0.05\pm0.1$&$-0.39$&$0.695$&\\
        & & MPC vs SPC &$0.04\pm0.1$ vs $0.19\pm0.2$&$-5.50$&$0.000$&\!\!\!\!***\\
        & & PC-M vs SPC &$0.05\pm0.1$ vs $0.19\pm0.2$&$-4.91$&$0.000$&\!\!\!\!***\\
        \bottomrule
    \end{tabular}
\end{table}

\begin{table}[t]
    \caption{Two-sample, unequal variance t-tests for difference in means for SF graphs with $|\nodeSet|=50$. Significance levels: 0 '***', 0.001 '**', 0.01 '*', 0.05 '.', 0.1 ' ' 1. DoF: $n_a=n_b=39$.}
    \label{tab:tests_SF_50}
    \vspace{-0.2cm}
    \centering
    \setlength{\tabcolsep}{4pt} 
    \begin{tabular}{c|c|c|c|c|rl}
        \toprule
        \textbf{Type} & \textbf{Metric} & \textbf{Methods} & \textbf{Means$\pm$Std} & \textbf{t} & \multicolumn{2}{c}{\textbf{p-value}} \\
        \midrule
        \multirow{20}{*}{\textbf{SF2}} 
        & \multirow{10}{*}{\textbf{V-F1}}
            & PC-S vs CPC &$0.67\pm0.4$ vs $0.58\pm0.5$&$0.85$&$0.397$&\\
        & & PC-S vs MPC &$0.67\pm0.4$ vs $0.57\pm0.5$&$1.03$&$0.307$&\\
        & & PC-S vs PC-M &$0.67\pm0.4$ vs $0.61\pm0.5$&$0.56$&$0.579$&\\
        & & PC-S vs SPC &$0.67\pm0.4$ vs $1.00\pm0.0$&$-5.30$&$0.000$&\!\!\!\!***\\
        & & CPC vs MPC &$0.58\pm0.5$ vs $0.57\pm0.5$&$0.17$&$0.864$&\\
        & & CPC vs PC-M &$0.58\pm0.5$ vs $0.61\pm0.5$&$-0.27$&$0.787$&\\
        & & CPC vs SPC &$0.58\pm0.5$ vs $1.00\pm0.0$&$-5.44$&$0.000$&\!\!\!\!***\\
        & & MPC vs PC-M &$0.57\pm0.5$ vs $0.61\pm0.5$&$-0.44$&$0.661$&\\
        & & MPC vs SPC &$0.57\pm0.5$ vs $1.00\pm0.0$&$-5.58$&$0.000$&\!\!\!\!***\\
        & & PC-M vs SPC &$0.61\pm0.5$ vs $1.00\pm0.0$&$-5.08$&$0.000$&\!\!\!\!***\\
        \cmidrule{2-7}
        & \multirow{10}{*}{\textbf{AH-F1}} 
            & PC-S vs CPC &$0.63\pm0.4$ vs $0.53\pm0.4$&$1.07$&$0.290$&\\
        & & PC-S vs MPC &$0.63\pm0.4$ vs $0.51\pm0.4$&$1.29$&$0.201$&\\
        & & PC-S vs PC-M &$0.63\pm0.4$ vs $0.56\pm0.4$&$0.75$&$0.454$&\\
        & & PC-S vs SPC &$0.63\pm0.4$ vs $0.90\pm0.0$&$-4.58$&$0.000$&\!\!\!\!***\\
        & & CPC vs MPC &$0.53\pm0.4$ vs $0.51\pm0.4$&$0.21$&$0.832$&\\
        & & CPC vs PC-M &$0.53\pm0.4$ vs $0.56\pm0.4$&$-0.29$&$0.773$&\\
        & & CPC vs SPC &$0.53\pm0.4$ vs $0.90\pm0.0$&$-5.26$&$0.000$&\!\!\!\!***\\
        & & MPC vs PC-M &$0.51\pm0.4$ vs $0.56\pm0.4$&$-0.50$&$0.617$&\\
        & & MPC vs SPC &$0.51\pm0.4$ vs $0.90\pm0.0$&$-5.50$&$0.000$&\!\!\!\!***\\
        & & PC-M vs SPC &$0.56\pm0.4$ vs $0.90\pm0.0$&$-4.85$&$0.000$&\!\!\!\!***\\
        \midrule
        \multirow{20}{*}{\textbf{SF4}} 
        & \multirow{10}{*}{\textbf{V-F1}} 
            & PC-S vs CPC &$0.28\pm0.4$ vs $0.46\pm0.4$&$-1.87$&$0.065$&\!\!\!\!.\\
        & & PC-S vs MPC &$0.28\pm0.4$ vs $0.42\pm0.4$&$-1.39$&$0.167$&\\
        & & PC-S vs PC-M &$0.28\pm0.4$ vs $0.69\pm0.4$&$-4.34$&$0.000$&\!\!\!\!***\\
        & & PC-S vs SPC &$0.28\pm0.4$ vs $0.99\pm0.0$&$-11.16$&$0.000$&\!\!\!\!***\\
        & & CPC vs MPC &$0.46\pm0.4$ vs $0.42\pm0.4$&$0.46$&$0.649$&\\
        & & CPC vs PC-M &$0.46\pm0.4$ vs $0.69\pm0.4$&$-2.28$&$0.025$&\!\!\!\!*\\
        & & CPC vs SPC &$0.46\pm0.4$ vs $0.99\pm0.0$&$-7.40$&$0.000$&\!\!\!\!***\\
        & & MPC vs PC-M &$0.42\pm0.4$ vs $0.69\pm0.4$&$-2.75$&$0.007$&\!\!\!\!**\\
        & & MPC vs SPC &$0.42\pm0.4$ vs $0.99\pm0.0$&$-8.07$&$0.000$&\!\!\!\!***\\
        & & PC-M vs SPC &$0.69\pm0.4$ vs $0.99\pm0.0$&$-4.42$&$0.000$&\!\!\!\!***\\
        \cmidrule{2-7}
        & \multirow{10}{*}{\textbf{AH-F1}} 
            & PC-S vs CPC &$0.25\pm0.4$ vs $0.40\pm0.4$&$-1.80$&$0.076$&\!\!\!\!.\\
        & & PC-S vs MPC &$0.25\pm0.4$ vs $0.36\pm0.4$&$-1.29$&$0.201$&\\
        & & PC-S vs PC-M &$0.25\pm0.4$ vs $0.59\pm0.4$&$-4.12$&$0.000$&\!\!\!\!***\\
        & & PC-S vs SPC &$0.25\pm0.4$ vs $0.83\pm0.1$&$-10.12$&$0.000$&\!\!\!\!***\\
        & & CPC vs MPC &$0.40\pm0.4$ vs $0.36\pm0.4$&$0.50$&$0.619$&\\
        & & CPC vs PC-M &$0.40\pm0.4$ vs $0.59\pm0.4$&$-2.15$&$0.034$&\!\!\!\!*\\
        & & CPC vs SPC &$0.40\pm0.4$ vs $0.83\pm0.1$&$-6.77$&$0.000$&\!\!\!\!***\\
        & & MPC vs PC-M &$0.36\pm0.4$ vs $0.59\pm0.4$&$-2.68$&$0.009$&\!\!\!\!**\\
        & & MPC vs SPC &$0.36\pm0.4$ vs $0.83\pm0.1$&$-7.57$&$0.000$&\!\!\!\!***\\
        & & PC-M vs SPC &$0.59\pm0.4$ vs $0.83\pm0.1$&$-4.04$&$0.000$&\!\!\!\!***\\
        \bottomrule
    \end{tabular}
\end{table}

\begin{table*}[ht]
    \centering
    \caption{ArrowHead Precision and Recall for ER$d$ and SF$d$ graphs of 10 and 50 nodes. $d$ is the number of edges per node in the true DAG. Bold if significantly different from the runner-up (according to a t-test, $\alpha=0.05$).}
    \label{tab:prec_rec_AH}
    \begin{tabular}{cc|r|cccc}
    \toprule
    & & \textbf{Method} & \textbf{ER2} & \textbf{ER4} & \textbf{SF2} & \textbf{SF4} \\ 
    \midrule
    \multirow{10}{*}{\textbf{\rotatebox{90}{$|\nodeSet| = 10$}}} & \multirow{5}{*}{\textbf{\rotatebox{90}{Precision}}} 
    & PC-Stable & 0.4$\pm$0.3 & 0.16$\pm$0.18 & 0.73$\pm$0.25 & 0.42$\pm$0.35 \\
    & & CPC & 0.46$\pm$0.29 & 0.16$\pm$0.19 & 0.82$\pm$0.14 & 0.44$\pm$0.29 \\
    & & MPC & 0.45$\pm$0.28 & 0.16$\pm$0.19 & 0.82$\pm$0.14 & 0.47$\pm$0.29 \\
    & & PC-Max & 0.55$\pm$0.24 & 0.08$\pm$0.13 & 0.80$\pm$0.23 & 0.50$\pm$0.33 \\
    & & Shapley-PC & \textbf{0.70$\pm$0.18} & \textbf{0.29$\pm$0.20} & \textbf{0.91$\pm$0.11} & \textbf{0.71$\pm$0.20} \\ 
    \cmidrule{2-7}
    & \multirow{5}{*}{\textbf{\rotatebox{90}{Recall}}} 
    & PC-Stable & 0.34$\pm$0.26 & 0.14$\pm$0.15 & 0.62$\pm$0.22 & 0.26$\pm$0.22 \\
    & & CPC & 0.39$\pm$0.25 & 0.14$\pm$0.16 & 0.68$\pm$0.13 & 0.28$\pm$0.19 \\
    & & MPC & 0.39$\pm$0.25 & 0.14$\pm$0.16 & 0.68$\pm$0.13 & 0.30$\pm$0.19 \\
    & & PC-Max & 0.47$\pm$0.21 & 0.07$\pm$0.11 & 0.68$\pm$0.19 & 0.32$\pm$0.22 \\
    & & Shapley-PC & \textbf{0.59$\pm$0.16} & \textbf{0.22$\pm$0.14} & \textbf{0.76$\pm$0.08} & \textbf{0.44$\pm$0.14} \\ 
    \midrule
    \multirow{10}{*}{\textbf{\rotatebox{90}{$|\nodeSet| = 50$}}} & \multirow{5}{*}{\textbf{\rotatebox{90}{Precision}}} 
    & PC-Stable & 0.29$\pm$0.35 & 0.06$\pm$0.13 & 0.65$\pm$0.38 & 0.27$\pm$0.38 \\
    & & CPC & 0.46$\pm$0.41 & 0.11$\pm$0.18 & 0.56$\pm$0.46 & 0.44$\pm$0.42 \\
    & & MPC & 0.40$\pm$0.39 & 0.06$\pm$0.14 & 0.53$\pm$0.46 & 0.39$\pm$0.42 \\
    & & PC-Max & 0.72$\pm$0.31 & 0.08$\pm$0.18 & 0.58$\pm$0.46 & 0.63$\pm$0.40 \\
    & & Shapley-PC & \textbf{0.86$\pm$0.05} & \textbf{0.33$\pm$0.25} & \textbf{0.93$\pm$0.04} & \textbf{0.89$\pm$0.07} \\ 
    \cmidrule{2-7}
    & \multirow{5}{*}{\textbf{\rotatebox{90}{Recall}}} 
    & PC-Stable & 0.22$\pm$0.27 & 0.03$\pm$0.06 & 0.61$\pm$0.36 & 0.24$\pm$0.33 \\
    & & CPC & 0.36$\pm$0.32 & 0.04$\pm$0.07 & 0.50$\pm$0.42 & 0.38$\pm$0.37 \\
    & & MPC & 0.31$\pm$0.30 & 0.03$\pm$0.07 & 0.49$\pm$0.43 & 0.34$\pm$0.36 \\
    & & PC-Max & 0.56$\pm$0.25 & 0.03$\pm$0.08 & 0.54$\pm$0.42 & 0.56$\pm$0.36 \\
    & & Shapley-PC & \textbf{0.67$\pm$0.07} & \textbf{0.14$\pm$0.11} & \textbf{0.86$\pm$0.05} & \textbf{0.79$\pm$0.08} \\ 
    \bottomrule
    \end{tabular}
\end{table*}

\begin{table*}[ht]
    \centering
    \caption{V-structure Precision and Recall for ER$d$ and SF$d$ graphs of 10 and 50 nodes. $d$ is the number of edges per node in the true DAG. Bold if significantly different from the runner-up (according to a t-test, $\alpha=0.05$).}
    \label{tab:prec_rec_V}
    \begin{tabular}{cc|r|cccc}
    \toprule
    & & \textbf{Method} & \textbf{ER2} & \textbf{ER4} & \textbf{SF2} & \textbf{SF4} \\ 
    \midrule
    \multirow{10}{*}{\textbf{\rotatebox{90}{$\lvert \nodeSet \rvert = 10$}}} & \multirow{5}{*}{\textbf{\rotatebox{90}{Precision}}} 
    & PC-Stable & 0.46$\pm$0.37 & 0.35$\pm$0.41 & 0.88$\pm$0.27 & 0.56$\pm$0.45 \\
    & & CPC & 0.60$\pm$0.36 & 0.32$\pm$0.41 & 0.96$\pm$0.09 & 0.65$\pm$0.42 \\
    & & MPC & 0.60$\pm$0.37 & 0.31$\pm$0.40 & 0.95$\pm$0.09 & 0.68$\pm$0.40 \\
    & & PC-Max & \textbf{0.72$\pm$0.33} & 0.19$\pm$0.39 & 0.92$\pm$0.22 & 0.65$\pm$0.42 \\
    & & Shapley-PC & \textbf{0.85$\pm$0.21} & 0.55$\pm$0.47 & \textbf{1.0$\pm$0.01} & \textbf{0.86$\pm$0.25} \\ 
    \cmidrule{2-7}
    & \multirow{5}{*}{\textbf{\rotatebox{90}{Recall}}} 
    & PC-Stable & 0.48$\pm$0.37 & 0.27$\pm$0.39 & 0.77$\pm$0.30 & 0.47$\pm$0.40 \\
    & & CPC & 0.60$\pm$0.35 & 0.25$\pm$0.36 & 0.84$\pm$0.22 & 0.50$\pm$0.36 \\
    & & MPC & 0.60$\pm$0.35 & 0.24$\pm$0.37 & 0.85$\pm$0.22 & 0.56$\pm$0.37 \\
    & & PC-Max & 0.67$\pm$0.32 & 0.09$\pm$0.24 & 0.88$\pm$0.23 & 0.56$\pm$0.39 \\
    & & Shapley-PC & \textbf{0.91$\pm$0.14} & \textbf{0.35$\pm$0.42} & \textbf{0.99$\pm$0.04} & \textbf{0.84$\pm$0.24} \\ 
    \midrule
    \multirow{10}{*}{\textbf{\rotatebox{90}{$\lvert \nodeSet \rvert = 50$}}} & \multirow{5}{*}{\textbf{\rotatebox{90}{Precision}}} 
    & PC-Stable & 0.32$\pm$0.38 & 0.07$\pm$0.16 & 0.70$\pm$0.41 & 0.31$\pm$0.42 \\
    & & CPC & 0.50$\pm$0.44 & 0.12$\pm$0.21 & 0.59$\pm$0.49 & 0.47$\pm$0.45 \\
    & & MPC & 0.42$\pm$0.41 & 0.07$\pm$0.16 & 0.57$\pm$0.50 & 0.42$\pm$0.45 \\
    & & PC-Max & 0.80$\pm$0.34 & 0.12$\pm$0.26 & 0.62$\pm$0.49 & 0.69$\pm$0.43 \\
    & & Shapley-PC & \textbf{0.97$\pm$0.04} & \textbf{0.44$\pm$0.34} & \textbf{1.0$\pm$0.0} & \textbf{0.98$\pm$0.03} \\ 
    \cmidrule{2-7}
    & \multirow{5}{*}{\textbf{\rotatebox{90}{Recall}}} 
    & PC-Stable & 0.30$\pm$0.36 & 0.09$\pm$0.20 & 0.64$\pm$0.39 & 0.27$\pm$0.38 \\
    & & CPC & 0.50$\pm$0.44 & 0.18$\pm$0.30 & 0.58$\pm$0.48 & 0.46$\pm$0.44 \\
    & & MPC & 0.42$\pm$0.42 & 0.10$\pm$0.23 & 0.56$\pm$0.49 & 0.41$\pm$0.44 \\
    & & PC-Max & 0.82$\pm$0.35 & 0.13$\pm$0.29 & 0.61$\pm$0.48 & 0.68$\pm$0.43 \\
    & & Shapley-PC & \textbf{0.99$\pm$0.02} & \textbf{0.53$\pm$0.39} & \textbf{1.0$\pm$0.0} & \textbf{0.99$\pm$0.02} \\ 
    \bottomrule
    \end{tabular}
\end{table*}

\clearpage

\begin{table*}[ht]
    \centering
    \caption{Structural Interventional Distance (SID, the lower the better) for ER$d$ and SF$d$ graphs of 10 and 50 nodes. $d$ is the number of edges per node in the true DAG. Bold if significantly different from the runner-up (according to a t-test, $\alpha=0.05$).}
    \label{tab:sid}
    \begin{tabular}{cc|r|cccc}
    \toprule
    & & \textbf{Method} & \textbf{ER2} & \textbf{ER4} & \textbf{SF2} & \textbf{SF4} \\ 
    \midrule
    \multirow{10}{*}{\textbf{\rotatebox{90}{$\lvert \nodeSet \rvert = 10$}}} & \multirow{5}{*}{\textbf{\rotatebox{90}{SID-Low}}} 
    & PC-Stable & 42$\pm$17 & 72$\pm$7 & 14$\pm$10 & 37$\pm$14 \\
    & & CPC & 39$\pm$14 & 70$\pm$6 & 13$\pm$10 & 42$\pm$12 \\
    & & MPC & 40$\pm$13 & 70$\pm$6 & 13$\pm$10 & 42$\pm$12 \\
    & & PC-Max & 35$\pm$11 & 71$\pm$9 & 11$\pm$7 & 36$\pm$14 \\
    & & Shapley-PC & \textbf{29$\pm$13} & \textbf{65$\pm$10} & \textbf{7$\pm$4} & \textbf{28$\pm$12} \\ 
    \cmidrule{2-7}
    & \multirow{5}{*}{\textbf{\rotatebox{90}{SID-High}}} 
    & PC-Stable & 59$\pm$14 & 78$\pm$6 & 27$\pm$12 & 57$\pm$14 \\
    & & CPC & 57$\pm$11 & 76$\pm$5 & 26$\pm$12 & 60$\pm$14 \\
    & & MPC & 57$\pm$10 & 76$\pm$5 & 26$\pm$11 & 57$\pm$15 \\
    & & PC-Max & 56$\pm$8 & 79$\pm$6 & 22$\pm$8 & \textbf{50$\pm$13} \\
    & & Shapley-PC & \textbf{51$\pm$11} & 77$\pm$6 & \textbf{20$\pm$7} & \textbf{48$\pm$13} \\ 
    \midrule
    \multirow{10}{*}{\textbf{\rotatebox{90}{$\lvert \nodeSet \rvert = 50$}}} & \multirow{5}{*}{\textbf{\rotatebox{90}{SID-Low}}} 
    & PC-Stable & 792$\pm$196 & 1906$\pm$155 & 189$\pm$84 & 394$\pm$172 \\
    & & CPC & 559$\pm$168 & 1904$\pm$189 & 128$\pm$59 & 308$\pm$116 \\
    & & MPC & 644$\pm$156 & 1880$\pm$157 & \textbf{119$\pm$57} & 315$\pm$107 \\
    & & PC-Max & \textbf{494$\pm$127} & \textbf{1682$\pm$197} & \textbf{113$\pm$57} & \textbf{216$\pm$109} \\
    & & Shapley-PC & \textbf{471$\pm$140} & \textbf{1569$\pm$172} & \textbf{93$\pm$43} & \textbf{173$\pm$85} \\ 
    \cmidrule{2-7}
    & \multirow{5}{*}{\textbf{\rotatebox{90}{SID-High}}} 
    & PC-Stable & 1073$\pm$254 & 2110$\pm$139 & 263$\pm$96 & 484$\pm$189 \\
    & & CPC & 900$\pm$244 & 2188$\pm$110 & 221$\pm$74 & 426$\pm$155 \\
    & & MPC & 953$\pm$229 & 2136$\pm$120 & \textbf{199$\pm$64} & 419$\pm$143 \\
    & & PC-Max & \textbf{803$\pm$251} & \textbf{2012$\pm$191} & \textbf{186$\pm$75} & \textbf{311$\pm$147} \\
    & & Shapley-PC & \textbf{792$\pm$268} & \textbf{1962$\pm$172} & \textbf{170$\pm$54} & \textbf{272$\pm$125} \\ 
    \bottomrule
    \end{tabular}
\end{table*}

\begin{table*}[ht]
    \centering
\begin{tabular}{lr|ccc|ccc}
\toprule
\multicolumn{2}{l}{Nodes} & \multicolumn{1}{c}{ER1} & \multicolumn{1}{c}{ER2} & \multicolumn{1}{c}{ER4} & \multicolumn{1}{c}{SF1} & \multicolumn{1}{c}{SF2} & \multicolumn{1}{c}{SF4} \\
\multicolumn{2}{l}{10} & \multicolumn{1}{c}{SHD} & \multicolumn{1}{c}{SHD} & \multicolumn{1}{c}{SHD} & \multicolumn{1}{c}{SHD} & \multicolumn{1}{c}{SHD} & \multicolumn{1}{c}{SHD} \\
\midrule
&PC-Stable & 1.3$\pm$1.8 & 13.0$\pm$4.2 & 35.4$\pm$3.6 & 2.0$\pm$1.9 & 6.5$\pm$2.4 & 18.5$\pm$3.4 \\
&CPC & 0.9$\pm$1.3 & 12.8$\pm$4.1 & 34.9$\pm$3.1 & 1.6$\pm$1.5 & 6.2$\pm$2.4 & 20.4$\pm$3.4 \\
&MPC & 0.9$\pm$1.3 & 13.1$\pm$4.2 & 34.8$\pm$3.0 & 1.6$\pm$1.6 & 6.2$\pm$2.3 & 20.1$\pm$3.4 \\
&PC-Max & 0.9$\pm$1.3 & 12.0$\pm$3.1 & 36.7$\pm$2.5 & 1.7$\pm$1.6 & 5.9$\pm$2.3 & \textbf{18.3$\pm$3.6} \\
&Shapley-PC & 1.1$\pm$1.6 & 10.7$\pm$4.1 & \textbf{34.4$\pm$3.5} & 1.6$\pm$1.6 & \textbf{4.9$\pm$1.8} & \textbf{17.1$\pm$3.2} \\
\midrule 
\multicolumn{2}{l}{50} & \multicolumn{6}{l}{}  \\
\midrule
& PC-Stable & 6.9$\pm$4.1 & 53.3$\pm$8.6 & 188.0$\pm$10.8 & 5.5$\pm$2.5 & 17.5$\pm$5.9 & 34.6$\pm$13.0 \\
& CPC & \textbf{4.8$\pm$2.3} & 42.5$\pm$9.9 & 180.5$\pm$8.7 & 4.5$\pm$1.8 & \textbf{14.6$\pm$4.0} & 31.7$\pm$11.4 \\
& MPC & \textbf{4.9$\pm$2.5} & 45.4$\pm$8.4 & 183.3$\pm$12.6 & 4.6$\pm$1.8 & \textbf{14.6$\pm$4.5} & 32.6$\pm$11.2 \\
& PC-Max & \textbf{4.8$\pm$2.4} & \textbf{39.5$\pm$7.6} & 178.1$\pm$8.5 & 4.7$\pm$1.9 & \textbf{13.5$\pm$4.0} & 28.7$\pm$13.1 \\
& Shapley-PC & \textbf{4.9$\pm$2.4} & \textbf{38.5$\pm$8.1} & 171.7$\pm$11.0 & 4.7$\pm$2.0 & \textbf{13.8$\pm$4.7} & 26.7$\pm$11.9 \\
\bottomrule
\end{tabular}
    \caption{Structural Hamming Distance (SHD, the lower the better) for ER$d$ and SF$d$ graphs of 10 and 50 nodes. $d$ is the number of edges per node in the true DAG. Bold if significantly different from the runner-up (according to a t-test, $\alpha=0.05$).}
    \label{tab:shd}
\end{table*}

\clearpage
\begin{table}[t]
    \caption{Two-sample, unequal variance t-tests for difference in means for Alarm and Insurance Data. Significance levels: 0 '***', 0.001 '**', 0.01 '*' 0.05 '.' 0.1 ' ' 1. DoF: $n_a=n_b=9$.}
    \label{tab:tests_alarm_insurance}
    \centering
    \vspace{-0.2cm}
    \setlength{\tabcolsep}{4pt} 
    \begin{tabular}{c|c|c|c|c|rl}
        \toprule
        \textbf{Dataset} & \textbf{Metric} & \textbf{Methods} & \textbf{Means$\pm$Std} & \textbf{t} & \multicolumn{2}{c}{\textbf{p-value}} \\
        \midrule
        \multirow{20}{*}{\textbf{Alarm}} 
        & \multirow{10}{*}{\textbf{V-F1}}
            & CPC vs MPC &$0.53\pm0.4$ vs $0.24\pm0.4$&$1.75$&$0.098$&\!\!\!\!.\\
        & & CPC vs PC-S &$0.53\pm0.4$ vs $0.18\pm0.3$&$2.37$&$0.030$&\!\!\!\!*\\
        & & CPC vs PC-M &$0.53\pm0.4$ vs $0.24\pm0.4$&$1.73$&$0.101$&\\
        & & CPC vs SPC &$0.53\pm0.4$ vs $0.85\pm0.1$&$-2.67$&$0.024$&\!\!\!\!*\\
        & & MPC vs PC-S &$0.24\pm0.4$ vs $0.18\pm0.3$&$0.39$&$0.700$&\\
        & & MPC vs PC-M &$0.24\pm0.4$ vs $0.24\pm0.4$&$-0.01$&$0.991$&\\
        & & MPC vs SPC &$0.24\pm0.4$ vs $0.85\pm0.1$&$-5.04$&$0.001$&\!\!\!\!***\\
        & & PC-S vs PC-M &$0.18\pm0.3$ vs $0.24\pm0.4$&$-0.40$&$0.692$&\\
        & & PC-S vs SPC &$0.18\pm0.3$ vs $0.85\pm0.1$&$-7.19$&$0.000$&\!\!\!\!***\\
        & & PC-M vs SPC &$0.24\pm0.4$ vs $0.85\pm0.1$&$-4.97$&$0.001$&\!\!\!\!***\\
        \cmidrule{2-7}
        & \multirow{10}{*}{\textbf{AH-F1}} 
            & CPC vs MPC &$0.37\pm0.3$ vs $0.16\pm0.3$&$1.78$&$0.091$&\!\!\!\!.\\
        & & CPC vs PC-S &$0.37\pm0.3$ vs $0.13\pm0.2$&$2.33$&$0.032$&\!\!\!\!*\\
        & & CPC vs PC-M &$0.37\pm0.3$ vs $0.16\pm0.3$&$1.80$&$0.089$&\!\!\!\!.\\
        & & CPC vs SPC &$0.37\pm0.3$ vs $0.57\pm0.0$&$-2.38$&$0.041$&\!\!\!\!*\\
        & & MPC vs PC-S &$0.16\pm0.3$ vs $0.13\pm0.2$&$0.33$&$0.743$&\\
        & & MPC vs PC-M &$0.16\pm0.3$ vs $0.16\pm0.3$&$0.01$&$0.994$&\\
        & & MPC vs SPC &$0.16\pm0.3$ vs $0.57\pm0.0$&$-4.81$&$0.001$&\!\!\!\!***\\
        & & PC-S vs PC-M &$0.13\pm0.2$ vs $0.16\pm0.3$&$-0.33$&$0.749$&\\
        & & PC-S vs SPC &$0.13\pm0.2$ vs $0.57\pm0.0$&$-6.64$&$0.000$&\!\!\!\!***\\
        & & PC-M vs SPC &$0.16\pm0.3$ vs $0.57\pm0.0$&$-4.84$&$0.001$&\!\!\!\!***\\
        \midrule
        \multirow{20}{*}{\textbf{Insurance}} 
        & \multirow{10}{*}{\textbf{V-F1}} 
            & CPC vs MPC &$0.05\pm0.1$ vs $0.02\pm0.1$&$0.73$&$0.474$&\\
        & & CPC vs PC-S &$0.05\pm0.1$ vs $0.03\pm0.1$&$0.31$&$0.757$&\\
        & & CPC vs PC-M &$0.05\pm0.1$ vs $0.07\pm0.2$&$-0.35$&$0.732$&\\
        & & CPC vs SPC &$0.05\pm0.1$ vs $0.42\pm0.2$&$-4.54$&$0.001$&\!\!\!\!***\\
        & & MPC vs PC-S &$0.02\pm0.1$ vs $0.03\pm0.1$&$-0.34$&$0.736$&\\
        & & MPC vs PC-M &$0.02\pm0.1$ vs $0.07\pm0.2$&$-0.89$&$0.391$&\\
        & & MPC vs SPC &$0.02\pm0.1$ vs $0.42\pm0.2$&$-5.17$&$0.000$&\!\!\!\!***\\
        & & PC-S vs PC-M &$0.03\pm0.1$ vs $0.07\pm0.2$&$-0.59$&$0.567$&\\
        & & PC-S vs SPC &$0.03\pm0.1$ vs $0.42\pm0.2$&$-4.71$&$0.000$&\!\!\!\!***\\
        & & PC-M vs SPC &$0.07\pm0.2$ vs $0.42\pm0.2$&$-3.83$&$0.001$&\!\!\!\!**\\
        \cmidrule{2-7}
        & \multirow{10}{*}{\textbf{AH-F1}} 
            & CPC vs MPC &$0.04\pm0.1$ vs $0.02\pm0.1$&$0.72$&$0.484$&\\
        & & CPC vs PC-S &$0.04\pm0.1$ vs $0.02\pm0.1$&$0.59$&$0.563$&\\
        & & CPC vs PC-M &$0.04\pm0.1$ vs $0.04\pm0.1$&$-0.10$&$0.920$&\\
        & & CPC vs SPC &$0.04\pm0.1$ vs $0.21\pm0.1$&$-3.78$&$0.002$&\!\!\!\!**\\
        & & MPC vs PC-S &$0.02\pm0.1$ vs $0.02\pm0.1$&$-0.11$&$0.911$&\\
        & & MPC vs PC-M &$0.02\pm0.1$ vs $0.04\pm0.1$&$-0.75$&$0.465$&\\
        & & MPC vs SPC &$0.02\pm0.1$ vs $0.21\pm0.1$&$-4.76$&$0.000$&\!\!\!\!***\\
        & & PC-S vs PC-M &$0.02\pm0.1$ vs $0.04\pm0.1$&$-0.64$&$0.532$&\\
        & & PC-S vs SPC &$0.02\pm0.1$ vs $0.21\pm0.1$&$-4.54$&$0.000$&\!\!\!\!***\\
        & & PC-M vs SPC &$0.04\pm0.1$ vs $0.21\pm0.1$&$-3.47$&$0.003$&\!\!\!\!**\\
        \bottomrule
    \end{tabular}
\end{table}

\begin{table}[t]
    \caption{Two-sample, unequal variance t-tests for difference in means for Ecoli70 and Mehra Data. Significance levels: 0 '***', 0.001 '**', 0.01 '*' 0.05 '.' 0.1 ' ' 1. DoF: $n_a=n_b=9$.}
    \label{tab:tests_ecoli70_mehra}
    \vspace{-0.2cm}
    \centering
    \setlength{\tabcolsep}{4pt} 
    \begin{tabular}{c|c|c|c|c|rl}
        \toprule
        \textbf{Dataset} & \textbf{Metric} & \textbf{Methods} & \textbf{Means$\pm$Std} & \textbf{t} & \multicolumn{2}{c}{\textbf{p-value}} \\
        \midrule
        \multirow{20}{*}{\textbf{Ecoli70}} 
        & \multirow{10}{*}{\textbf{V-F1}}
            & CPC vs MPC &$0.01\pm0.0$ vs $0.60\pm0.1$&$-19.57$&$0.000$&\!\!\!\!***\\
        & & CPC vs PC-S &$0.01\pm0.0$ vs $0.29\pm0.2$&$-3.93$&$0.003$&\!\!\!\!**\\
        & & CPC vs PC-M &$0.01\pm0.0$ vs $0.72\pm0.3$&$-7.88$&$0.000$&\!\!\!\!***\\
        & & CPC vs SPC &$0.01\pm0.0$ vs $0.89\pm0.1$&$-23.53$&$0.000$&\!\!\!\!***\\
        & & MPC vs PC-S &$0.60\pm0.1$ vs $0.29\pm0.2$&$3.88$&$0.002$&\!\!\!\!**\\
        & & MPC vs PC-M &$0.60\pm0.1$ vs $0.72\pm0.3$&$-1.33$&$0.210$&\\
        & & MPC vs SPC &$0.60\pm0.1$ vs $0.89\pm0.1$&$-6.19$&$0.000$&\!\!\!\!***\\
        & & PC-S vs PC-M &$0.29\pm0.2$ vs $0.72\pm0.3$&$-3.72$&$0.002$&\!\!\!\!**\\
        & & PC-S vs SPC &$0.29\pm0.2$ vs $0.89\pm0.1$&$-7.39$&$0.000$&\!\!\!\!***\\
        & & PC-M vs SPC &$0.72\pm0.3$ vs $0.89\pm0.1$&$-1.74$&$0.108$&\\
        \cmidrule{2-7}
        & \multirow{10}{*}{\textbf{AH-F1}} 
            & CPC vs MPC &$0.01\pm0.0$ vs $0.55\pm0.1$&$-27.65$&$0.000$&\!\!\!\!***\\
        & & CPC vs PC-S &$0.01\pm0.0$ vs $0.32\pm0.2$&$-4.27$&$0.002$&\!\!\!\!**\\
        & & CPC vs PC-M &$0.01\pm0.0$ vs $0.60\pm0.2$&$-8.52$&$0.000$&\!\!\!\!***\\
        & & CPC vs SPC &$0.01\pm0.0$ vs $0.73\pm0.1$&$-28.17$&$0.000$&\!\!\!\!***\\
        & & MPC vs PC-S &$0.55\pm0.1$ vs $0.32\pm0.2$&$3.12$&$0.011$&\!\!\!\!*\\
        & & MPC vs PC-M &$0.55\pm0.1$ vs $0.60\pm0.2$&$-0.68$&$0.510$&\\
        & & MPC vs SPC &$0.55\pm0.1$ vs $0.73\pm0.1$&$-5.64$&$0.000$&\!\!\!\!***\\
        & & PC-S vs PC-M &$0.32\pm0.2$ vs $0.60\pm0.2$&$-2.82$&$0.011$&\!\!\!\!*\\
        & & PC-S vs SPC &$0.32\pm0.2$ vs $0.73\pm0.1$&$-5.42$&$0.000$&\!\!\!\!***\\
        & & PC-M vs SPC &$0.60\pm0.2$ vs $0.73\pm0.1$&$-1.80$&$0.099$&\!\!\!\!.\\
        \midrule
        \multirow{20}{*}{\textbf{Mehra}} 
        & \multirow{10}{*}{\textbf{V-F1}} 
            & CPC vs MPC &$0.45\pm0.4$ vs $0.26\pm0.4$&$1.04$&$0.311$&\\
        & & CPC vs PC-S &$0.45\pm0.4$ vs $0.01\pm0.0$&$3.49$&$0.007$&\!\!\!\!**\\
        & & CPC vs PC-M &$0.45\pm0.4$ vs $0.28\pm0.5$&$0.91$&$0.377$&\\
        & & CPC vs SPC &$0.45\pm0.4$ vs $0.76\pm0.3$&$-1.98$&$0.065$&\!\!\!\!.\\
        & & MPC vs PC-S &$0.26\pm0.4$ vs $0.01\pm0.0$&$1.88$&$0.092$&\!\!\!\!.\\
        & & MPC vs PC-M &$0.26\pm0.4$ vs $0.28\pm0.5$&$-0.10$&$0.925$&\\
        & & MPC vs SPC &$0.26\pm0.4$ vs $0.76\pm0.3$&$-3.10$&$0.007$&\!\!\!\!**\\
        & & PC-S vs PC-M &$0.01\pm0.0$ vs $0.28\pm0.5$&$-1.89$&$0.091$&\!\!\!\!.\\
        & & PC-S vs SPC &$0.01\pm0.0$ vs $0.76\pm0.3$&$-8.30$&$0.000$&\!\!\!\!***\\
        & & PC-M vs SPC &$0.28\pm0.5$ vs $0.76\pm0.3$&$-2.85$&$0.012$&\!\!\!\!*\\
        \cmidrule{2-7}
        & \multirow{10}{*}{\textbf{AH-F1}} 
            & CPC vs MPC &$0.27\pm0.2$ vs $0.14\pm0.2$&$1.24$&$0.230$&\\
        & & CPC vs PC-S &$0.27\pm0.2$ vs $0.01\pm0.0$&$3.49$&$0.007$&\!\!\!\!**\\
        & & CPC vs PC-M &$0.27\pm0.2$ vs $0.15\pm0.2$&$1.18$&$0.253$&\\
        & & CPC vs SPC &$0.27\pm0.2$ vs $0.41\pm0.2$&$-1.45$&$0.166$&\\
        & & MPC vs PC-S &$0.14\pm0.2$ vs $0.01\pm0.0$&$1.82$&$0.101$&\!\!\!\!.\\
        & & MPC vs PC-M &$0.14\pm0.2$ vs $0.15\pm0.2$&$-0.04$&$0.965$&\\
        & & MPC vs SPC &$0.14\pm0.2$ vs $0.41\pm0.2$&$-2.99$&$0.009$&\!\!\!\!**\\
        & & PC-S vs PC-M &$0.01\pm0.0$ vs $0.15\pm0.2$&$-1.83$&$0.100$&\!\!\!\!.\\
        & & PC-S vs SPC &$0.01\pm0.0$ vs $0.41\pm0.2$&$-8.24$&$0.000$&\!\!\!\!***\\
        & & PC-M vs SPC &$0.15\pm0.2$ vs $0.41\pm0.2$&$-2.87$&$0.011$&\!\!\!\!*\\
        \bottomrule
    \end{tabular}
\end{table}

\clearpage

\clearpage


\clearpage
\subsection{CIT Comparison}
\label{app:cit_comp}
Here we provide a comparison of performance and runtime when changing the CIT used to establish independence.

\subsubsection{Synthetic Data}  
\paragraph{Comparison between Fisher Z and KCI Tests.}  
Tables~\ref{tab:KCI_comp_gauss} and \ref{tab:KCI_comp_nongauss} compare the performance of Shapley-PC and other PC-based methods using Fisher Z and KCI tests across various noise distributions.  
\begin{itemize}
    \item \textbf{Gaussian Noise (Table~\ref{tab:KCI_comp_gauss}):} Fisher Z, aligned with the data's assumptions, achieves higher F1 scores. For example, under Gaussian noise with $d=2$ graphs (ER2 or SF2), Shapley-PC with Fisher Z achieves the highest AH-F1 and V-F1 scores with statistical significance.  
    \item \textbf{Non-Gaussian Noise (Table~\ref{tab:KCI_comp_nongauss}):} While KCI’s nonparametric nature should theoretically perform better, the results are mixed. KCI does not consistently outperform Fisher Z, and variances remain comparable due to the limited sample size ($s=N/|V|=100$). This trend holds across individual noise types (Tables~\ref{tab:KCI_comp_exp}, \ref{tab:KCI_comp_gum}, and \ref{tab:KCI_comp_uni}).  
\end{itemize}  
\paragraph{Influence of Noise Distribution and Graph Structure.}  
Under Gaussian noise, Fisher Z achieves better performance, especially with Shapley-PC (Table~\ref{tab:KCI_comp_gauss}). Non-Gaussian distributions challenge Fisher Z's assumptions, narrowing performance gaps (Tables~\ref{tab:KCI_comp_nongauss} and \ref{tab:KCI_comp_all}).  

Performance trends from the main text persist:  
\begin{itemize}
    \item \textbf{Graph Type (ER vs.\ SF):} SF graphs consistently achieve higher F1 scores due to their hub structure, which simplifies identifying dependencies (Tables~\ref{tab:KCI_comp_gauss} and \ref{tab:KCI_comp_nongauss}).  
    \item \textbf{Edge Density ($d=2$ vs.\ $d=4$):} Sparser graphs ($d=2$) are easier to learn due to smaller and more reliable conditioning sets, yielding higher F1 scores compared to denser graphs ($d=4$). Shapley-PC maintains a leading position in both cases.  
\end{itemize}  
\paragraph{Runtime and Complexity.}  
KCI’s computational complexity is substantially higher than Fisher Z, making it impractical to use large sample sizes in our experiments. For example, the runtime for a single independence test is approximately 0.1 seconds for Fisher Z compared to 200 seconds for KCI (Table~\ref{tab:kci_comp_runtime}). Consequently, we used $N/|V|=100$ as a practical compromise. While the trends observed are consistent with theoretical expectations, this small sample size introduces higher variance, particularly in KCI results, and limits the ability to fully realise its potential advantages. These caveats should be considered when interpreting the comparison.

\begin{table}[hb]
    \centering
    \begin{tabular}{r|cc|cc}
    \toprule
     & \multicolumn{2}{c|}{\textbf{KCI}} & \multicolumn{2}{c}{\textbf{Fisher Z}} \\
    $|\nodeSet|$ & ER & SF & ER & SF \\
    \midrule
        PC-Stable & 181.7 & 96.5 & 0.086 & 0.082 \\
        CPC & 223.3 & 143.2 & 0.1 & 0.104 \\
        MPC & 218.1 & 143.9 & 0.102 & 0.104 \\
        PC-Max & 220.4 & 143.8 & 0.105 & 0.109 \\
        Shapley-PC & 215.9 & 141.1 & 0.103 & 0.106 \\
        \bottomrule
    \end{tabular}
    \caption{Runtime comparison between KCI and Fisher Z: median elapsed time in seconds for ER and SF graphs with nodes $|\nodeSet|=\{10\}$.}
    \label{tab:kci_comp_runtime}
\end{table}

\subsubsection{\texttt{bnlearn} Data}  
\begin{figure*}
    \centering
    \includegraphics[width=0.75\linewidth]{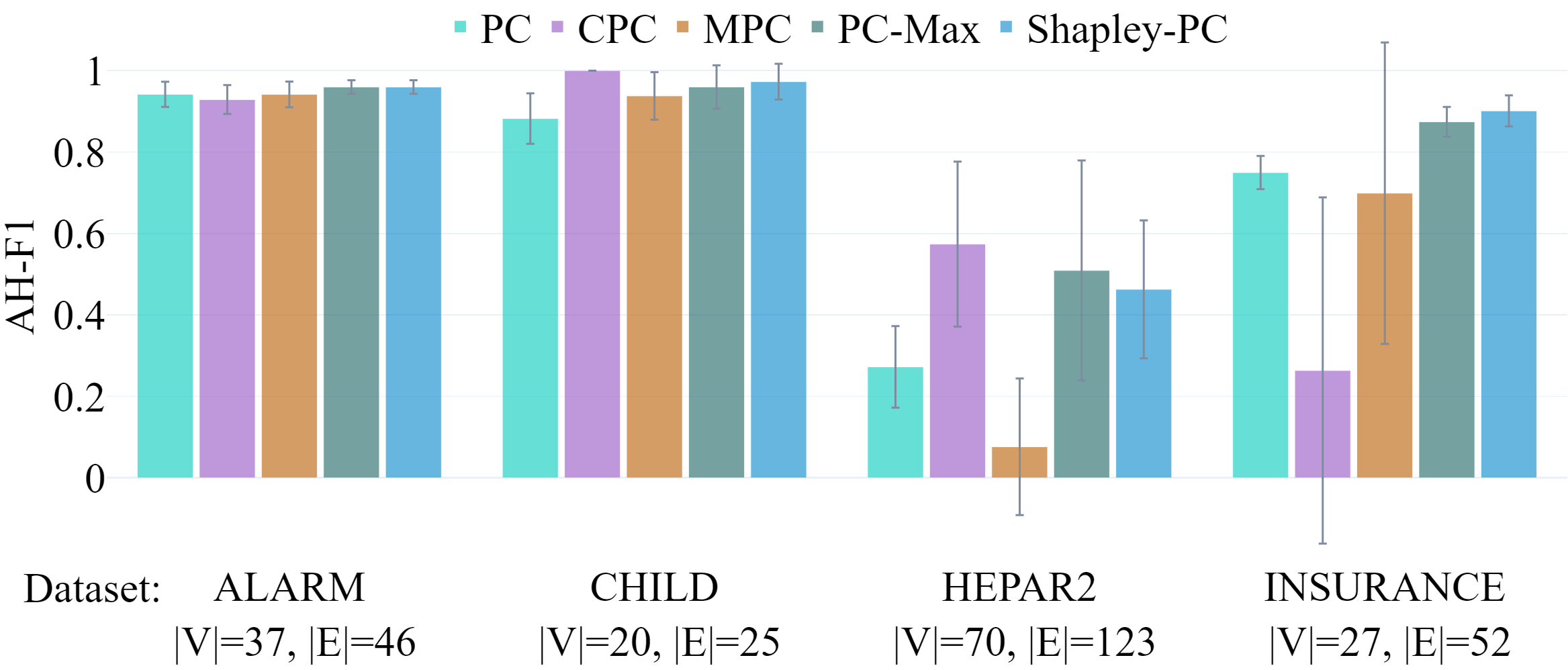}
    \caption{ArrowHead F1 using $\chi^2$ for the discrete BNs of the \texttt{bnlearn} repository.}
    \label{fig:AH_real_chisq_discrete}
\end{figure*}

\begin{figure*}
    \centering
    \includegraphics[width=0.75\linewidth]{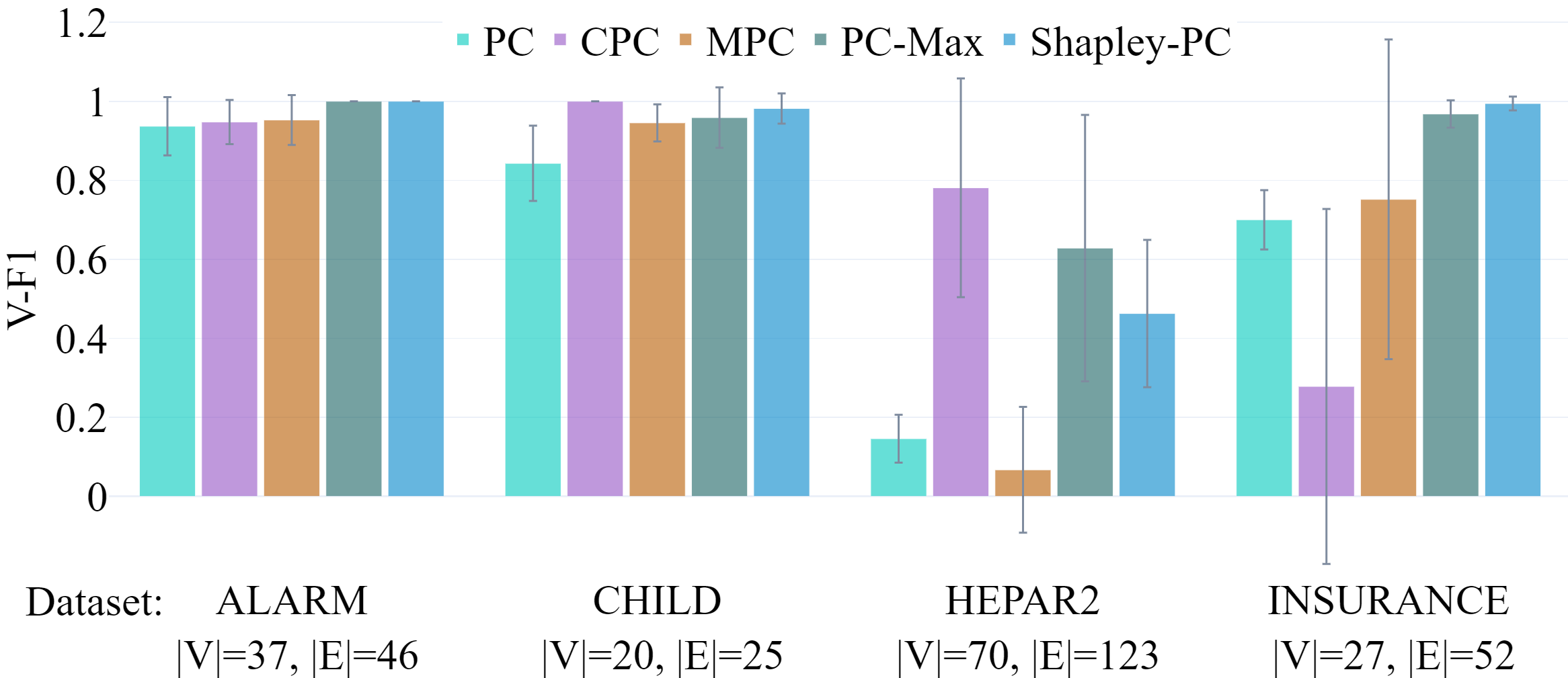}
    \caption{V-Structure F1 using $\chi^2$ for the discrete BNs of the \texttt{bnlearn} repository.}
    \label{fig:V_real_chisq_discrete}
\end{figure*}

\paragraph{Using $\chi^2$ for Discrete Bayesian networks (BNs).}  
In this section, we report results using the $\chi^2$ test, which is better suited for discrete data. Since our synthetic data is all continuous, we analyse potential differences deriving from usage of a different CIT from the one used in the main paper, using the discrete BNs in the \texttt{bnlearn} repository. Results are presented in Fig.~\ref{fig:AH_real_chisq_discrete} and \ref{fig:V_real_chisq_discrete}. By comparing these results to those obtained with the Fisher Z test (Fig.~\ref{fig:pseudoreal}), we gain valuable insights into how the choice of CI test affects performance.  
\begin{itemize}
    \item \textbf{Alarm:} All methods achieve near-optimal performance, with no significant differences across methods.
    \item \textbf{Child:} Here, all PC variants outperform PC-Stable, reflecting the ability of Shapley-PC and other Step 2-enhanced methods to improve performance even when the CI test is well-suited to the data.  
    \item \textbf{Insurance:} Shapley-PC and PC-Max perform significantly better than all other methods. The advantage of these methods in scenarios with more complex dependencies demonstrates their ability to leverage the additional information provided by $p$-values.  
    \item \textbf{Hepar2:} PC-Stable and MPC perform significantly worse than all other methods, while Shapley-PC remains competitive. This highlights Shapley-PC’s adaptability to varying scenarios where CPC and MPC rules might instead fai.  
\end{itemize}  
Comparing these results to those obtained with the Fisher Z test reveals crucial differences. Fisher Z, designed for continuous, linear-Gaussian data, is not well-suited for discrete data. For the \textbf{Alarm} and \textbf{Insurance} datasets, this mismatch leads to a significant performance degradation for all methods except Shapley-PC, which demonstrates a remarkable degree of robustness even under suboptimal conditions.  

This resilience is less pronounced for the \textbf{Child} and \textbf{Hepar2} datasets, where the performance of all methods declines when using Fisher Z. These results highlight the importance of aligning the CI test with the data type to achieve optimal performance. However, even in these scenarios, Shapley-PC maintains a competitive edge, demonstrating its adaptability and reliability across diverse datasets and CI tests.  

Overall, these findings underscore the flexibility of Shapley-PC in handling both appropriate and inappropriate CI tests, and its strong performance when paired with tests suited to the data type, such as $\chi^2$ for discrete BNs.

\begin{table*}[t]
    \caption[Shapley-PC - Synthetic Experiments]{ArrowHead (AH) and V-structure (V) F1 Scores $\pm$ std for ER$d$ and SF$d$ graphs of nodes $\lvert\nodeSet\lvert = 10$. $d$ is the number of edges per node in the true DAG. The proportional sample size is $N/|V|=s=100$ and all considered noise distributions. Bold if significantly different from the runner-up according to a t-test ($\alpha=0.05$)}
    \label{tab:KCI_comp_all}
    \centering
    \begin{tabular}{cc|r|clclclcl}
    \toprule
    & & \textbf{Method} & \textbf{ER2} & & \textbf{ER4} & & \textbf{SF2} & & \textbf{SF4} & \\ 
    \midrule
    \multirow{10}{*}{\textbf{\rotatebox{90}{\textbf{Fisher Z}}}} & \multirow{5}{*}{\textbf{\rotatebox{90}{\textbf{AH-F1}}}} 
    & PC-Stable    & 0.33$\pm$0.18 & & 0.16$\pm$0.12 & & 0.58$\pm$0.18 & & 0.28$\pm$0.17 & \\
    & & CPC        & 0.40$\pm$0.16 & & 0.16$\pm$0.11 & & 0.55$\pm$0.23 & & 0.26$\pm$0.17 & \\
    & & MPC        & 0.42$\pm$0.17 & & 0.15$\pm$0.11 & & 0.59$\pm$0.20 & & 0.31$\pm$0.15 & \\
    & & PC-Max     & \textbf{0.43$\pm$0.16} & & 0.16$\pm$0.13 & & \textbf{0.62$\pm$0.19} & & \textbf{0.34$\pm$0.17} & \\
    & & Shapley-PC & \textbf{0.50$\pm$0.18} & & 0.17$\pm$0.10 & & \textbf{0.68$\pm$0.11} & & \textbf{0.41$\pm$0.15} & \\
    \cmidrule{2-11}
    & \multirow{5}{*}{\textbf{\rotatebox{90}{\textbf{V-F1}}}} 
    & PC-Stable    & 0.43$\pm$0.32 & & 0.17$\pm$0.33 & & 0.79$\pm$0.25 & & 0.49$\pm$0.37 & \\
    & & CPC        & 0.59$\pm$0.31 & & 0.17$\pm$0.31 & & 0.75$\pm$0.30 & & 0.47$\pm$0.40 & \\
    & & MPC        & 0.61$\pm$0.31 & & 0.15$\pm$0.30 & & 0.79$\pm$0.26 & & 0.57$\pm$0.35 & \\
    & & PC-Max     & \textbf{0.66$\pm$0.30} & & 0.18$\pm$0.35 & & 0.87$\pm$0.23 & & 0.63$\pm$0.36 & \\
    & & Shapley-PC & \textbf{0.78$\pm$0.26} & & 0.20$\pm$0.34 & & \textbf{0.95$\pm$0.10} & & \textbf{0.78$\pm$0.27} & \\
    \midrule
    \multirow{10}{*}{\textbf{\rotatebox{90}{\textbf{KCI}}}} & \multirow{5}{*}{\textbf{\rotatebox{90}{\textbf{AH-F1}}}} 
    & PC-Stable    & 0.37$\pm$0.16 & & 0.12$\pm$0.10 & & 0.54$\pm$0.15 & & 0.21$\pm$0.16 & \\
    & & CPC        & 0.38$\pm$0.14 & & 0.13$\pm$0.10 & & 0.58$\pm$0.10 & & 0.22$\pm$0.13 & \\
    & & MPC        & 0.38$\pm$0.16 & & 0.13$\pm$0.10 & & 0.56$\pm$0.15 & & \textbf{0.23$\pm$0.14} & \\
    & & PC-Max     & 0.41$\pm$0.15 & & 0.13$\pm$0.11 & & 0.58$\pm$0.15 & & \textbf{0.31$\pm$0.14} & \\
    & & Shapley-PC & 0.45$\pm$0.19 & & 0.14$\pm$0.11 & & \textbf{0.62$\pm$0.10} & & \textbf{0.33$\pm$0.15} & \\
    \cmidrule{2-11}
    & \multirow{5}{*}{\textbf{\rotatebox{90}{\textbf{V-F1}}}} 
    & PC-Stable    & 0.58$\pm$0.34   & & 0.18$\pm$0.30 & & 0.77$\pm$0.23 & & 0.40$\pm$0.41 & \\
    & & CPC        & 0.60$\pm$0.34 & & 0.15$\pm$0.30 & & 0.83$\pm$0.20 & & 0.43$\pm$0.38 & \\
    & & MPC        & 0.60$\pm$0.35 & & 0.16$\pm$0.31 & & 0.82$\pm$0.25 & & 0.44$\pm$0.38 & \\
    & & PC-Max     & 0.64$\pm$0.32 & & 0.15$\pm$0.33 & & \textbf{0.84$\pm$0.20} & & \textbf{0.62$\pm$0.43} & \\
    & & Shapley-PC & 0.71$\pm$0.34 & & 0.17$\pm$0.33 & & \textbf{0.91$\pm$0.17} & & \textbf{0.68$\pm$0.41} & \\
    \bottomrule
    \end{tabular}
\end{table*}

\begin{table*}[t]
    \caption[Shapley-PC - Synthetic Experiments]{ArrowHead (AH) and V-structure (V) F1 Scores $\pm$ std for ER$d$ and SF$d$ graphs of nodes $\lvert\nodeSet\lvert = 10$. $d$ is the number of edges per node in the true DAG. The proportional sample size is $N/|V|=s=100$ and the noise is Gaussian. Bold if significantly different from the runner-up according to a t-test ($\alpha=0.05$)}
    \label{tab:KCI_comp_gauss}
    \centering
    \begin{tabular}{cc|r|clclclcl}
    \toprule
    & & \textbf{Method} & \textbf{ER2} & & \textbf{ER4} & & \textbf{SF2} & & \textbf{SF4} & \\
    \midrule
    \multirow{10}{*}{\textbf{\rotatebox{90}{\textbf{Fisher Z}}}} & \multirow{5}{*}{\textbf{\rotatebox{90}{\textbf{AH-F1}}}} 
    & PC-Stable   & 0.31$\pm$0.23 & & 0.20$\pm$0.09 & & 0.64$\pm$0.10 & & 0.26$\pm$0.12 & \\
    & & CPC        & \textbf{0.41$\pm$0.18} & & 0.15$\pm$0.06 & & 0.54$\pm$0.23 & & 0.23$\pm$0.15 & \\
    & & MPC        & \textbf{0.43$\pm$0.21} & & 0.15$\pm$0.07 & & 0.57$\pm$0.14 & & 0.28$\pm$0.13 & \\
    & & PC-Max     & \textbf{0.50$\pm$0.07} & & 0.20$\pm$0.12 & & \textbf{0.66$\pm$0.09} & & \textbf{0.28$\pm$0.20} & \\
    & & Shapley-PC & \textbf{0.52$\pm$0.08} & & 0.18$\pm$0.10 & & \textbf{0.70$\pm$0.08} & & \textbf{0.40$\pm$0.10} & \\
    \cmidrule{2-11}
    & \multirow{5}{*}{\textbf{\rotatebox{90}{\textbf{V-F1}}}} 
    & PC-Stable   & 0.39$\pm$0.30 & & 0.22$\pm$0.37 & & 0.86$\pm$0.16 & & 0.50$\pm$0.33 & \\
    & & CPC        & \textbf{0.60$\pm$0.33} & & 0.23$\pm$0.39 & & 0.70$\pm$0.33 & & 0.31$\pm$0.30 & \\
    & & MPC        & \textbf{0.62$\pm$0.33} & & 0.22$\pm$0.37 & & 0.74$\pm$0.21 & & 0.50$\pm$0.28 & \\
    & & PC-Max     & \textbf{0.79$\pm$0.17} & & 0.30$\pm$0.48 & & \textbf{0.92$\pm$0.10} & & 0.51$\pm$0.41 & \\
    & & Shapley-PC & \textbf{0.83$\pm$0.11} & & 0.23$\pm$0.39 & & \textbf{0.97$\pm$0.05} & & \textbf{0.74$\pm$0.29} & \\
    \midrule
    \multirow{10}{*}{\textbf{\rotatebox{90}{\textbf{KCI}}}} & \multirow{5}{*}{\textbf{\rotatebox{90}{\textbf{AH-F1}}}} 
    & PC-Stable    & 0.27$\pm$0.14 & & 0.09$\pm$0.11 & & 0.53$\pm$0.15 & & 0.24$\pm$0.16 & \\
    & & CPC        & 0.32$\pm$0.17 & & 0.10$\pm$0.10 & & 0.57$\pm$0.09 & & 0.27$\pm$0.15 & \\
    & & MPC        & 0.27$\pm$0.19 & & 0.10$\pm$0.14 & & 0.57$\pm$0.11 & & 0.30$\pm$0.15 & \\
    & & PC-Max     & 0.32$\pm$0.17 & & 0.11$\pm$0.09 & & 0.60$\pm$0.10 & & 0.30$\pm$0.14 & \\
    & & Shapley-PC & 0.38$\pm$0.16 & & 0.11$\pm$0.08 & & 0.61$\pm$0.09 & & 0.33$\pm$0.16 & \\
    \cmidrule{2-11}
    & \multirow{5}{*}{\textbf{\rotatebox{90}{\textbf{V-F1}}}} 
    & PC-Stable   & 0.38$\pm$0.37 & & 0.11$\pm$0.32 & & 0.81$\pm$0.24 & & 0.52$\pm$0.38 & \\
    & & CPC        & 0.46$\pm$0.43 & & 0.15$\pm$0.30 & & 0.86$\pm$0.15 & & 0.49$\pm$0.38 & \\
    & & MPC        & 0.35$\pm$0.42 & & 0.09$\pm$0.30 & & 0.85$\pm$0.15 & & 0.55$\pm$0.36 & \\
    & & PC-Max     & 0.46$\pm$0.44 & & 0.10$\pm$0.32 & & 0.94$\pm$0.09 & & 0.64$\pm$0.46 & \\
    & & Shapley-PC & 0.53$\pm$0.38 & & 0.11$\pm$0.30 & & 0.95$\pm$0.09 & & 0.70$\pm$0.39 & \\
    \bottomrule
    \end{tabular}
\end{table*}

\begin{table*}[t]
    \caption[Shapley-PC - Synthetic Experiments]{ArrowHead (AH) and V-structure (V) F1 Scores $\pm$ std for ER$d$ and SF$d$ graphs of nodes $\lvert\nodeSet\lvert = 10$. $d$ is the number of edges per node in the true DAG. The proportional sample size is $N/|V|=s=100$ and the noise is Non-Gaussian. Bold if significantly different from the runner-up according to a t-test ($\alpha=0.05$).}
    \label{tab:KCI_comp_nongauss}
    \centering
    \begin{tabular}{cc|r|clclclcl}
    \toprule
    & & \textbf{Method} & \textbf{ER2} & & \textbf{ER4} & & \textbf{SF2} & & \textbf{SF4} & \\
    \midrule
    \multirow{10}{*}{\textbf{\rotatebox{90}{\textbf{Fisher Z}}}} & \multirow{5}{*}{\textbf{\rotatebox{90}{\textbf{AH-F1}}}} 
    & PC-Stable    & 0.33$\pm$0.17 & & 0.15$\pm$0.13 & & 0.55$\pm$0.20 & & 0.29$\pm$0.19 & \\
    & & CPC        & \textbf{0.40$\pm$0.16} & & 0.16$\pm$0.12 & & 0.56$\pm$0.23 & & 0.28$\pm$0.18 & \\
    & & MPC        & \textbf{0.41$\pm$0.16} & & 0.14$\pm$0.12 & & \textbf{0.59$\pm$0.21} & & 0.33$\pm$0.16 & \\
    & & PC-Max     & \textbf{0.41$\pm$0.18} & & 0.15$\pm$0.13 & & \textbf{0.61$\pm$0.22} & & \textbf{0.37$\pm$0.15} & \\
    & & Shapley-PC & \textbf{0.50$\pm$0.20} & & 0.17$\pm$0.10 & & \textbf{0.68$\pm$0.12} & & \textbf{0.41$\pm$0.16} & \\
    \cmidrule{2-11}
    & \multirow{5}{*}{\textbf{\rotatebox{90}{\textbf{V-F1}}}} 
    & PC-Stable    & 0.45$\pm$0.33 & & 0.15$\pm$0.32 & & 0.77$\pm$0.27 & & 0.49$\pm$0.39 & \\
    & & CPC        & 0.59$\pm$0.31 & & 0.15$\pm$0.28 & & 0.77$\pm$0.30 & & 0.52$\pm$0.42 & \\
    & & MPC        & 0.60$\pm$0.30 & & 0.13$\pm$0.28 & & 0.80$\pm$0.28 & & 0.59$\pm$0.37 & \\
    & & PC-Max     & \textbf{0.62$\pm$0.33} & & 0.14$\pm$0.29 & & 0.85$\pm$0.26 & & \textbf{0.67$\pm$0.34} & \\
    & & Shapley-PC & \textbf{0.76$\pm$0.30} & & 0.19$\pm$0.33 & & 0.95$\pm$0.11 & & \textbf{0.79$\pm$0.26} & \\
    \midrule
    \multirow{10}{*}{\textbf{\rotatebox{90}{\textbf{KCI}}}} & \multirow{5}{*}{\textbf{\rotatebox{90}{\textbf{AH-F1}}}} 
    & PC-Stable    & 0.41$\pm$0.15 & & 0.13$\pm$0.09 & & 0.55$\pm$0.15 & & 0.20$\pm$0.16 & \\
    & & CPC        & 0.40$\pm$0.12 & & 0.14$\pm$0.10 & & 0.58$\pm$0.16 & & 0.20$\pm$0.12 & \\
    & & MPC        & 0.42$\pm$0.13 & & 0.14$\pm$0.10 & & 0.56$\pm$0.20 & & 0.21$\pm$0.14 & \\
    & & PC-Max     & 0.43$\pm$0.14 & & 0.13$\pm$0.11 & & 0.57$\pm$0.20 & & \textbf{0.31$\pm$0.13} & \\
    & & Shapley-PC & 0.47$\pm$0.19 & & 0.15$\pm$0.11 & & 0.62$\pm$0.15 & & \textbf{0.33$\pm$0.15} & \\
    \cmidrule{2-11}
    & \multirow{5}{*}{\textbf{\rotatebox{90}{\textbf{V-F1}}}} 
    & PC-Stable    & 0.64$\pm$0.31 & & 0.21$\pm$0.36 & & 0.75$\pm$0.23 & & 0.36$\pm$0.41 & \\
    & & CPC        & 0.64$\pm$0.30 & & 0.16$\pm$0.30 & & 0.82$\pm$0.21 & & 0.41$\pm$0.38 & \\
    & & MPC        & 0.68$\pm$0.29 & & 0.18$\pm$0.31 & & 0.80$\pm$0.28 & & 0.40$\pm$0.38 & \\
    & & PC-Max     & 0.70$\pm$0.25 & & 0.16$\pm$0.34 & & 0.81$\pm$0.26 & & \textbf{0.62$\pm$0.43} & \\
    & & Shapley-PC & 0.77$\pm$0.30 & & 0.19$\pm$0.34 & & 0.90$\pm$0.19 & & \textbf{0.67$\pm$0.42} & \\
    \bottomrule
    \end{tabular}
\end{table*}

\begin{table*}[t]
    \caption[Shapley-PC - Synthetic Experiments]{ArrowHead (AH) and V-structure (V) F1 Scores $\pm$ std for ER$d$ and SF$d$ graphs of nodes $\lvert\nodeSet\lvert = 10$. $d$ is the number of edges per node in the true DAG. The proportional sample size is $N/|V|=s=100$ and the noise is Exponential. Bold if significantly different from the runner-up according to a t-test ($\alpha=0.05$).}
    \label{tab:KCI_comp_exp}
    \centering
    \begin{tabular}{cc|r|clclclcl}
    \toprule
    & & \textbf{Method} & \textbf{ER2} & & \textbf{ER4} & & \textbf{SF2} & & \textbf{SF4} & \\ 
    \midrule
    \multirow{10}{*}{\textbf{\rotatebox{90}{\textbf{Fisher Z}}}} & \multirow{5}{*}{\textbf{\rotatebox{90}{\textbf{AH-F1}}}} 
    & PC-Stable   & 0.34$\pm$0.16 & & 0.16$\pm$0.14 & & 0.57$\pm$0.10 & & 0.28$\pm$0.21 & \\
    & & CPC        & 0.38$\pm$0.18 & & 0.15$\pm$0.12 & & \textbf{0.58$\pm$0.24} & & 0.28$\pm$0.18 & \\
    & & MPC        & 0.41$\pm$0.15 & & 0.14$\pm$0.11 & & \textbf{0.64$\pm$0.15} & & \textbf{0.32$\pm$0.15} & \\
    & & PC-Max     & 0.37$\pm$0.23 & & 0.10$\pm$0.12 & & \textbf{0.68$\pm$0.11} & & \textbf{0.43$\pm$0.10} & \\
    & & Shapley-PC & 0.44$\pm$0.23 & & 0.17$\pm$0.12 & & \textbf{0.71$\pm$0.09} & & \textbf{0.44$\pm$0.14} & \\
    \cmidrule{2-11}
    & \multirow{5}{*}{\textbf{\rotatebox{90}{\textbf{V-F1}}}} 
    & PC-Stable   & 0.46$\pm$0.37 & & 0.17$\pm$0.36 & & 0.80$\pm$0.14 & & 0.44$\pm$0.41 & \\
    & & CPC        & 0.54$\pm$0.36 & & 0.18$\pm$0.30 & & \textbf{0.80$\pm$0.30} & & 0.42$\pm$0.41 & \\
    & & MPC        & 0.62$\pm$0.31 & & 0.13$\pm$0.28 & & \textbf{0.88$\pm$0.17} & & \textbf{0.49$\pm$0.39} & \\
    & & PC-Max     & 0.58$\pm$0.42 & & 0.07$\pm$0.21 & & \textbf{0.94$\pm$0.08} & & \textbf{0.77$\pm$0.29} & \\
    & & Shapley-PC & 0.69$\pm$0.39 & & 0.12$\pm$0.25 & & \textbf{0.98$\pm$0.03} & & \textbf{0.83$\pm$0.18} & \\
    \midrule
    \multirow{10}{*}{\textbf{\rotatebox{90}{\textbf{KCI}}}} & \multirow{5}{*}{\textbf{\rotatebox{90}{\textbf{AH-F1}}}} 
    & PC-Stable   & \textbf{0.44$\pm$0.22} & & 0.14$\pm$0.11 & & 0.54$\pm$0.17 & & 0.19$\pm$0.19 & \\
    & & CPC        & 0.40$\pm$0.15 & & 0.16$\pm$0.10 & & 0.59$\pm$0.23 & & 0.20$\pm$0.11 & \\
    & & MPC        & \textbf{0.43$\pm$0.14} & & 0.18$\pm$0.09 & & 0.58$\pm$0.24 & & 0.18$\pm$0.15 & \\
    & & PC-Max     & \textbf{0.49$\pm$0.14} & & 0.14$\pm$0.13 & & 0.60$\pm$0.22 & & \textbf{0.34$\pm$0.14} & \\
    & & Shapley-PC & \textbf{0.53$\pm$0.13} & & 0.15$\pm$0.12 & & 0.62$\pm$0.24 & & \textbf{0.37$\pm$0.18} & \\
    \cmidrule{2-11}
    & \multirow{5}{*}{\textbf{\rotatebox{90}{\textbf{V-F1}}}} 
    & PC-Stable   & \textbf{0.69$\pm$0.36 } & & 0.21$\pm$0.35 & & 0.69$\pm$0.29 & & 0.29$\pm$0.41 & \\
    & & CPC        & \textbf{ 0.57$\pm$0.36} & & 0.20$\pm$0.32 & & 0.80$\pm$0.30 & & 0.36$\pm$0.32 & \\
    & & MPC        & 0.63$\pm$0.30 & & 0.25$\pm$0.33 & & 0.80$\pm$0.34 & & 0.29$\pm$0.33 & \\
    & & PC-Max     & \textbf{0.80$\pm$0.19} & & 0.27$\pm$0.44 & & 0.83$\pm$0.30 & & \textbf{0.69$\pm$0.38} & \\
    & & Shapley-PC & \textbf{0.88$\pm$0.16} & & 0.28$\pm$0.41 & & 0.86$\pm$0.31 & & \textbf{0.71$\pm$0.39} & \\
    \bottomrule
    \end{tabular}
\end{table*}

\begin{table*}[t]
    \caption[Shapley-PC - Synthetic Experiments]{ArrowHead (AH) and V-structure (V) F1 Scores $\pm$ std for ER$d$ and SF$d$ graphs of nodes $\lvert\nodeSet\lvert = 10$. $d$ is the number of edges per node in the true DAG. The proportional sample size is $N/|V|=s=100$ and the noise is Gumbel. Bold if significantly different from the runner-up according to a t-test ($\alpha=0.05$).}
    \label{tab:KCI_comp_gum}
    \centering
    \begin{tabular}{cc|r|clclclcl}
    \toprule
    & & \textbf{Method} & \textbf{ER2} & & \textbf{ER4} & & \textbf{SF2} & & \textbf{SF4} & \\ 
    \midrule
    \multirow{10}{*}{\textbf{\rotatebox{90}{\textbf{Fisher Z}}}} & \multirow{5}{*}{\textbf{\rotatebox{90}{\textbf{AH-F1}}}} 
    & PC-Stable   & 0.33$\pm$0.21 & & 0.16$\pm$0.13 & & 0.58$\pm$0.26 & & 0.28$\pm$0.15 & \\
    & & CPC        & \textbf{0.43$\pm$0.12} & & 0.19$\pm$0.12 & & 0.59$\pm$0.24 & & 0.24$\pm$0.20 & \\
    & & MPC        & \textbf{0.43$\pm$0.15} & & 0.16$\pm$0.13 & & 0.63$\pm$0.25 & & 0.30$\pm$0.16 & \\
    & & PC-Max     & \textbf{0.47$\pm$0.12} & & 0.18$\pm$0.11 & & 0.54$\pm$0.31 & & 0.31$\pm$0.20 & \\
    & & Shapley-PC & \textbf{0.53$\pm$0.17} & & 0.15$\pm$0.09 & & 0.69$\pm$0.13 & & 0.40$\pm$0.16 & \\
    \cmidrule{2-11}
    & \multirow{5}{*}{\textbf{\rotatebox{90}{\textbf{V-F1}}}} 
    & PC-Stable   & 0.45$\pm$0.33 & & 0.17$\pm$0.36 & & 0.78$\pm$0.34 & &           0.47$\pm$0.33 & \\
    & & CPC        & 0.67$\pm$0.20 & & 0.07$\pm$0.21 & & 0.80$\pm$0.30 & &          \textbf{0.49$\pm$0.44} & \\
    & & MPC        & 0.64$\pm$0.24 & & 0.07$\pm$0.21 & & 0.82$\pm$0.31 & &          \textbf{0.53$\pm$0.36} & \\
    & & PC-Max     & \textbf{0.75$\pm$0.14} & & 0.07$\pm$0.21 & & 0.74$\pm$0.41 & & \textbf{0.55$\pm$0.40} & \\
    & & Shapley-PC & \textbf{0.84$\pm$0.15} & & 0.07$\pm$0.21 & & 0.95$\pm$0.13 & & \textbf{0.80$\pm$0.14} & \\
    \midrule
    \multirow{10}{*}{\textbf{\rotatebox{90}{\textbf{KCI}}}} & \multirow{5}{*}{\textbf{\rotatebox{90}{\textbf{AH-F1}}}} 
    & PC-Stable   & 0.39$\pm$0.13 & & 0.14$\pm$0.08 & & 0.52$\pm$0.16 & & 0.22$\pm$0.18 & \\
    & & CPC        & 0.42$\pm$0.11 & & 0.15$\pm$0.10 & & 0.57$\pm$0.13 & & 0.21$\pm$0.16 & \\
    & & MPC        & 0.45$\pm$0.12 & & 0.10$\pm$0.08 & & 0.52$\pm$0.23 & & 0.25$\pm$0.16 & \\
    & & PC-Max     & 0.44$\pm$0.14 & & 0.15$\pm$0.12 & & 0.54$\pm$0.23 & & 0.32$\pm$0.12 & \\
    & & Shapley-PC & 0.42$\pm$0.27 & & 0.17$\pm$0.13 & & 0.65$\pm$0.10 & & 0.32$\pm$0.14 & \\
    \cmidrule{2-11}
    & \multirow{5}{*}{\textbf{\rotatebox{90}{\textbf{V-F1}}}} 
    & PC-Stable   & 0.67$\pm$0.27 & & 0.21$\pm$0.30 & & 0.69$\pm$0.20 & & 0.42$\pm$0.46 & \\
    & & CPC        & 0.75$\pm$0.20 & & 0.20$\pm$0.12 & & \textbf{0.77$\pm$0.13} & & \textbf{0.43$\pm$0.40} & \\
    & & MPC        & 0.82$\pm$0.20 & & 0.12$\pm$0.12 & & \textbf{0.70$\pm$0.31} & & \textbf{0.47$\pm$0.42} & \\
    & & PC-Max     & 0.75$\pm$0.18 & & 0.05$\pm$0.16 & & \textbf{0.72$\pm$0.31} & & \textbf{0.59$\pm$0.43} & \\
    & & Shapley-PC & 0.69$\pm$0.39 & & 0.15$\pm$0.34 & & \textbf{0.91$\pm$0.11} & & \textbf{0.66$\pm$0.46} & \\
    \bottomrule
    \end{tabular}
\end{table*}

\begin{table*}[t]
    \caption[Shapley-PC - Synthetic Experiments]{ArrowHead (AH) and V-structure (V) F1 Scores $\pm$ std for ER$d$ and SF$d$ graphs of nodes $\lvert\nodeSet\lvert = 10$. $d$ is the number of edges per node in the true DAG. The proportional sample size is $N/|V|=s=100$ and the noise is Uniform. Bold if significantly different from the runner-up according to a t-test ($\alpha=0.05$).}
    \label{tab:KCI_comp_uni}
    \centering
    \begin{tabular}{cc|r|clclclcl}
    \toprule
    & & \textbf{Method} & \textbf{ER2} & & \textbf{ER4} & & \textbf{SF2} & & \textbf{SF4} & \\ 
    \midrule
    \multirow{10}{*}{\textbf{\rotatebox{90}{\textbf{Fisher Z}}}} & \multirow{5}{*}{\textbf{\rotatebox{90}{\textbf{AH-F1}}}} 
    & PC-Stable   & 0.34$\pm$0.15 & & 0.13$\pm$0.13 & & 0.52$\pm$0.22 & & 0.30$\pm$0.23 & \\
    & & CPC        & \textbf{0.40$\pm$0.18} & & 0.13$\pm$0.12 & & 0.49$\pm$0.22 & & 0.31$\pm$0.18 & \\
    & & MPC        & \textbf{0.40$\pm$0.18} & & 0.13$\pm$0.12 & & 0.52$\pm$0.23 & & 0.37$\pm$0.16 & \\
    & & PC-Max     & \textbf{0.40$\pm$0.18} & & 0.16$\pm$0.14 & & 0.60$\pm$0.16 & & 0.36$\pm$0.14 & \\
    & & Shapley-PC & \textbf{0.53$\pm$0.20} & & 0.18$\pm$0.08 & & 0.63$\pm$0.14 & & 0.39$\pm$0.20 & \\
    \cmidrule{2-11}
    & \multirow{5}{*}{\textbf{\rotatebox{90}{\textbf{V-F1}}}} 
    & PC-Stable   & 0.42$\pm$0.33 & & 0.12$\pm$0.25 & & 0.72$\pm$0.32 & & 0.55$\pm$0.44 & \\
    & & CPC        & \textbf{0.54$\pm$0.37} & & 0.19$\pm$0.34 & & 0.70$\pm$0.31 & & 0.65$\pm$0.41 & \\
    & & MPC        & \textbf{0.54$\pm$0.37} & & 0.19$\pm$0.34 & & 0.71$\pm$0.32 & & 0.74$\pm$0.35 & \\
    & & PC-Max     & \textbf{0.53$\pm$0.34} & & 0.28$\pm$0.39 & & 0.86$\pm$0.16 & & 0.68$\pm$0.32 & \\
    & & Shapley-PC & \textbf{0.76$\pm$0.32} & & 0.38$\pm$0.43 & & 0.92$\pm$0.15 & & 0.73$\pm$0.40 & \\
    \midrule
    \multirow{10}{*}{\textbf{\rotatebox{90}{\textbf{KCI}}}} & \multirow{5}{*}{\textbf{\rotatebox{90}{\textbf{AH-F1}}}} 
    & PC-Stable   & 0.40$\pm$0.11 & & 0.11$\pm$0.09 & & 0.58$\pm$0.11 & & 0.19$\pm$0.12 & \\
    & & CPC        & 0.39$\pm$0.12 & & 0.12$\pm$0.10 & & 0.58$\pm$0.12 & & 0.19$\pm$0.10 & \\
    & & MPC        & 0.37$\pm$0.12 & & 0.16$\pm$0.11 & & 0.59$\pm$0.14 & & \textbf{0.21$\pm$0.11} & \\
    & & PC-Max     & 0.37$\pm$0.12 & & 0.11$\pm$0.09 & & 0.57$\pm$0.13 & & \textbf{0.28$\pm$0.15} & \\
    & & Shapley-PC & 0.47$\pm$0.14 & & 0.14$\pm$0.09 & & 0.60$\pm$0.10 & & \textbf{0.31$\pm$0.13} & \\
    \cmidrule{2-11}
    & \multirow{5}{*}{\textbf{\rotatebox{90}{\textbf{V-F1}}}} 
    & PC-Stable   & 0.57$\pm$0.32 & & 0.20$\pm$0.42 & & 0.89$\pm$0.13 & & 0.38$\pm$0.40 & \\
    & & CPC        & 0.61$\pm$0.33 & & 0.17$\pm$0.36 & & 0.88$\pm$0.13 & & 0.44$\pm$0.44 & \\
    & & MPC        & 0.60$\pm$0.32 & & 0.17$\pm$0.36 & & 0.89$\pm$0.11 & & 0.43$\pm$0.40 & \\
    & & PC-Max     & 0.55$\pm$0.31 & & 0.17$\pm$0.36 & & 0.87$\pm$0.16 & & 0.57$\pm$0.50 & \\
    & & Shapley-PC & 0.74$\pm$0.31 & & 0.13$\pm$0.28 & & 0.94$\pm$0.09 & & 0.65$\pm$0.46 & \\
    \bottomrule
    \end{tabular}
\end{table*}

\end{document}